\pdfoutput=1

\documentclass[11pt]{article}

\usepackage[final]{acl}

\usepackage{times}
\usepackage{latexsym}
\usepackage{url}
\usepackage{booktabs}
\usepackage{adjustbox}
\usepackage{colortbl}
\usepackage{threeparttable}
\usepackage{makecell}
\usepackage{multirow} 
\usepackage{subcaption}
\usepackage{amsmath}
\usepackage{amssymb}
\usepackage{amsthm}
\usepackage{capt-of}
\usepackage{soul}
\usepackage{pifont}
\usepackage{enumitem}

\newtheorem{proposition}{Proposition}

\newtheorem{corollary}{Corollary}
\newtheorem{remark}{Remark}
\newtheorem{law}{Law}

\usepackage{listings}
\definecolor{codeblue}{rgb}{0.25,0.5,0.5}
\usepackage[T1]{fontenc}

\usepackage[utf8]{inputenc}

\usepackage{microtype}

\usepackage{inconsolata}

\usepackage{graphicx}

\title{Towards the Law of Capacity Gap in Distilling Language Models}

\author{Chen Zhang\textsuperscript{\ding{168}}, Qiuchi Li\textsuperscript{\ding{72}}, Dawei Song\textsuperscript{\ding{168}\ding{70}}\Thanks{ Dawei Song is the corresponding author.}, Zheyu Ye\textsuperscript{\ding{170}}, Yan Gao\textsuperscript{\ding{170}}, Yan Hu\textsuperscript{\ding{170}} \\
  \textsuperscript{\ding{168}}Beijing Institute of Technology\quad\textsuperscript{\ding{72}}University of Copenhagen \\
  \textsuperscript{\ding{70}}The Open University\quad\textsuperscript{\ding{170}}Xiaohongshu \\
  \texttt{chenzhang9702@outlook.com} \\}

\begin{document}
\maketitle

\begin{abstract}
Language model (LM) distillation aims at distilling the knowledge in a large teacher LM to a small student one. As a critical issue facing LM distillation, a superior student often arises from a teacher of a relatively small scale instead of a larger one, especially in the presence of substantial capacity gap between the teacher and student. This issue, often referred to as the \textit{curse of capacity gap}, suggests that there is likely an optimal teacher yielding the best-performing student along the scaling course of the teacher. Consequently, distillation trials on teachers of a wide range of scales are called for to determine the optimal teacher, which becomes computationally intensive in the context of large LMs (LLMs). This paper addresses this critical bottleneck by providing the \textit{law of capacity gap} inducted from a preliminary study on distilling a broad range of small-scale (<3B) LMs, where the optimal teacher consistently scales linearly with the student scale across different model and data scales. By extending the law to LLM distillation on a larger scale (7B), we succeed in obtaining versatile LLMs that outperform a wide array of competitors.\footnote{The repository is at \url{https://github.com/GeneZC/MiniMA}. The model collection is placed at \url{https://huggingface.co/collections/GeneZC/minima-family-6695f13d461de4eea59d83a3}}
\end{abstract}

%

\section{Introduction}

Language model (LM) distillation is designed to distill from a large teacher LM to a small student LM. With the regime of the teacher-student paradigm, it is expected that the student LM could mimic the behaviors of the teacher LM and achieve appealing performance as the teacher LM does~\citep{DBLP:journals/corr/HintonVD15,DBLP:journals/corr/SanhDC19,DBLP:conf/emnlp/JiaoYSJCL0L20}. 

\begin{figure}[ht]
    \centering
    \includegraphics[width=0.47\textwidth]{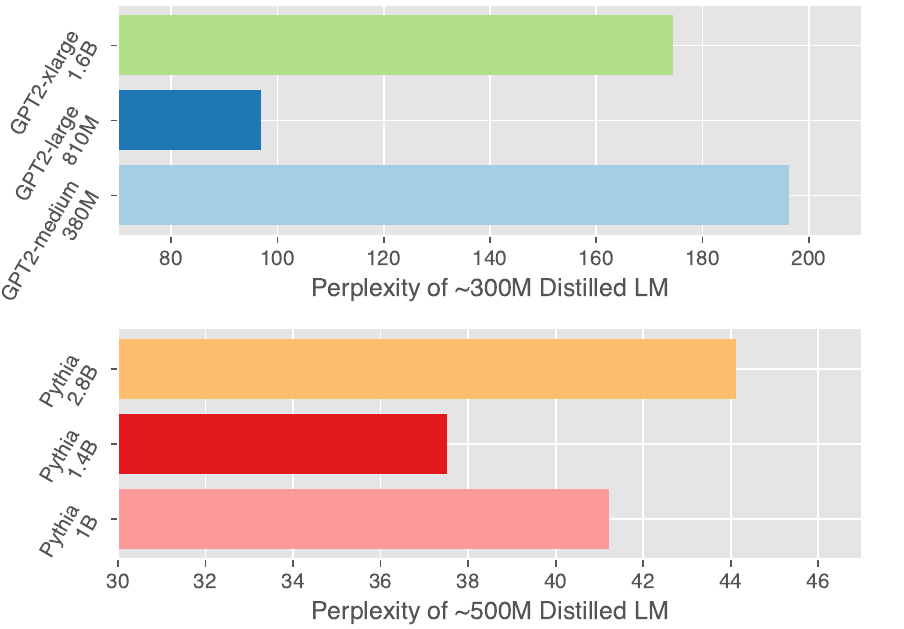}
    \caption{The curse of capacity gap. GPT2~\citep{DBLP:journals/corr/RadfordWC19} and Pythia~\citep{DBLP:conf/icml/BidermanSABOHKP23} distilled with OpenWebText~\citep{GokaslanC19}, and evaluated on WikiText2~\citep{DBLP:conf/iclr/MerityX0S17} in perplexity. \textit{curse}, the performance of a fixed student scale does not improve along the increased teacher scale.}
    \label{fig:intro_curse}
    \vspace{-5mm}
\end{figure}

\begin{figure}[ht]
    \centering
    \centering
    \includegraphics[width=0.40\textwidth]{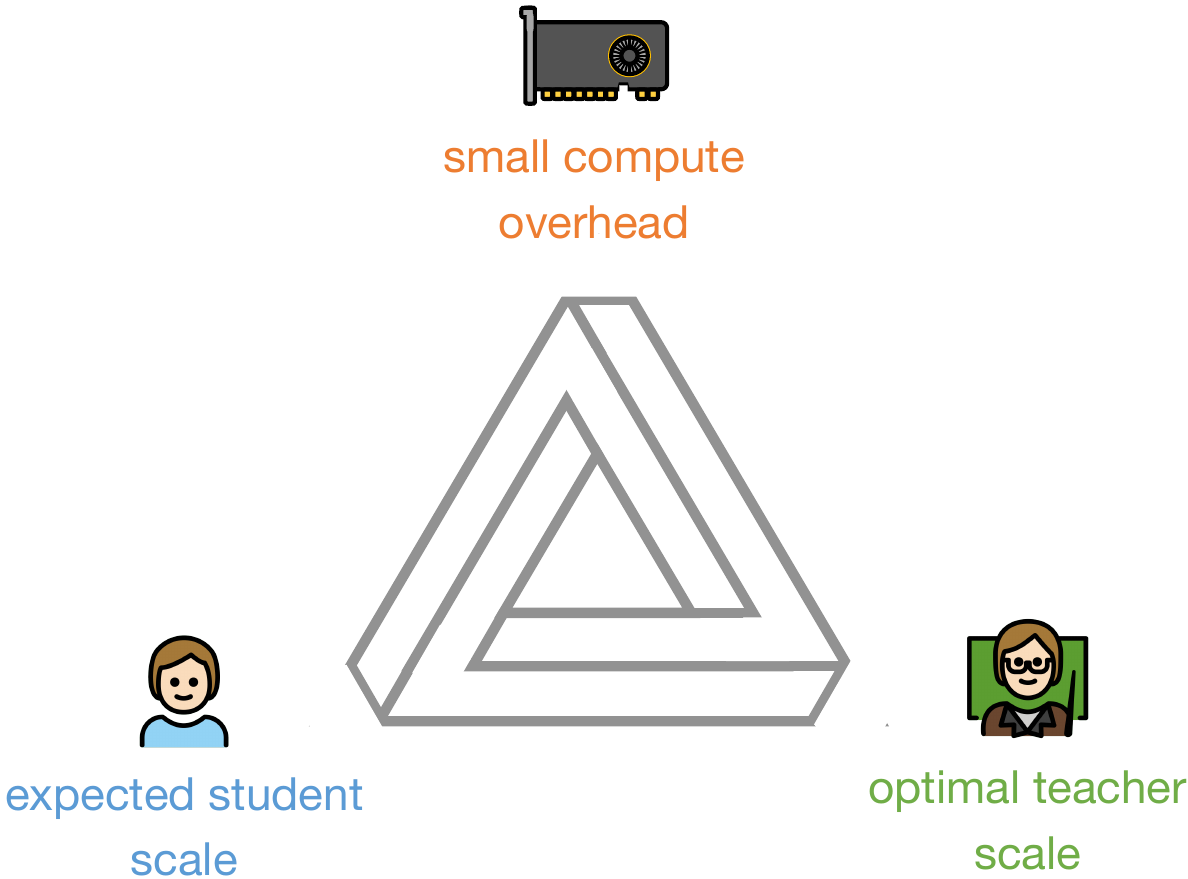}
    \caption{The curse of capacity gap can result in an impossible triangle in the era of LLMs. Optimal teacher scale according to expected student scale can not be yielded via small compute overhead, besides the one required by the oracle distillation.}
    \label{fig:impossible_triangle}
    \vspace{-5mm}
\end{figure}

However, it has been broadly observed that the performance of student LM is not always improved but often degraded as the teacher scales up, termed as the \textit{curse of capacity gap}. As shown in Figure~\ref{fig:intro_curse}, the optimal teacher that accords with the best student performance shall not be the largest one~\citep{DBLP:conf/aaai/MirzadehFLLMG20,DBLP:journals/corr/ZhangYW22,DBLP:conf/acl/ZhouXM22,DBLP:conf/emnlp/YangZS22}. Consequently, computationally enumerative distillation trials need to be conducted from different scales of teacher LMs in order to obtain the optimal teacher. 
This can evolve into a critical bottleneck in the context of large LMs~\citep[LLMs,][]{DBLP:conf/nips/BrownMRSKDNSSAA20,DBLP:journals/jmlr/ChowdheryNDBMRBCSGSSTMRBTSPRDHPBAI23,DBLP:journals/corr/TouvronLI23,OpenAI22}. Essentially, an \textit{impossible triangle} has been evoked among expected student scale, optimal teacher scale, and small compute overhead as shown in Figure~\ref{fig:impossible_triangle}.

This paper addresses the impossible triangle of capacity gap by exploring whether a fixed relation exists between the expected student scale and its optimal teacher scale, namely the \textit{law of capacity gap}. Through a pilot study, we empirically reveal a constantly linear correlation between the student scale and optimal teacher scale, which holds perfectly across varying model and data scales. This transforms the impossible triangle to a possible one by eliminating the compute overhead involved in determining the optimal teacher scale.

To empirically validate the broad applicability of the law, we are driven to distill LMs of larger scales based off the law. Specifically, we distill common LLMs to the most suitable student scales under the law, and compared the performance of the student LMs to a wide range of efficient designs on an extensive variety of benchmarks. The results demonstrate that the law still holds and the student LMs bring superior performance over baselines with the same compute budget. The student LMs also possess strong instruction-following capabilities after finetuning, surpassing an array of scale-matched competitors and rivaling LLMs of larger scales.

To sum up, our contributions are summarized to three folds:
\begin{itemize}[noitemsep,topsep=1pt]
    \item We have put forth the impossible triangle as a result of the curse of capacity gap, and pointed out the critical bottleneck it brings to distillation in the era of LLMs.
    \item We have discovered and formally proposed the law of capacity gap, where the expected student scale and its optimal teacher scale follow a linear scaling. It eliminates the compute overhead in distilling LLMs.
    \item The law has led to compute-efficient distillation of LLMs on a larger scale, with a notable improvement on fundamental language understanding tasks as well as instruction-following benchmarks.
\end{itemize}

\section{Background}

\subsection{LM Distillation}

Ever since the birth of transformer-derived LMs~\citep{DBLP:conf/nips/VaswaniSPUJGKP17,DBLP:conf/naacl/DevlinCLT19,DBLP:journals/corr/LiuOG19,DBLP:journals/corr/RadfordNS18,DBLP:journals/jmlr/RaffelSRLNMZLL20}, a surge of interests have been put around compression of LMs~\citep{DBLP:conf/kdd/BucilaCN06}, in which distillation of LMs~\citep{DBLP:journals/corr/HintonVD15,DBLP:journals/corr/SanhDC19,DBLP:conf/emnlp/JiaoYSJCL0L20} is a crucial topic other than quantization~\citep{DBLP:conf/icml/KimGYMK21,DBLP:conf/icml/XiaoLSWDH23,DBLP:journals/corr/DettmersPH23}, pruning~\citep{DBLP:conf/acl/XiaZC22,DBLP:conf/icml/FrantarA23,DBLP:journals/corr/MaFW23}, and dynamic networking (early exiting, moefication, compute elasticity, etc.)~\citep{DBLP:conf/acl/XinTLYL20,DBLP:conf/acl/LiSSYQH20,DBLP:conf/acl/ZhangL00S022,DBLP:conf/naacl/ZuoZLHZC22,DBLP:journals/corr/ZhangWS23}. While initial focus has been allocated to task-specific distillation that distills finetuned LMs with data of finetuning scale~\citep{DBLP:conf/emnlp/SunCGL19}, later focus has shifted to task-agnostic distillation that directly distills pretrained LM with data of pretraining scale~\citep{DBLP:conf/acl/WangBHDW21}, mainly due to its remarkable performance and task flexibility delivered by the task-agnostic property.

Recent LLMs have boosted a new emergence of compression of LLMs~\citep{DBLP:journals/corr/FrantarAH22,DBLP:journals/corr/SunLB23,DBLP:conf/acl/HsiehLYNFRKLP23,DBLP:journals/corr/ZhuLL23}. However, there is a lack of research in LLM distillation, especially in a task-agnostic fashion~\citep{DBLP:conf/icml/FuPOSK23,DBLP:journals/corr/ZhangYWS23,DBLP:journals/corr/JhaGS23,DBLP:journals/corr/GuDWH23,DBLP:journals/corr/AgarwalVS23}. This is at least partly, if not exclusively, owing to the intensive compute demand of task-agnostic distillation of LLMs, which becomes a fatal bottleneck in the presence of the curse of capacity gap.

It should be noted that there are a huge body of studies concentrating on pseudo (alias black-box) distillation~\citep{DBLP:conf/emnlp/KimR16} either at pretraining~\citep{DBLP:journals/corr/GunasekarZA23}, supervised finetuning~\citep{DBLP:conf/acl/HoSY23,DBLP:conf/acl/MagisterMAMS23,DBLP:journals/corr/XuSZ23}, or preference optimization~\citep{DBLP:journals/corr/TunstallBL23} stage. The popularity of pseudo distillation is largely attributed to that many powerful LLMs are proprietary (e.g., GPT4). In this context, the concern of capacity gap might be rather waived along the reduction of teacher knowledge from informative distributed probabilities to one-hot labels.

\subsection{Curse of Capacity Gap}

It was originally recognized that the curse of capacity gap exists in distillation of vision models~\citep{DBLP:conf/aaai/MirzadehFLLMG20}, and recently shown that the curse lies in distillation of LMs~\citep{DBLP:conf/acl/ZhangYLWXWS23} via thorough inspections. The curse of capacity essentially states that a larger teacher LM would not always lead to a better student LM in spite of a stronger performance itself. It also points out that the student is impacted not only by the teacher performance but also by the capacity gap, thus a tradeoff. Consequently, the optimal capacity gap is located somewhere that needs to be discovered as in Figure~\ref{fig:intro_curse}. In addition to the intuitive explanation, a theoretical justification could be found in Theorem 1 in~\citet{DBLP:conf/acl/ZhangYLWXWS23}.

Although previous studies have striven to lift the curse of capacity gap~\citep{DBLP:journals/corr/ZhangYW22,DBLP:conf/acl/ZhouXM22,DBLP:conf/emnlp/YangZS22,DBLP:conf/acl/ZhangYLWXWS23}, the curse can only be partially lifted with regard to the increasingly large scales of teachers, say LLMs. Taking into account the resource demand of LLMs, this work takes a new perspective on the curse, inspired to a certain degree by the law of k-bit quantization~\citep{DBLP:conf/icml/DettmersZ23}. The new view motivates us to unveil the buried law beneath the curse instead of sticking with the efforts on lifting it. Even though the optimal capacity gap has been proven to exist~\cite{DBLP:conf/acl/ZhangYLWXWS23}, it is difficult to derive a mathematical law to determine it. Inspired by the spirits of scaling law in language modeling~\citep{DBLP:journals/corr/KaplanMH20}, we carry out empirical studies to deduct a potential law.

\section{Exploring the Law of Capacity Gap}

Regarding the impossible triangle, we are motivated to explore whether the optimal teacher scale can be determined from the expected student scale with little compute overhead. For this purpose, we presume that there exists a perfectly fitted relation between the expected student scale and the optimal teacher scale, referred to as the \textit{law of capacity gap}:
\begin{proposition} 
Provided a to-be-distilled student of an expected scale, the teacher of an optimal scale can be uniquely determined through a scaling relation.
\end{proposition}

Once verified, the law can remarkably spare computing efforts and resources in deciding in which scale should a preferred teacher be. 

\subsection{Overview}

A pilot study is carried out to investigate whether such a law of capacity gap holds, where we target at conducting small-scale pilot explorations and extrapolating them to large-scale ones. We believe the approximation is acceptable referring to up-to-date experience~\citep{DBLP:journals/corr/XiePD23,DBLP:journals/corr/KaplanMH20} that suggests the validity of the extrapolation. 

Briefly, in each of the comparably small-scale exploration, we attempt to 1) take an LM of a specific scale at a time as a teacher LM $\mathcal{T}$, 2) prune the teacher LM to a specific sparsity as a student LM $\mathcal{S}$, 3) distill from the teacher LM to the student LM, and 4) observe the scaling relation between the optimal teacher LM resulting in the best performance and the corresponding student LM.
\subsection{Method}



\paragraph{Pruning}

In ignorance of architectural differences and details such as biases and normalizations, a causal LM is typically decomposed into a stack of transformer layers, each of which further includes a multi-head self-attention (MHA) block and a feedforward network (FFN) block. We mainly prune the attention heads of MHA blocks, the intermediate neurons of FFN blocks, and the hidden states (thereby embeddings) based on their parameter expressive scores~\citep{DBLP:conf/iclr/MolchanovTKAK17}.

Inspired by structured pruning~\citep{DBLP:conf/nips/MichelLN19}, we attach a set of variables $\xi$, $\nu$ per layer to the attention heads, the intermediate neurons; and a variable $\mu$ shared across layers~\citep{DBLP:conf/acl/XiaZC22} to the hidden states, as shown below:
\begin{equation}
\begin{aligned}
    \text{MHA}(\mathbf{x}_{i})
    =&\sum_{j=1}^{A}\xi_{j}\cdot\text{Attn}(\mu\cdot\mathbf{x}_{\leq i};\mathbf{W}^{\sf Q}_{j},\mathbf{W}^{\sf K}_{j}) \\
    &\cdot\mu\cdot\mathbf{x}_{\leq i}\mathbf{W}^{\sf V}_{j}\mathbf{W}^{\sf O}_{j}, \\
    \text{FFN}(\mathbf{x}_{i})=&\sum_{k=1}^{I}\nu_{k}\cdot\text{g}(\mathbf{x}_{i}\mathbf{W}^{\sf I}_{k})\mathbf{W}^{\sf O}_{k},
\end{aligned}
\end{equation}
where $\mathbf{x}_{i}$ serves as a token hidden state vector, and $\text{g}(\cdot)$ is an activation function. The $j$-th head among $A$ heads is parameterized by $\mathbf{W}^{\sf Q}_{j}$, $\mathbf{W}^{\sf K}_{j}$, $\mathbf{W}^{\sf V}_{j}$, and $\mathbf{W}^{\sf O}_{j}$. The $k$-th neuron among $I$ neurons is parameterized by $\mathbf{W}^{\sf I}_{k}$ and $\mathbf{W}^{\sf O}_{k}$ respectively.

The parameter expressive scores are then calculated through accumulated absolute gradients. A higher expressive score indicates that the corresponding parameter has a higher contribution to the model and hence a lower priority of pruning. The expressive scores of the attention heads, intermediate neurons and hidden states are determined by:
\begin{equation}
\begin{aligned}
    \mathbb{I}^{\sf head}_{j}&=\ \mathbb{E}_{\mathbf{x}_{i}}\left|\frac{\partial\mathcal{L}^{\sf pruning}}{\partial\xi_{j}}\right|, \\
    \mathbb{I}^{\sf neuron}_{j}&=\ \mathbb{E}_{\mathbf{x}_{i}}\left|\frac{\partial\mathcal{L}^{\sf pruning}}{\partial\nu_{j}}\right|,\\
    \mathbb{I}^{\sf hidden}&=\ \mathbb{E}_{\mathbf{x}_{i}}\left|\frac{\partial\mathcal{L}^{\sf pruning}}{\partial\mu}\right|,
\end{aligned}
\end{equation}
\begin{equation}
    \mathcal{L}^{\sf pruning}=\sum_{i}^{|\mathcal{V}|}-\mathbf{y}^{\sf truth}_{i}\text{log}\ \mathbf{y}_{i}^{\sf teacher},    
\end{equation}
where $\mathbb{E}$ represents expectation, $\mathbf{y}$ is the post-softmax probability distribution over the whole vocabulary $\mathcal{V}$. Following~\citet{DBLP:conf/iclr/MolchanovTKAK17}, we apply $\ell_{2}$ normalization to the expressive scores to prevent skewed pruning ratios across layers.

In the pruning procedure, the attention heads, intermediate neurons and hidden states are pruned separately with the same sparsity. For each component, the parameters are ranked according to their expressive scores as calculated above, and the ones with the lowest scores are sequentially pruned until the desired sparsity is reached. Since the sparsity of hidden states and attention heads (or intermediate neurons) have a product effect to the overall sparsity, the individual components are set to a sparsity ratio $p^{\prime}=1-\sqrt{1-p}$ for reaching the target sparsity $p$ of the whole model. It is noteworthy that the squared-root form of hidden state sparsity makes the embeddings be less pruned, but it makes a minimal difference compared with the parameters pruned in MHA and FFN blocks.

\paragraph{Distillation}

We conduct distillation from the teacher to the student with the most basic token-level cross-entropy loss~\citep{DBLP:journals/corr/HintonVD15} as below:
\begin{equation}
\begin{aligned}
    \mathcal{L}^{\sf distillation}&=\frac{1}{2}\sum_{i}^{|\mathcal{V}|}-\mathbf{y}_{i}^{\sf teacher}\text{log}\ \mathbf{y}_{i}^{\sf student} \\
    &+\frac{1}{2}\sum_{i}^{|\mathcal{V}|}-\mathbf{y}^{\sf truth}_{i}\text{log}\ \mathbf{y}_{i}^{\sf student}.
\end{aligned}
\end{equation}

While there are alternative distillation objectives, such as the ones on the sequence level~\citep{DBLP:conf/acl/Wen0DM23}, they are not taken into consideration due to a relatively low efficiency.

\subsection{Setup}

The explorations involve the pruning and distillation of GPT2~\citep{DBLP:journals/corr/RadfordWC19} and Pythia~\citep{DBLP:conf/icml/BidermanSABOHKP23} with OpenWebText~\citep{GokaslanC19}. Each of these LM series comprises LMs of different scales, i.e., GPT2-base (140M), -medium (380M), -large (810M), -xlarge (1.6B); and Pythia-70M, -160M, -410M, -1B, -1.4B, -2.8B. Each LM is distilled into student LMs of varying sizes through sparsity $p$. The perplexity on the WikiText2 and last word prediction accuracy of LAMBADA are computed as the performance indicators of student LMs. 
OpenWebText consists of $\sim$40G web data that amounts to 3.9B tokens, which are in a similar data scale to that of GPT2 pretraining. 
The data scale variation is also considered by reducing the number of training tokens. We use a subset of the data ($\sim$5\%) for pruning here and hereafter for a high pruning efficiency.


\begin{figure*}[ht]
    \centering
    \begin{subfigure}[b]{0.445\textwidth}
        \centering
        \includegraphics[width=\textwidth]{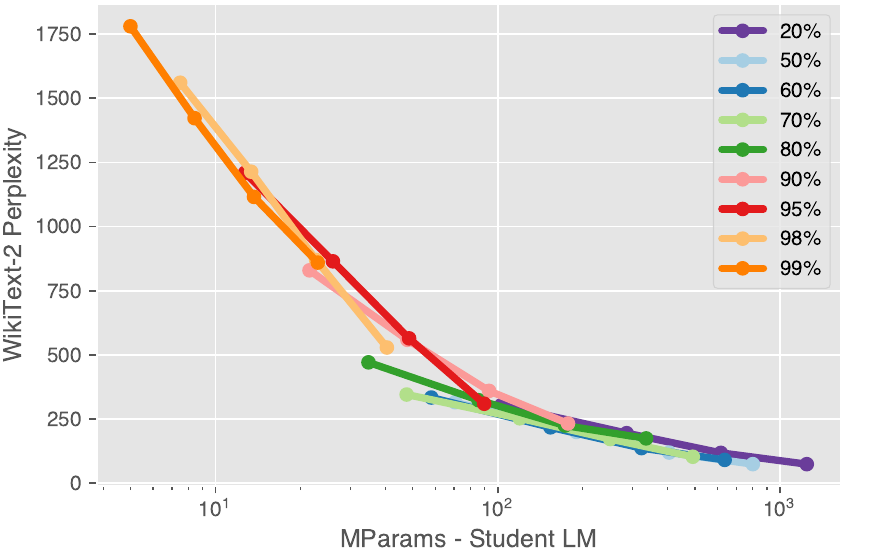}
        \caption{GPT2 on WikiText2.}
        \label{fig:obs_gpt2_wikitext2}
    \end{subfigure}%
    \hfill
    \begin{subfigure}[b]{0.437\textwidth}
        \centering
        \includegraphics[width=\textwidth]{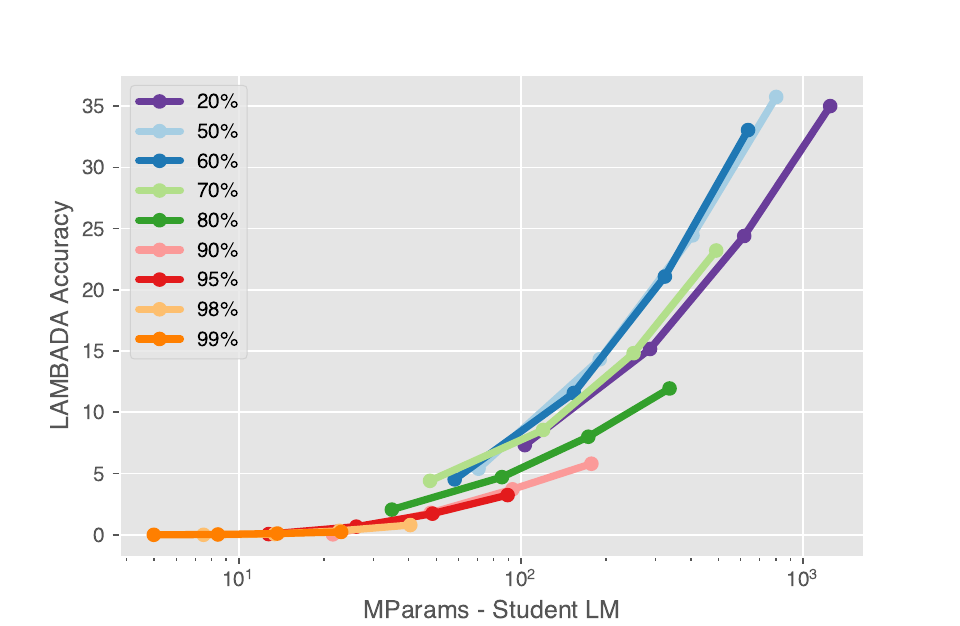}
        \caption{GPT2 on LAMBADA.}
        \label{fig:obs_gpt2_lambada}
    \end{subfigure}
    \begin{subfigure}[b]{0.439\textwidth}
        \centering
        \includegraphics[width=\textwidth]{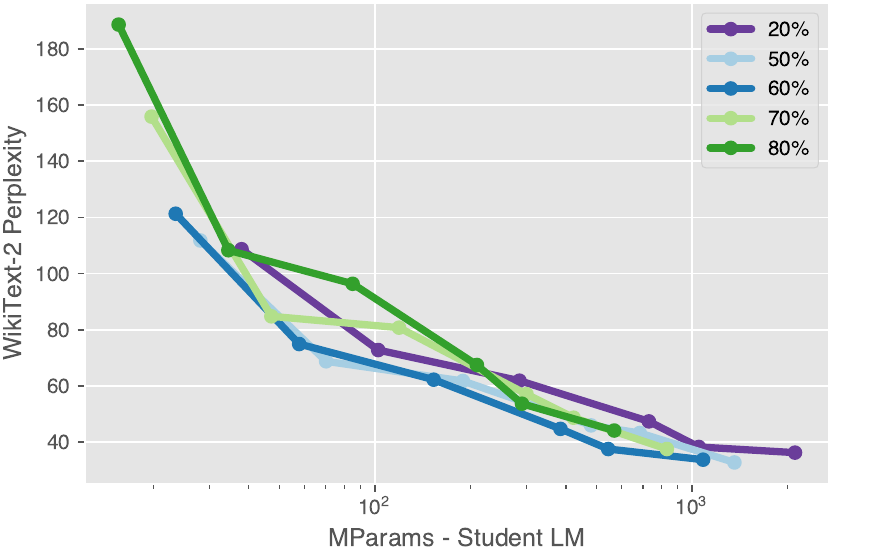}
        \caption{Pythia on WikiText2.}
        \label{fig:obs_pythia_wikitext2}
    \end{subfigure}
    \hfill
    \begin{subfigure}[b]{0.437\textwidth}
        \centering
        \includegraphics[width=\textwidth]{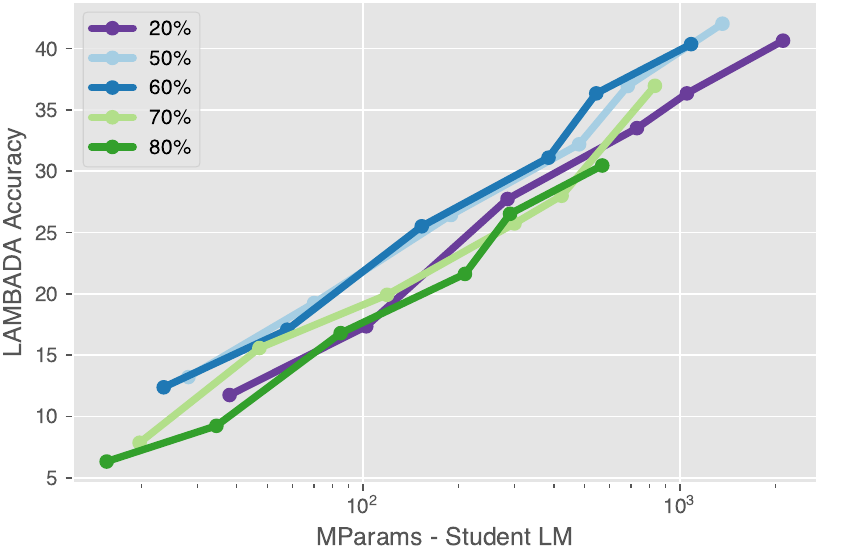}
        \caption{Pythia on LAMBADA.}
        \label{fig:obs_pythia_lambada}
    \end{subfigure}
    \begin{subfigure}[b]{0.439\textwidth}
        \centering
        \includegraphics[width=\textwidth]{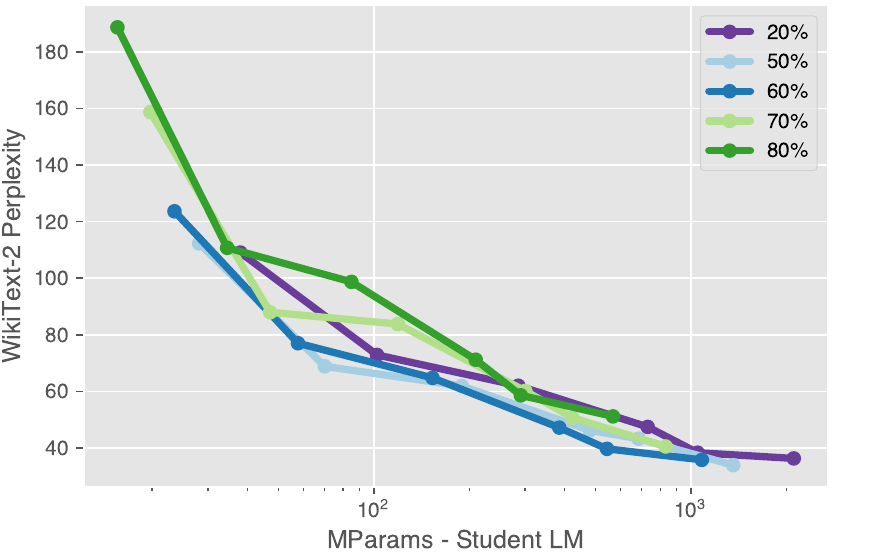}
        \caption{Pythia on WikiText2 with fewer tokens.}
        \label{fig:obs_pythia_wikitext2_token}
    \end{subfigure}
    \hfill
    \begin{subfigure}[b]{0.437\textwidth}
        \centering
        \includegraphics[width=\textwidth]{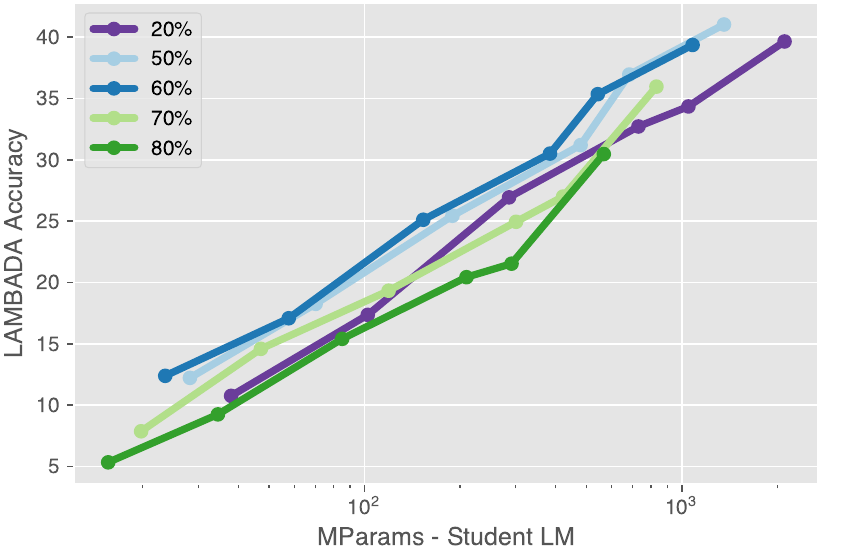}
        \caption{Pythia on LAMBADA with fewer tokens.}
        \label{fig:obs_pythia_lambada_token}
    \end{subfigure}
    \caption{The observations from the distillation of GPT2 and Pythia series. Students are evaluated on the test set of WikiText2~\citep{DBLP:conf/iclr/MerityX0S17} in perplexity and the test set of LAMBADA~\citep{DBLP:conf/acl/PapernoKLPBPBBF16} in last word prediction accuracy. A line of a color represents a sparsity, where each point in it represents a student pruned and distilled from a teacher at such sparsity. }
    \label{fig:obs}
\end{figure*}

\subsection{Observations}

The explorations produce results of distillation teacher-student pairs with different capacity gaps so that we can form the scaling plots as in Figure~\ref{fig:obs}. In each plot, a colored line corresponds to a certain sparsity, and points with the same horizontal order in all lines correspond to the same teacher LM. For exploring the law of capacity gap, the following hypotheses should be examined:

\quad \textbf{H1: the capacity gap influences the student performance.}

\quad \textbf{H2: the optimal teacher scale exists in finite spaces, given a student scale.}

\quad \textbf{H3: the optimal teacher scale does not vary across settings, given a student scale.}

It can be observed from the plots that a) for a fixed-scale teacher LM, the performance of student always improves or stabilizes along the increment of its scale; b) when the student scale is fixed, its performance invariably follows an increasing-steady-decreasing pattern as the gradual increase in the teacher scale, implying the existence of an optimal teacher. The findings support \textbf{H1} and \textbf{H2}, respectively.

Furthermore, we discover that the optimal teacher scale for a certain student scale is consistent across distinct student scales and data scales, at an approximate sparsity of $\sim$60\%. This unearths that \textbf{H3} also holds true.

\begin{figure}[ht]
    \centering
    \includegraphics[width=0.47\textwidth]{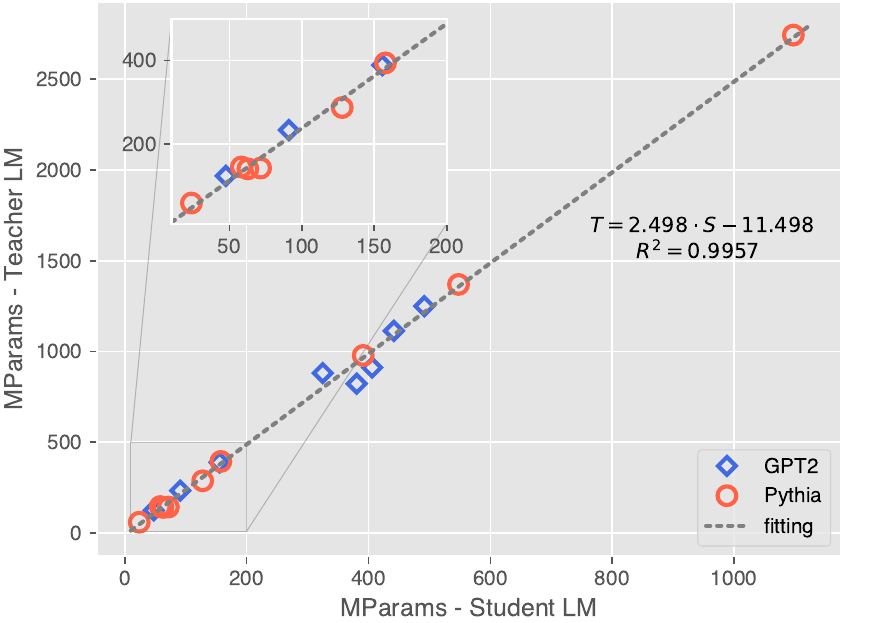}
    \caption{The curse of capacity gap can be leaned towards the law of capacity gap. 
    \textit{law}, the optimal teacher scale exists and remains linear to the student scale. Each point stands for the best teacher scale for a fixed student scale.}
    \label{fig:intro_curse_to_law}
\end{figure}

\subsection{Characterization}

The validity of the three hypotheses suggests the existence of the law of capacity gap, i.e., a consistent relation between the scale of target student and the optimal teacher scale. We attempt to move a step further by characterizing the law. 

For this purpose, we collect the optimal teacher LMs for different student LMs from the explorations and draw a scatter plot for the obtained (student scale, teacher scale) pairs, as shown in Figure~\ref{fig:intro_curse_to_law}. It can be seen that the points appear to locate on a straight line. Therefore, we apply fitting to the points as below:
\begin{equation}
    \mathbb{T}^{\star}\approx \alpha\cdot\mathbb{S}+\beta
\end{equation}

The line of best fit is obtained with $\alpha=2.498$ and $\beta=-11.498$, with $R^2=0.9957$ indicating a perfect fit. This means that the curse of capacity gap can be appropriately viewed as a law:

\begin{law}
    Given a to-be-distilled student $\mathcal{S}$ of an expected scale $\mathbb{S}$, a teacher $\mathcal{T}^{\star}$ of an optimal scale $\mathbb{T}$ should be: 
    \begin{equation}
        \mathbb{T}^{\star}\approx2.498\cdot\mathbb{S}-11.498\approx 2.5\cdot\mathbb{S}
    \end{equation}
\end{law}

And the law promptly triggers a follow-up corollary:
\begin{corollary}
    On condition that a to-be-distilled student $\mathcal{S}$ is of a scale $0.4\cdot\mathbb{T}^{\star}$, a teacher $\mathcal{T}$ of a scale $\mathbb{T}^{\star}$ should be the optimal teacher for $\mathcal{S}$.
\end{corollary}

\begin{remark}
The law of capacity gap is drawn under specific design choices. Nevertheless, we do consider the law under different teacher architectures (e.g., GPT2 in Figure~\ref{fig:obs_gpt2_wikitext2} versus Pythia in Figure~\ref{fig:obs_pythia_wikitext2}), data scales (e.g., full tokens in Figure~\ref{fig:obs_pythia_lambada} versus fewer tokens in Figure~\ref{fig:obs_pythia_lambada_token}), pruning paradigms (c.f., Section~\ref{sec:ablation}), and distillation objectives (c.f., Section~\ref{sec:ablation}). Empirical evidence suggests that the basic law universally applies to all these circumstances.
\end{remark}



There is a trend where small language models (SLMs) are distilled rather than trained from scratch~\citep{DBLP:journals/corr/GunerWW24,DBLP:journals/corr/RivierePS24}. Interestingly, almost all these distilled SLMs use teacher models whose scales roughly follow our linear law. 

\section{Extrapolating the Law of Capacity Gap}

We have drawn the law of capacity gap from small-scale explorations. Yet, whether the law holds for LLMs on a large scale remains unknown. To answer this question, we carry out a series of experiments on extrapolating the law to larger-scale experiments to exhibit the extrapolating ability of the law.


\subsection{Setup}

\begin{table}[ht]
    \centering
    \caption{The statistics of data mixture used for distillation.}
    \begin{adjustbox}{width=0.39\textwidth,center}
    \begin{tabular}{lrrr}
    \toprule
        \textbf{Dataset} & \textbf{Epochs} & \textbf{Tokens} & \textbf{Proportion} \\
    \midrule
        Pile & 0.16 & 57B & 45.2\% \\
        GitHub & 0.50 & 30B & 23.8\% \\
        WuDao & 0.78 & 39B & 31.0\% \\
    \midrule
        Mixture & 1.00 & 126B & 100.0\% \\
    \bottomrule
    \end{tabular}
    \end{adjustbox}
    \label{tab:minima_data_stat}
\end{table}

\paragraph{Target}

We distill popular LLaMA2-7B~\citep{DBLP:journals/corr/TouvronMS23} and LLaMA3.1-8B~\citep{DBLP:journals/corr/DubeyJP24} to 0.4$\times$7$\approx$3B LMs named \textsc{MiniMA}. Considering potential applications to Chinese tasks, we do not directly leverage LLaMA2-7B released by Meta but use an adapted one. The adapted LLaMA2-7B is expanded with a Chinese vocabulary, incrementally trained on a mixture of subsampled Pile~\citep{DBLP:journals/corr/GaoBB21}, GitHub~\citep{RedPajama23}, and WuDao~\citep{DBLP:journals/aiopen/YuanZDDLCZYT21}, as detailed in Appendix~\ref{app:llama_adaptation}. In contrast, LLaMA3.1-8B is already armed with a Chinese-enhanced vocabulary. The mixed data is also utilized to conduct pruning and distillation, and the statistics are in Table~\ref{tab:minima_data_stat}.

Taking a step further, we finetune \textsc{MiniMA} on instruction-following data and get \textsc{MiniChat}. The instruction-following data covers a broad collection of both single-turn question-answering pairs and multi-turn conversation sessions, which totally adds up to 1.1M examples. The statistics are detailed in Appendix~\ref{app:stat_inst_data} due to the space limitation.

\paragraph{Implementation}

Different from the pruning in pilot explorations, we additionally impose a heuristic rule to the pruning priority of \textsc{MiniMA} so that it can produce symmetric shapes across layers and make \textsc{MiniMA} universally applicable without any monkey-patches in succeeding use. The heuristic pruning priority first converts the previously used global priority to a local layer-wise one, then compensates for the loss of the entire-layer pruning ability by heuristically dropping off the 4 bottommost and 4 topmost layers, and accordingly pruning few hidden states (as well as embeddings plus language model head), MHA heads and FFN neurons. The reasons behind such design choice are detailed in Appendix~\ref{app:minima_pruning}.

The distillation takes in data packed as sequences of 4,096 tokens. The batch size is set to 1,024 ($\sim$4M tokens). The learning rate is 3e-4, and the weight decay is 1e-1. The training lasts for only 1 epoch. The learning rate is scheduled to warm up linearly for the first 1\% steps of all and decay down in a sinusoidal way for the rest steps. Gradients whose norms accumulate over 1.0 are necessarily clipped. The training is executed on 16 A100 80G GPUs. The training efficiency is guaranteed by DeepSpeed Zero2~\citep{DBLP:conf/kdd/RasleyRRH20} and FlashAttention~\citep{DBLP:conf/nips/DaoFERR22}. Gradient checkpointing is enabled to further reduce the memory footprint and bfloat16 precision is chosen to improve the training stability.

\begin{table*}[ht]
    \centering
    \caption{The results of \textsc{MiniMA} on standard benchmarks. The best results are \textbf{boldfaced}. The results marked with \textsuperscript{\dag} are made compute-comparable to \textsc{MiniMA}. The results marked with \textsuperscript{\ddag} are produced with a larger corpus due to a larger vocabulary.}
    \begin{adjustbox}{width=0.97\textwidth,center}
    \begin{tabular}{lrrcccccc}
    \toprule
        \textbf{LM} & \textbf{Params} & \textbf{Tokens} & \makecell[c]{\textbf{MMLU}\\\textbf{Acc}} & \makecell[c]{\textbf{CEval}\\\textbf{Acc}} & \makecell[c]{\textbf{DROP}\\\textbf{EM Score}} & \makecell[c]{\textbf{BBH}\\\textbf{EM Score}} & \makecell[c]{\textbf{GSM8K}\\\textbf{Maj1@1}} & \makecell[c]{\textbf{HumanEval}\\\textbf{Pass@1}} \\
    \midrule
        LLaMA2 & 7 B & 2 T & 46.00 & 34.40 & 31.57 & 32.02 & 14.10 & 12.80 \\
        \quad - ShortGPT & 3 B & 0 B & 25.57 & 26.79 & 8.72 & 7.53 & 4.52 & 0.00 \\
        \quad - LayerChop & 3 B & 126 B & 25.76 & 27.23 & 10.45 & 12.23 & 6.97 & 3.66 \\
        \quad - ShearedLLaMA & 3 B & 126 B & 25.39 & 27.79 & 20.41 & 30.21 & 5.08 & 8.54 \\
        \quad - ShearedLLaMA\textsuperscript{\dag} & 3 B & 252 B & 26.15 & 27.81& 20.03 & 30.11 & 5.26 & 9.01 \\
        \quad - \textsc{MiniMA} & 3 B & 126 B & \textbf{28.51} & \textbf{28.23} & \textbf{22.50} & \textbf{31.61} & \textbf{8.11} & \textbf{10.98} \\
    \midrule
        LLaMA2~\citeyearpar{DBLP:journals/corr/TouvronMS23}  & 13 B & 2 T & 55.34 & 41.60 & 43.40 & 38.07 & 24.11 & 14.02 \\
        \quad - \textsc{MiniMA} & 3 B & 126 B & 26.82 & 28.23 & 18.92 & 30.01 & 5.31 & 10.37 \\
    \midrule
        LLaMA2~\citeyearpar{DBLP:journals/corr/TouvronMS23}  & 70 B & 2 T & 68.62 & 53.86 & 63.31 & 51.58 & 53.37 & 28.66 \\
        \quad - \textsc{MiniMA} & 3 B & 126 B & 26.91 & 24.89 & 19.03 & 28.55 & 3.56 & 7.32 \\
    \midrule
        LLaMA3.1~\citeyearpar{DBLP:journals/corr/DubeyJP24}  & 8 B & 15 T & 64.75 & 52.45 & 49.02 & 41.00 & 48.82 & 34.15 \\
        \quad - \textsc{MiniMA}\textsuperscript{\ddag} & 3 B & 126 B & 31.72 & 29.89 & 30.46 & 32.06 & 10.42 & 18.46 \\
    \bottomrule
    \end{tabular}
    \end{adjustbox}
    \label{tab:minima_standard_bench}
\end{table*}

\begin{figure}[t]
    \centering
    \includegraphics[width=0.475\textwidth]{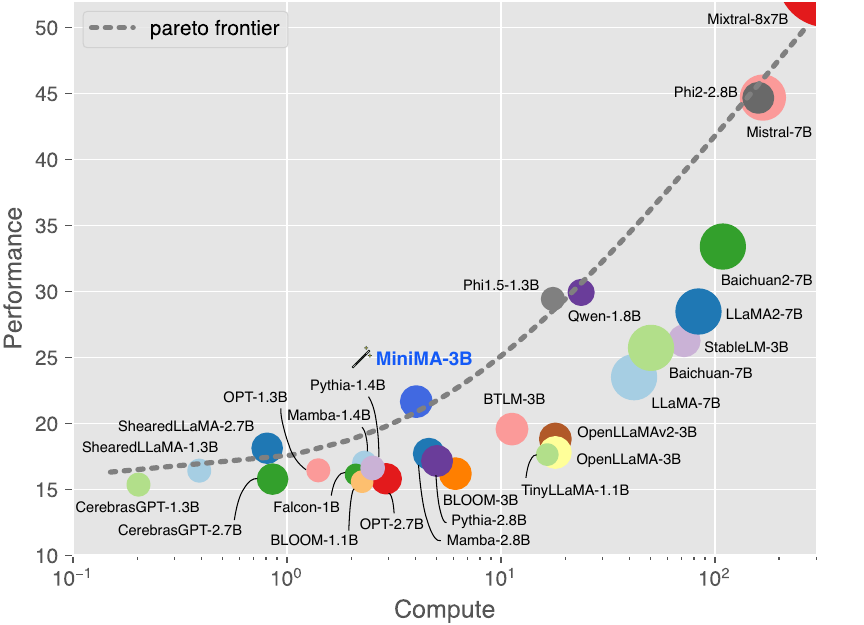}
    \caption{The new compute-performance pareto frontier is yielded by \textsc{MiniMA}, namely \textsc{MiniMA} is more compute-efficient given any compute budget than existing LMs. The radius of each circle stands for the model scale. \textit{performance}: average task measure as each detailed in Appendix~\ref{app:bench_lm}. \textit{compute}: estimated training compute in $\times$10\textsuperscript{9} TFLOPs as detailed in Appendix~\ref{app:est_compute}.}
    \label{fig:intro_pareto_frontier}
\end{figure}

The finetuning receives data padded as sequences of 4,096 tokens. The batch size is set to 256 ($\sim$1M tokens). The learning rate is 2e-5, and the weight decay is 1e-1. The training lasts for 3 epochs with early stopping after continuously 2 epochs without validation improvement. The learning rate is scheduled to warm up linearly for the first 10\% steps of all and decay down in a sinusoidal way for the rest steps. Gradients whose norms accumulate over 1 are necessarily clipped. The training is executed on 8 A100 80G GPUs. The training efficiency is again guaranteed likewise by those in the distillation.

\subsection{Main Results}

\paragraph{Standard Benchmarks}

We evaluate \textsc{MiniMA} against an extensive set of baselines on several standard benchmarks including MMLU~\citep{DBLP:conf/iclr/HendrycksBBZMSS21}, CEval~\citep{DBLP:journals/corr/HuangBZ23}, DROP~\citep{DBLP:conf/naacl/DuaWDSS019}, BBH~\citep{DBLP:conf/acl/SuzgunSSGTCCLCZ23}, GSM8K~\citep{DBLP:journals/corr/CobbeKB21}, and HumanEval~\citep{DBLP:journals/corr/ChenTJ21}. The introductions and evaluation protocols of these datasets are detailed in Appendix~\ref{app:bench_eval_proto}. We compare \textsc{MiniMA} against a few models involving efficient designs, such as layer dropping: ShortGPT~\citep{DBLP:journals/corr/MenXZ24}, LayerChop~\citep{DBLP:journals/corr/JhaGS23}, and pruning: ShearedLLaMA~\citep{DBLP:journals/corr/XiaGZC23}.

The results in Table~\ref{tab:minima_standard_bench} demonstrate that LLaMA2-7B is exactly the best among considered LLaMA2-\{7,13,70\}B. And the performance of \textsc{MiniMA} consistently degrades along the increased scales of teacher LMs. This phenomenon implies the practical value of the law. Moreover, \textsc{MiniMA} achieves superior performance over baselines with respect to knowledge (MMLU, CEval), reasoning (DROP, BBH, GSM8K), and coding (HumanEval). Even with similar compute consumed, \textsc{MiniMA} yet outperforms ShearedLLaMA. This indicates that distillation, especially when promoted by the law, is competitive among other efficient designs. Further, \textsc{MiniMA} from LLaMA3.1 is better than that from LLaMA2, showcasing the prominence of \textsc{MiniMA} via adopting more promising teacher LMs. The training loss-level comparisons among them are also detailed in Appendix~\ref{app:minima_train_loss} for potential interests.

From a broader comparison that involves existing LMs of various scales, as in Figure~\ref{fig:intro_pareto_frontier}, we could clearly see that \textsc{MiniMA} basically builds an excitingly new compute-performance Pareto frontier therein.

\paragraph{GPT4 Assessments}

\begin{table}[t]
    \centering
    \caption{The results of \textsc{MiniChat} on GPT4 assessments. Macro average scores are reported across fields in these two datasets. The better results are \textbf{boldfaced}.}
    \begin{adjustbox}{width=0.5\textwidth,center}
    \begin{tabular}{lcc}
    \toprule
        \textbf{LM Pair} & \makecell[c]{\textbf{Vicuna-Bench}\\\textbf{Macro Avg}} & \makecell[c]{\textbf{BELLE-Bench}\\\textbf{Macro Avg}} \\
    \midrule
        \textsc{MiniChat} \\
        v.s. OpenBuddy-3B & \textbf{7.64}: 5.42 & \textbf{7.77}: 6.81 \\
        v.s. BiLLa-7B & 7.24: \textbf{7.41} & \textbf{7.73}: 7.49 \\
        v.s. ChatGLM-6B & \textbf{7.63}: 5.63 & \textbf{7.44}: 7.23 \\
        v.s. Phoenix-7B & \textbf{7.35}: 6.95 & \textbf{7.62}: 7.21 \\
        v.s. ChatGLM2-6B & \textbf{7.35}: 7.30 & 7.40: \textbf{8.00} \\
    \bottomrule
    \end{tabular}
    \end{adjustbox}
    \label{tab:minichat_gpt4_assess}
\end{table}

We compare \textsc{MiniChat} to some instruction-following LMs on Vicuna-Bench~\citep{Vicuna23} and BELLE-bench~\citep{DBLP:journals/corr/JiGD23}, which demand GPT4 for pairwise assessments. The introductions and evaluation protocols of these datasets are detailed in Appendix~\ref{app:gpt4_eval_proto}. The regarded instruction-following baselines encompass scale-matched ones: OpenBuddy-3B~\citep{OpenBuddy23}, and scale-mismatched ones: BiLLa-7B~\citep{BiLLA23}, ChatGLM-6B~\citep{ChatGLM23}, Phoenix-7B~\citep{DBLP:journals/corr/ChenJC23}, ChatGLM2-6B~\citep{ChatGLM223}.

The results in Table~\ref{tab:minichat_gpt4_assess} disclose that \textsc{MiniChat} surpasses scale-matched 3B baseline OpenBuddy-3B by a landslide, and even keeps competitive with scale-mismatched baselines. In detail, \textsc{MiniChat} only falls behind BiLLa-7B on Vicuna-Bench and ChatGLM2-6B on BELLE-Bench. The results of \textsc{MiniChat} on standard benchmarks are detailed in Appendix~\ref{app:minichat_bench} in case of curiosity. The field-specific scores of mentioned LMs are provided in Figure~\ref{fig:intro_ability_radar_vicuna} in the form of ability radar, and further detailed in Appendix~\ref{app:gpt4_eval_field}. The micro evaluation displays the reasonable ability realized by \textsc{MiniChat}. The consistency of GPT4 evaluation to human evaluation is detailed in Appendix~\ref{app:human_elo}.

\begin{figure}[ht]
    \centering
    \includegraphics[width=0.5\textwidth]{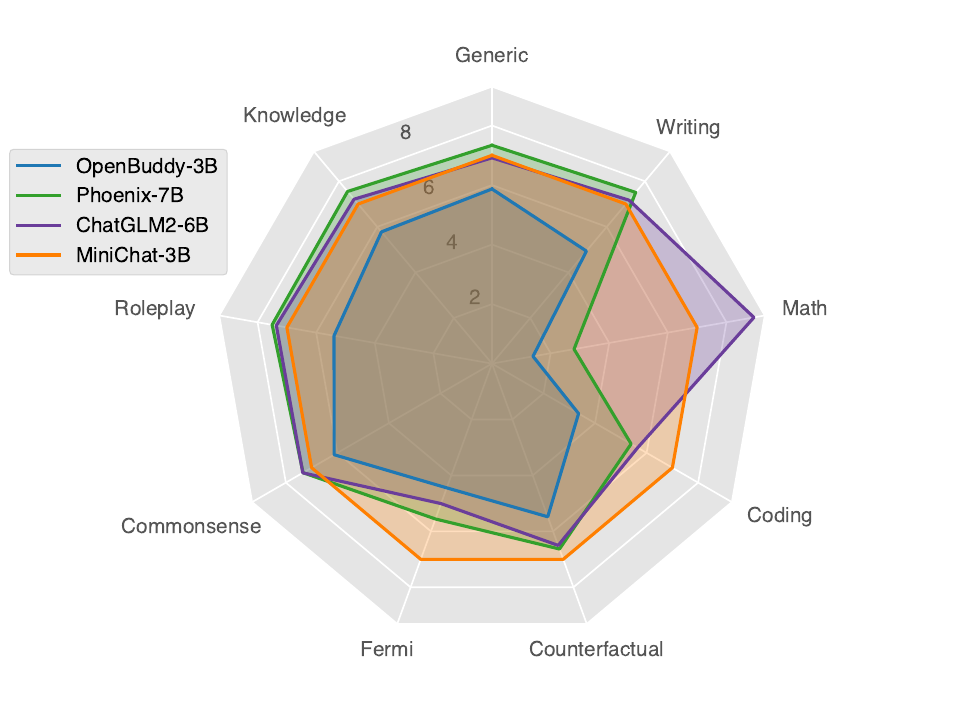}
    \caption{The reasonable ability radar is realized by \textsc{MiniChat}. The head-to-head comparison scores on Vicuna-Bench of each baseline are rebased according to \textsc{MiniChat}.}
    \label{fig:intro_ability_radar_vicuna}
\end{figure}

\begin{figure}[t]
    \centering
    \includegraphics[width=0.47\textwidth]{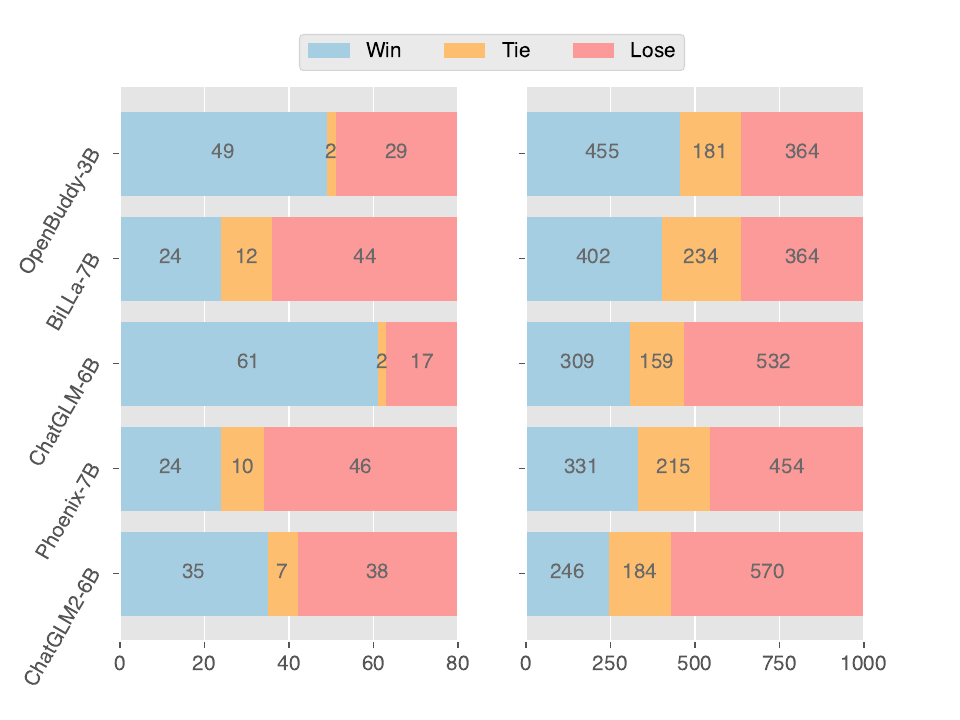}
    \captionof{figure}{The win-lose-tie portions of \textsc{MiniChat} versus its competitors. The left is reported on the Vicuna-Bench while the right is reported on the BELLE-Bench.}
    \label{fig:minichat_gpt_assess}
\end{figure}

Albeit this, the winners in macro average scores are not necessarily the winners in win-lose-tie battles as detailed in Figure~\ref{fig:minichat_gpt_assess}. For example, \textsc{MiniChat} is way better than Phoenix on Vicuna-Bench in terms of Macro Average yet lags behind Phoenix in terms of win rate. In summary, the excellence of \textsc{MiniChat} encourages us to further explore the application of \textsc{MiniMA}.

\subsection{Ablation Studies}
\label{sec:ablation}

\begin{table*}[t]
    \centering
    \caption{The results of ablation studies.}
    \begin{adjustbox}{width=0.87\textwidth,center}
    \begin{tabular}{lcccccc}
    \toprule
        \textbf{LM} & \makecell[c]{\textbf{MMLU}\\\textbf{Acc}} & \makecell[c]{\textbf{CEval}\\\textbf{Acc}} & \makecell[c]{\textbf{DROP}\\\textbf{EM Score}} & \makecell[c]{\textbf{BBH}\\\textbf{EM Score}} & \makecell[c]{\textbf{GSM8K}\\\textbf{Maj1@1}} & \makecell[c]{\textbf{HumanEval}\\\textbf{Pass@1}} \\
    \midrule
        \textsc{MiniMA} & 28.51 & 28.23 & 22.50 & 31.61 & 8.11 & 10.98 \\
        \quad- global pruning & 28.92 & 28.52 & 19.55 & 32.89 & 7.26 & 11.37 \\
        \quad- layer dropping & 27.59 & 26.95 & 21.96 & 30.37 & 6.85 & 9.86 \\
        \quad- patient knowledge & 28.69 & 28.37 & 22.51 & 31.89 & 8.00 & 10.64 \\
        \quad- teacher assistant & 27.69 & 27.49 & 21.31 & 30.02 & 7.72 & 9.58 \\
    \bottomrule
    \end{tabular}
    \end{adjustbox}
    \label{tab:minima_ablation_study}
\end{table*}

To further validate the law, we compare \textsc{MiniMA} to its variants with different pruning strategies: global pruning (i.e., asymmetric shaping), layer dropping~\citep[i.e., depth pruning,][]{DBLP:journals/corr/MenXZ24}, and different distillation techniques: patient knowledge~\citep[i.e., taking last-layer hidden states as knowledge,][]{DBLP:conf/emnlp/SunCGL19}, teacher assistant~\citep[i.e., inserting a 5B teacher assistant,][]{DBLP:conf/aaai/MirzadehFLLMG20}.

The results in Table~\ref{tab:minima_ablation_study} evidence that the specializations on the pruning priority of \textsc{MiniMA} would not harm (sometimes yet benefit to) the performance, when compared to either global pruning or layer dropping. This hints the rationality of symmetric shaping as described in hands-on implementation, and the adequacy of the employed pruning paradigm in drawing the law. Besides, the naive knowledge distillation~\citep{DBLP:journals/corr/HintonVD15} is averagely comparable to advanced tactics like patient knowledge distillation, coinciding with~\citet[][Table 19]{DBLP:journals/corr/MuralidharanSJ24}. This features the distillation objective in use would be relatively minor concerning the law. A counter-argument would be that the teacher assistant can not improve over ours, since the distillation path is sub-optimal. Nevertheless, no one can assure the distillation from LLaMA2-70B still follows this truth. As a whole, the above phenomena reflect the practical solidity of the law.

\section{Conclusions and Future Work}

In this paper, we propose a novel viewpoint towards the capacity gap in distilling LMs in the era of LLMs to counter the impossible triangle of capacity gap deviated from the curse of capacity gap. Via this port, we beneficially turn the curse of capacity gap into a law through pilot explorations. Activated by the law, we put forward extrapolation experiments resulting into valuable SLMs, establishing a state-of-the-art compute-performance Pareto frontier. We have also managed to draw a comprehensive panorama of MiniMA family by landing related models as in Appendix~\ref{app:minima_family}.

\section*{Limitations}

There are a few obvious limitations in this work: 1) we do not carry out very complete experiments for \textsc{MiniMA} while LLaMA3 grants a larger model at even 400B scale, 2) we do not consider data scale, data mixture, and distillation techniques as core scaling factors, and 3) we do not combine a preference optimization stage for \textsc{MiniChat} while approaches like DPO~\citep{DBLP:journals/corr/RafailovSM23} can principally alleviate the safety concerns and lift the user experiences. Hopefully, we could address these in future work.

\section*{Ethics Statements}

\textsc{MiniChat} is subject to the same well-recognized limitations of other LLMs, including potential for non-factual generation such as unqualified advice, and a propensity towards hallucinations. And like other LLMs, \textsc{MiniChat} may generate harmful, offensive, or biased content due to its training on publicly available online datasets. Not everyone who uses LLMs has good intentions, and \textsc{MiniChat} could potentially be used for nefarious purposes such as generating misinformation.

\section*{Acknowledgments}

This work is funded in part by the Natural Science Foundation of China (grant no: 62376027).


\bibliography{custom}

\begin{thebibliography}{116}
\providecommand{\natexlab}[1]{#1}

\bibitem[{Agarwal et~al.(2023)Agarwal, Vieillard, Stanczyk, Ramos, Geist, and Bachem}]{DBLP:journals/corr/AgarwalVS23}
Rishabh Agarwal, Nino Vieillard, Piotr Stanczyk, Sabela Ramos, Matthieu Geist, and Olivier Bachem. 2023.
\newblock \href {https://doi.org/10.48550/ARXIV.2306.13649} {{GKD:} generalized knowledge distillation for auto-regressive sequence models}.
\newblock \emph{CoRR}, abs/2306.13649.

\bibitem[{Bahdanau(2022)}]{FLOPs22}
Dzmitry Bahdanau. 2022.
\newblock \href {https://medium.com/@dzmitrybahdanau/the-flops-calculus-of-language-model-training-3b19c1f025e4} {The flops calculus of language model training}.

\bibitem[{Bai et~al.(2023)Bai, Bai, Chu, Cui, Dang, Deng, Fan, Ge, Han, Huang, Hui, Ji, Li, Lin, Lin, Liu, Liu, Lu, Lu, Ma, Men, Ren, Ren, Tan, Tan, Tu, Wang, Wang, Wang, Wu, Xu, Xu, Yang, Yang, Yang, Yang, Yao, Yu, Yuan, Yuan, Zhang, Zhang, Zhang, Zhang, Zhou, Zhou, Zhou, and Zhu}]{DBLP:journals/corr/BaiBC23}
Jinze Bai, Shuai Bai, Yunfei Chu, Zeyu Cui, Kai Dang, Xiaodong Deng, Yang Fan, Wenbin Ge, Yu~Han, Fei Huang, Binyuan Hui, Luo Ji, Mei Li, Junyang Lin, Runji Lin, Dayiheng Liu, Gao Liu, Chengqiang Lu, Keming Lu, Jianxin Ma, Rui Men, Xingzhang Ren, Xuancheng Ren, Chuanqi Tan, Sinan Tan, Jianhong Tu, Peng Wang, Shijie Wang, Wei Wang, Shengguang Wu, Benfeng Xu, Jin Xu, An~Yang, Hao Yang, Jian Yang, Shusheng Yang, Yang Yao, Bowen Yu, Hongyi Yuan, Zheng Yuan, Jianwei Zhang, Xingxuan Zhang, Yichang Zhang, Zhenru Zhang, Chang Zhou, Jingren Zhou, Xiaohuan Zhou, and Tianhang Zhu. 2023.
\newblock \href {https://doi.org/10.48550/ARXIV.2309.16609} {Qwen technical report}.
\newblock \emph{CoRR}, abs/2309.16609.

\bibitem[{Baichuan(2023)}]{Baichuan123}
Baichuan. 2023.
\newblock \href {https://github.com/baichuan-inc/Baichuan-7B} {A series of large language models developed by baichuan intelligent technology}.

\bibitem[{Biderman et~al.(2023)Biderman, Schoelkopf, Anthony, Bradley, O'Brien, Hallahan, Khan, Purohit, Prashanth, Raff, Skowron, Sutawika, and van~der Wal}]{DBLP:conf/icml/BidermanSABOHKP23}
Stella Biderman, Hailey Schoelkopf, Quentin~Gregory Anthony, Herbie Bradley, Kyle O'Brien, Eric Hallahan, Mohammad~Aflah Khan, Shivanshu Purohit, USVSN~Sai Prashanth, Edward Raff, Aviya Skowron, Lintang Sutawika, and Oskar van~der Wal. 2023.
\newblock \href {https://proceedings.mlr.press/v202/biderman23a.html} {Pythia: {A} suite for analyzing large language models across training and scaling}.
\newblock In \emph{International Conference on Machine Learning, {ICML} 2023, 23-29 July 2023, Honolulu, Hawaii, {USA}}, volume 202 of \emph{Proceedings of Machine Learning Research}, pages 2397--2430. {PMLR}.

\bibitem[{Brown et~al.(2020)Brown, Mann, Ryder, Subbiah, Kaplan, Dhariwal, Neelakantan, Shyam, Sastry, Askell, Agarwal, Herbert{-}Voss, Krueger, Henighan, Child, Ramesh, Ziegler, Wu, Winter, Hesse, Chen, Sigler, Litwin, Gray, Chess, Clark, Berner, McCandlish, Radford, Sutskever, and Amodei}]{DBLP:conf/nips/BrownMRSKDNSSAA20}
Tom~B. Brown, Benjamin Mann, Nick Ryder, Melanie Subbiah, Jared Kaplan, Prafulla Dhariwal, Arvind Neelakantan, Pranav Shyam, Girish Sastry, Amanda Askell, Sandhini Agarwal, Ariel Herbert{-}Voss, Gretchen Krueger, Tom Henighan, Rewon Child, Aditya Ramesh, Daniel~M. Ziegler, Jeffrey Wu, Clemens Winter, Christopher Hesse, Mark Chen, Eric Sigler, Mateusz Litwin, Scott Gray, Benjamin Chess, Jack Clark, Christopher Berner, Sam McCandlish, Alec Radford, Ilya Sutskever, and Dario Amodei. 2020.
\newblock \href {https://proceedings.neurips.cc/paper/2020/hash/1457c0d6bfcb4967418bfb8ac142f64a-Abstract.html} {Language models are few-shot learners}.
\newblock In \emph{Advances in Neural Information Processing Systems 33: Annual Conference on Neural Information Processing Systems 2020, NeurIPS 2020, December 6-12, 2020, virtual}.

\bibitem[{Bucila et~al.(2006)Bucila, Caruana, and Niculescu{-}Mizil}]{DBLP:conf/kdd/BucilaCN06}
Cristian Bucila, Rich Caruana, and Alexandru Niculescu{-}Mizil. 2006.
\newblock \href {https://doi.org/10.1145/1150402.1150464} {Model compression}.
\newblock In \emph{Proceedings of the Twelfth {ACM} {SIGKDD} International Conference on Knowledge Discovery and Data Mining, Philadelphia, PA, USA, August 20-23, 2006}, pages 535--541. {ACM}.

\bibitem[{Chaudhary(2023)}]{CodeAlpaca23}
Sahil Chaudhary. 2023.
\newblock \href {https://github.com/sahil280114/codealpaca} {Code alpaca: An instruction-following llama model for code generation}.

\bibitem[{Chen et~al.(2021)Chen, Tworek, Jun, Yuan, de~Oliveira~Pinto, Kaplan, Edwards, Burda, Joseph, Brockman, Ray, Puri, Krueger, Petrov, Khlaaf, Sastry, Mishkin, Chan, Gray, Ryder, Pavlov, Power, Kaiser, Bavarian, Winter, Tillet, Such, Cummings, Plappert, Chantzis, Barnes, Herbert{-}Voss, Guss, Nichol, Paino, Tezak, Tang, Babuschkin, Balaji, Jain, Saunders, Hesse, Carr, Leike, Achiam, Misra, Morikawa, Radford, Knight, Brundage, Murati, Mayer, Welinder, McGrew, Amodei, McCandlish, Sutskever, and Zaremba}]{DBLP:journals/corr/ChenTJ21}
Mark Chen, Jerry Tworek, Heewoo Jun, Qiming Yuan, Henrique~Pond{\'{e}} de~Oliveira~Pinto, Jared Kaplan, Harrison Edwards, Yuri Burda, Nicholas Joseph, Greg Brockman, Alex Ray, Raul Puri, Gretchen Krueger, Michael Petrov, Heidy Khlaaf, Girish Sastry, Pamela Mishkin, Brooke Chan, Scott Gray, Nick Ryder, Mikhail Pavlov, Alethea Power, Lukasz Kaiser, Mohammad Bavarian, Clemens Winter, Philippe Tillet, Felipe~Petroski Such, Dave Cummings, Matthias Plappert, Fotios Chantzis, Elizabeth Barnes, Ariel Herbert{-}Voss, William~Hebgen Guss, Alex Nichol, Alex Paino, Nikolas Tezak, Jie Tang, Igor Babuschkin, Suchir Balaji, Shantanu Jain, William Saunders, Christopher Hesse, Andrew~N. Carr, Jan Leike, Joshua Achiam, Vedant Misra, Evan Morikawa, Alec Radford, Matthew Knight, Miles Brundage, Mira Murati, Katie Mayer, Peter Welinder, Bob McGrew, Dario Amodei, Sam McCandlish, Ilya Sutskever, and Wojciech Zaremba. 2021.
\newblock \href {https://arxiv.org/abs/2107.03374} {Evaluating large language models trained on code}.
\newblock \emph{CoRR}, abs/2107.03374.

\bibitem[{Chen et~al.(2023)Chen, Jiang, Chen, Wang, Yu, Chen, Zhang, Liang, Zhang, Zhang, Li, Wan, Wang, and Li}]{DBLP:journals/corr/ChenJC23}
Zhihong Chen, Feng Jiang, Junying Chen, Tiannan Wang, Fei Yu, Guiming Chen, Hongbo Zhang, Juhao Liang, Chen Zhang, Zhiyi Zhang, Jianquan Li, Xiang Wan, Benyou Wang, and Haizhou Li. 2023.
\newblock \href {https://doi.org/10.48550/ARXIV.2304.10453} {Phoenix: Democratizing chatgpt across languages}.
\newblock \emph{CoRR}, abs/2304.10453.

\bibitem[{Chiang et~al.(2023)Chiang, Li, Lin, Sheng, Wu, Zhang, Zheng, Zhuang, Zhuang, Gonzalez, Stoica, and Xing}]{Vicuna23}
Wei-Lin Chiang, Zhuohan Li, Zi~Lin, Ying Sheng, Zhanghao Wu, Hao Zhang, Lianmin Zheng, Siyuan Zhuang, Yonghao Zhuang, Joseph~E. Gonzalez, Ion Stoica, and Eric~P. Xing. 2023.
\newblock \href {https://lmsys.org/blog/2023-03-30-vicuna/} {Vicuna: An open-source chatbot impressing gpt-4 with 90\%* chatgpt quality}.

\bibitem[{Chowdhery et~al.(2023)Chowdhery, Narang, Devlin, Bosma, Mishra, Roberts, Barham, Chung, Sutton, Gehrmann, Schuh, Shi, Tsvyashchenko, Maynez, Rao, Barnes, Tay, Shazeer, Prabhakaran, Reif, Du, Hutchinson, Pope, Bradbury, Austin, Isard, Gur{-}Ari, Yin, Duke, Levskaya, Ghemawat, Dev, Michalewski, Garcia, Misra, Robinson, Fedus, Zhou, Ippolito, Luan, Lim, Zoph, Spiridonov, Sepassi, Dohan, Agrawal, Omernick, Dai, Pillai, Pellat, Lewkowycz, Moreira, Child, Polozov, Lee, Zhou, Wang, Saeta, Diaz, Firat, Catasta, Wei, Meier{-}Hellstern, Eck, Dean, Petrov, and Fiedel}]{DBLP:journals/jmlr/ChowdheryNDBMRBCSGSSTMRBTSPRDHPBAI23}
Aakanksha Chowdhery, Sharan Narang, Jacob Devlin, Maarten Bosma, Gaurav Mishra, Adam Roberts, Paul Barham, Hyung~Won Chung, Charles Sutton, Sebastian Gehrmann, Parker Schuh, Kensen Shi, Sasha Tsvyashchenko, Joshua Maynez, Abhishek Rao, Parker Barnes, Yi~Tay, Noam Shazeer, Vinodkumar Prabhakaran, Emily Reif, Nan Du, Ben Hutchinson, Reiner Pope, James Bradbury, Jacob Austin, Michael Isard, Guy Gur{-}Ari, Pengcheng Yin, Toju Duke, Anselm Levskaya, Sanjay Ghemawat, Sunipa Dev, Henryk Michalewski, Xavier Garcia, Vedant Misra, Kevin Robinson, Liam Fedus, Denny Zhou, Daphne Ippolito, David Luan, Hyeontaek Lim, Barret Zoph, Alexander Spiridonov, Ryan Sepassi, David Dohan, Shivani Agrawal, Mark Omernick, Andrew~M. Dai, Thanumalayan~Sankaranarayana Pillai, Marie Pellat, Aitor Lewkowycz, Erica Moreira, Rewon Child, Oleksandr Polozov, Katherine Lee, Zongwei Zhou, Xuezhi Wang, Brennan Saeta, Mark Diaz, Orhan Firat, Michele Catasta, Jason Wei, Kathy Meier{-}Hellstern, Douglas Eck, Jeff Dean, Slav Petrov, and Noah Fiedel.
  2023.
\newblock \href {http://jmlr.org/papers/v24/22-1144.html} {Palm: Scaling language modeling with pathways}.
\newblock \emph{J. Mach. Learn. Res.}, 24:240:1--240:113.

\bibitem[{Chung et~al.(2022)Chung, Hou, Longpre, Zoph, Tay, Fedus, Li, Wang, Dehghani, Brahma, Webson, Gu, Dai, Suzgun, Chen, Chowdhery, Narang, Mishra, Yu, Zhao, Huang, Dai, Yu, Petrov, Chi, Dean, Devlin, Roberts, Zhou, Le, and Wei}]{DBLP:journals/corr/ChungHL22}
Hyung~Won Chung, Le~Hou, Shayne Longpre, Barret Zoph, Yi~Tay, William Fedus, Eric Li, Xuezhi Wang, Mostafa Dehghani, Siddhartha Brahma, Albert Webson, Shixiang~Shane Gu, Zhuyun Dai, Mirac Suzgun, Xinyun Chen, Aakanksha Chowdhery, Sharan Narang, Gaurav Mishra, Adams Yu, Vincent~Y. Zhao, Yanping Huang, Andrew~M. Dai, Hongkun Yu, Slav Petrov, Ed~H. Chi, Jeff Dean, Jacob Devlin, Adam Roberts, Denny Zhou, Quoc~V. Le, and Jason Wei. 2022.
\newblock \href {https://doi.org/10.48550/ARXIV.2210.11416} {Scaling instruction-finetuned language models}.
\newblock \emph{CoRR}, abs/2210.11416.

\bibitem[{Clark et~al.(2018)Clark, Cowhey, Etzioni, Khot, Sabharwal, Schoenick, and Tafjord}]{DBLP:journals/corr/ClarkCE18}
Peter Clark, Isaac Cowhey, Oren Etzioni, Tushar Khot, Ashish Sabharwal, Carissa Schoenick, and Oyvind Tafjord. 2018.
\newblock \href {https://arxiv.org/abs/1803.05457} {Think you have solved question answering? try arc, the {AI2} reasoning challenge}.
\newblock \emph{CoRR}, abs/1803.05457.

\bibitem[{Cobbe et~al.(2021)Cobbe, Kosaraju, Bavarian, Chen, Jun, Kaiser, Plappert, Tworek, Hilton, Nakano, Hesse, and Schulman}]{DBLP:journals/corr/CobbeKB21}
Karl Cobbe, Vineet Kosaraju, Mohammad Bavarian, Mark Chen, Heewoo Jun, Lukasz Kaiser, Matthias Plappert, Jerry Tworek, Jacob Hilton, Reiichiro Nakano, Christopher Hesse, and John Schulman. 2021.
\newblock \href {https://arxiv.org/abs/2110.14168} {Training verifiers to solve math word problems}.
\newblock \emph{CoRR}, abs/2110.14168.

\bibitem[{Cui et~al.(2023)Cui, Yang, and Yao}]{DBLP:journals/corr/CuiYY23}
Yiming Cui, Ziqing Yang, and Xin Yao. 2023.
\newblock \href {https://doi.org/10.48550/ARXIV.2304.08177} {Efficient and effective text encoding for chinese llama and alpaca}.
\newblock \emph{CoRR}, abs/2304.08177.

\bibitem[{Dao et~al.(2022)Dao, Fu, Ermon, Rudra, and R{\'{e}}}]{DBLP:conf/nips/DaoFERR22}
Tri Dao, Daniel~Y. Fu, Stefano Ermon, Atri Rudra, and Christopher R{\'{e}}. 2022.
\newblock \href {http://papers.nips.cc/paper\_files/paper/2022/hash/67d57c32e20fd0a7a302cb81d36e40d5-Abstract-Conference.html} {Flashattention: Fast and memory-efficient exact attention with io-awareness}.
\newblock In \emph{NeurIPS}.

\bibitem[{Dettmers et~al.(2023)Dettmers, Pagnoni, Holtzman, and Zettlemoyer}]{DBLP:journals/corr/DettmersPH23}
Tim Dettmers, Artidoro Pagnoni, Ari Holtzman, and Luke Zettlemoyer. 2023.
\newblock \href {https://doi.org/10.48550/ARXIV.2305.14314} {Qlora: Efficient finetuning of quantized llms}.
\newblock \emph{CoRR}, abs/2305.14314.

\bibitem[{Dettmers and Zettlemoyer(2023)}]{DBLP:conf/icml/DettmersZ23}
Tim Dettmers and Luke Zettlemoyer. 2023.
\newblock \href {https://proceedings.mlr.press/v202/dettmers23a.html} {The case for 4-bit precision: k-bit inference scaling laws}.
\newblock In \emph{International Conference on Machine Learning, {ICML} 2023, 23-29 July 2023, Honolulu, Hawaii, {USA}}, volume 202 of \emph{Proceedings of Machine Learning Research}, pages 7750--7774. {PMLR}.

\bibitem[{Devlin et~al.(2019)Devlin, Chang, Lee, and Toutanova}]{DBLP:conf/naacl/DevlinCLT19}
Jacob Devlin, Ming{-}Wei Chang, Kenton Lee, and Kristina Toutanova. 2019.
\newblock \href {https://doi.org/10.18653/V1/N19-1423} {{BERT:} pre-training of deep bidirectional transformers for language understanding}.
\newblock In \emph{Proceedings of the 2019 Conference of the North American Chapter of the Association for Computational Linguistics: Human Language Technologies, {NAACL-HLT} 2019, Minneapolis, MN, USA, June 2-7, 2019, Volume 1 (Long and Short Papers)}, pages 4171--4186. Association for Computational Linguistics.

\bibitem[{Dey et~al.(2023{\natexlab{a}})Dey, Gosal, Chen, Khachane, Marshall, Pathria, Tom, and Hestness}]{DBLP:journals/corr/DeyGC23}
Nolan Dey, Gurpreet Gosal, Zhiming Chen, Hemant Khachane, William Marshall, Ribhu Pathria, Marvin Tom, and Joel Hestness. 2023{\natexlab{a}}.
\newblock \href {https://doi.org/10.48550/ARXIV.2304.03208} {Cerebras-gpt: Open compute-optimal language models trained on the cerebras wafer-scale cluster}.
\newblock \emph{CoRR}, abs/2304.03208.

\bibitem[{Dey et~al.(2023{\natexlab{b}})Dey, Soboleva, Al{-}Khateeb, Yang, Pathria, Khachane, Muhammad, Chen, Myers, Steeves, Vassilieva, Tom, and Hestness}]{DBLP:journals/corr/DeySA23}
Nolan Dey, Daria Soboleva, Faisal Al{-}Khateeb, Bowen Yang, Ribhu Pathria, Hemant Khachane, Shaheer Muhammad, Zhiming Chen, Robert Myers, Jacob~Robert Steeves, Natalia Vassilieva, Marvin Tom, and Joel Hestness. 2023{\natexlab{b}}.
\newblock \href {https://doi.org/10.48550/ARXIV.2309.11568} {{BTLM-3B-8K:} 7b parameter performance in a 3b parameter model}.
\newblock \emph{CoRR}, abs/2309.11568.

\bibitem[{Dua et~al.(2019)Dua, Wang, Dasigi, Stanovsky, Singh, and Gardner}]{DBLP:conf/naacl/DuaWDSS019}
Dheeru Dua, Yizhong Wang, Pradeep Dasigi, Gabriel Stanovsky, Sameer Singh, and Matt Gardner. 2019.
\newblock \href {https://doi.org/10.18653/V1/N19-1246} {{DROP:} {A} reading comprehension benchmark requiring discrete reasoning over paragraphs}.
\newblock In \emph{Proceedings of the 2019 Conference of the North American Chapter of the Association for Computational Linguistics: Human Language Technologies, {NAACL-HLT} 2019, Minneapolis, MN, USA, June 2-7, 2019, Volume 1 (Long and Short Papers)}, pages 2368--2378. Association for Computational Linguistics.

\bibitem[{Dubey et~al.(2024)Dubey, Jauhri, Pandey, Kadian, Al{-}Dahle, Letman, Mathur, Schelten, Yang, Fan, Goyal, Hartshorn, Yang, Mitra, Sravankumar, Korenev, Hinsvark, Rao, Zhang, Rodriguez, Gregerson, Spataru, Rozi{\`{e}}re, Biron, Tang, Chern, Caucheteux, Nayak, Bi, Marra, McConnell, Keller, Touret, Wu, Wong, Ferrer, Nikolaidis, Allonsius, Song, Pintz, Livshits, Esiobu, Choudhary, Mahajan, Garcia{-}Olano, Perino, Hupkes, Lakomkin, AlBadawy, Lobanova, Dinan, Smith, Radenovic, Zhang, Synnaeve, Lee, Anderson, Nail, Mialon, Pang, Cucurell, Nguyen, Korevaar, Xu, Touvron, Zarov, Ibarra, Kloumann, Misra, Evtimov, Copet, Lee, Geffert, Vranes, Park, Mahadeokar, Shah, van~der Linde, Billock, Hong, Lee, Fu, Chi, Huang, Liu, Wang, Yu, Bitton, Spisak, Park, Rocca, Johnstun, Saxe, Jia, Alwala, Upasani, Plawiak, Li, Heafield, Stone, and et~al.}]{DBLP:journals/corr/DubeyJP24}
Abhimanyu Dubey, Abhinav Jauhri, Abhinav Pandey, Abhishek Kadian, Ahmad Al{-}Dahle, Aiesha Letman, Akhil Mathur, Alan Schelten, Amy Yang, Angela Fan, Anirudh Goyal, Anthony Hartshorn, Aobo Yang, Archi Mitra, Archie Sravankumar, Artem Korenev, Arthur Hinsvark, Arun Rao, Aston Zhang, Aur{\'{e}}lien Rodriguez, Austen Gregerson, Ava Spataru, Baptiste Rozi{\`{e}}re, Bethany Biron, Binh Tang, Bobbie Chern, Charlotte Caucheteux, Chaya Nayak, Chloe Bi, Chris Marra, Chris McConnell, Christian Keller, Christophe Touret, Chunyang Wu, Corinne Wong, Cristian~Canton Ferrer, Cyrus Nikolaidis, Damien Allonsius, Daniel Song, Danielle Pintz, Danny Livshits, David Esiobu, Dhruv Choudhary, Dhruv Mahajan, Diego Garcia{-}Olano, Diego Perino, Dieuwke Hupkes, Egor Lakomkin, Ehab AlBadawy, Elina Lobanova, Emily Dinan, Eric~Michael Smith, Filip Radenovic, Frank Zhang, Gabriel Synnaeve, Gabrielle Lee, Georgia~Lewis Anderson, Graeme Nail, Gr{\'{e}}goire Mialon, Guan Pang, Guillem Cucurell, Hailey Nguyen, Hannah Korevaar, Hu~Xu, Hugo
  Touvron, Iliyan Zarov, Imanol~Arrieta Ibarra, Isabel~M. Kloumann, Ishan Misra, Ivan Evtimov, Jade Copet, Jaewon Lee, Jan Geffert, Jana Vranes, Jason Park, Jay Mahadeokar, Jeet Shah, Jelmer van~der Linde, Jennifer Billock, Jenny Hong, Jenya Lee, Jeremy Fu, Jianfeng Chi, Jianyu Huang, Jiawen Liu, Jie Wang, Jiecao Yu, Joanna Bitton, Joe Spisak, Jongsoo Park, Joseph Rocca, Joshua Johnstun, Joshua Saxe, Junteng Jia, Kalyan~Vasuden Alwala, Kartikeya Upasani, Kate Plawiak, Ke~Li, Kenneth Heafield, Kevin Stone, and et~al. 2024.
\newblock \href {https://doi.org/10.48550/ARXIV.2407.21783} {The llama 3 herd of models}.
\newblock \emph{CoRR}, abs/2407.21783.

\bibitem[{Frantar and Alistarh(2023)}]{DBLP:conf/icml/FrantarA23}
Elias Frantar and Dan Alistarh. 2023.
\newblock \href {https://proceedings.mlr.press/v202/frantar23a.html} {Sparsegpt: Massive language models can be accurately pruned in one-shot}.
\newblock In \emph{International Conference on Machine Learning, {ICML} 2023, 23-29 July 2023, Honolulu, Hawaii, {USA}}, volume 202 of \emph{Proceedings of Machine Learning Research}, pages 10323--10337. {PMLR}.

\bibitem[{Frantar et~al.(2022)Frantar, Ashkboos, Hoefler, and Alistarh}]{DBLP:journals/corr/FrantarAH22}
Elias Frantar, Saleh Ashkboos, Torsten Hoefler, and Dan Alistarh. 2022.
\newblock \href {https://doi.org/10.48550/ARXIV.2210.17323} {{GPTQ:} accurate post-training quantization for generative pre-trained transformers}.
\newblock \emph{CoRR}, abs/2210.17323.

\bibitem[{Fu et~al.(2023)Fu, Peng, Ou, Sabharwal, and Khot}]{DBLP:conf/icml/FuPOSK23}
Yao Fu, Hao Peng, Litu Ou, Ashish Sabharwal, and Tushar Khot. 2023.
\newblock \href {https://proceedings.mlr.press/v202/fu23d.html} {Specializing smaller language models towards multi-step reasoning}.
\newblock In \emph{International Conference on Machine Learning, {ICML} 2023, 23-29 July 2023, Honolulu, Hawaii, {USA}}, volume 202 of \emph{Proceedings of Machine Learning Research}, pages 10421--10430. {PMLR}.

\bibitem[{Gao et~al.(2021)Gao, Biderman, Black, Golding, Hoppe, Foster, Phang, He, Thite, Nabeshima, Presser, and Leahy}]{DBLP:journals/corr/GaoBB21}
Leo Gao, Stella Biderman, Sid Black, Laurence Golding, Travis Hoppe, Charles Foster, Jason Phang, Horace He, Anish Thite, Noa Nabeshima, Shawn Presser, and Connor Leahy. 2021.
\newblock \href {https://arxiv.org/abs/2101.00027} {The pile: An 800gb dataset of diverse text for language modeling}.
\newblock \emph{CoRR}, abs/2101.00027.

\bibitem[{Geng and Liu(2023)}]{OpenLLaMA23}
Xinyang Geng and Hao Liu. 2023.
\newblock \href {https://github.com/openlm-research/open_llama} {Openllama: An open reproduction of llama}.

\bibitem[{Gokaslan and Cohen(2019)}]{GokaslanC19}
Aaron Gokaslan and Vanya Cohen. 2019.
\newblock \href {http://Skylion007.github.io/OpenWebTextCorpus} {Openwebtext corpus}.

\bibitem[{Gu and Dao(2023)}]{DBLP:journals/corr/GuD23}
Albert Gu and Tri Dao. 2023.
\newblock \href {https://doi.org/10.48550/ARXIV.2312.00752} {Mamba: Linear-time sequence modeling with selective state spaces}.
\newblock \emph{CoRR}, abs/2312.00752.

\bibitem[{Gu et~al.(2023)Gu, Dong, Wei, and Huang}]{DBLP:journals/corr/GuDWH23}
Yuxian Gu, Li~Dong, Furu Wei, and Minlie Huang. 2023.
\newblock \href {https://doi.org/10.48550/ARXIV.2306.08543} {Knowledge distillation of large language models}.
\newblock \emph{CoRR}, abs/2306.08543.

\bibitem[{Gunasekar et~al.(2023)Gunasekar, Zhang, Aneja, Mendes, Giorno, Gopi, Javaheripi, Kauffmann, de~Rosa, Saarikivi, Salim, Shah, Behl, Wang, Bubeck, Eldan, Kalai, Lee, and Li}]{DBLP:journals/corr/GunasekarZA23}
Suriya Gunasekar, Yi~Zhang, Jyoti Aneja, Caio C{\'{e}}sar~Teodoro Mendes, Allie~Del Giorno, Sivakanth Gopi, Mojan Javaheripi, Piero Kauffmann, Gustavo de~Rosa, Olli Saarikivi, Adil Salim, Shital Shah, Harkirat~Singh Behl, Xin Wang, S{\'{e}}bastien Bubeck, Ronen Eldan, Adam~Tauman Kalai, Yin~Tat Lee, and Yuanzhi Li. 2023.
\newblock \href {https://doi.org/10.48550/ARXIV.2306.11644} {Textbooks are all you need}.
\newblock \emph{CoRR}, abs/2306.11644.

\bibitem[{Gunter et~al.(2024)Gunter, Wang, Wang, Pang, Narayanan, Zhang, Zhang, Chen, Chiu, Qiu, Gopinath, Yap, Yin, Nan, Weers, Yin, Huang, Wang, Lu, Peebles, Ye, Lee, Du, Chen, Keunebroek, Wiseman, Evans, Lei, Rathod, Kong, Du, Li, Wang, Gao, Ahmed, Xu, Lu, Rashid, Jose, Doane, Bencomo, Vanderby, Hansen, Jain, Anupama, Kamal, Wu, Brum, Maalouf, Erdenebileg, Dulhanty, Moritz, Kang, Jimenez, Ladd, Shi, Bai, Chu, Hohman, Kotek, Coleman, Li, Bigham, Cao, Lai, Cheung, Shan, Zhou, Li, Qin, Singh, Vega, Zou, Heckman, Gardiner, Bowler, Cordell, Cao, Hay, Shahdadpuri, Godwin, Dighe, Rachapudi, Tantawi, Frigg, Davarnia, Shah, Guha, Sirovica, Ma, Ma, Wang, Kim, Jayaram, Shankar, Paidi, Kumar, Wang, Zheng, and Cheng}]{DBLP:journals/corr/GunerWW24}
Tom Gunter, Zirui Wang, Chong Wang, Ruoming Pang, Andy Narayanan, Aonan Zhang, Bowen Zhang, Chen Chen, Chung{-}Cheng Chiu, David Qiu, Deepak Gopinath, Dian~Ang Yap, Dong Yin, Feng Nan, Floris Weers, Guoli Yin, Haoshuo Huang, Jianyu Wang, Jiarui Lu, John Peebles, Ke~Ye, Mark Lee, Nan Du, Qibin Chen, Quentin Keunebroek, Sam Wiseman, Syd Evans, Tao Lei, Vivek Rathod, Xiang Kong, Xianzhi Du, Yanghao Li, Yongqiang Wang, Yuan Gao, Zaid Ahmed, Zhaoyang Xu, Zhiyun Lu, Al~Rashid, Albin~Madappally Jose, Alec Doane, Alfredo Bencomo, Allison Vanderby, Andrew Hansen, Ankur Jain, Anupama~Mann Anupama, Areeba Kamal, Bugu Wu, Carolina Brum, Charlie Maalouf, Chinguun Erdenebileg, Chris Dulhanty, Dominik Moritz, Doug Kang, Eduardo Jimenez, Evan Ladd, Fangping Shi, Felix Bai, Frank Chu, Fred Hohman, Hadas Kotek, Hannah~Gillis Coleman, Jane Li, Jeffrey~P. Bigham, Jeffery Cao, Jeff Lai, Jessica Cheung, Jiulong Shan, Joe Zhou, John Li, Jun Qin, Karanjeet Singh, Karla Vega, Kelvin Zou, Laura Heckman, Lauren Gardiner, Margit Bowler,
  Maria Cordell, Meng Cao, Nicole Hay, Nilesh Shahdadpuri, Otto Godwin, Pranay Dighe, Pushyami Rachapudi, Ramsey Tantawi, Roman Frigg, Sam Davarnia, Sanskruti Shah, Saptarshi Guha, Sasha Sirovica, Shen Ma, Shuang Ma, Simon Wang, Sulgi Kim, Suma Jayaram, Vaishaal Shankar, Varsha Paidi, Vivek Kumar, Xin Wang, Xin Zheng, and Walker Cheng. 2024.
\newblock \href {https://doi.org/10.48550/ARXIV.2407.21075} {Apple intelligence foundation language models}.
\newblock \emph{CoRR}, abs/2407.21075.

\bibitem[{Hendrycks et~al.(2021)Hendrycks, Burns, Basart, Zou, Mazeika, Song, and Steinhardt}]{DBLP:conf/iclr/HendrycksBBZMSS21}
Dan Hendrycks, Collin Burns, Steven Basart, Andy Zou, Mantas Mazeika, Dawn Song, and Jacob Steinhardt. 2021.
\newblock \href {https://openreview.net/forum?id=d7KBjmI3GmQ} {Measuring massive multitask language understanding}.
\newblock In \emph{9th International Conference on Learning Representations, {ICLR} 2021, Virtual Event, Austria, May 3-7, 2021}. OpenReview.net.

\bibitem[{Hinton et~al.(2015)Hinton, Vinyals, and Dean}]{DBLP:journals/corr/HintonVD15}
Geoffrey~E. Hinton, Oriol Vinyals, and Jeffrey Dean. 2015.
\newblock \href {https://arxiv.org/abs/1503.02531} {Distilling the knowledge in a neural network}.
\newblock \emph{CoRR}, abs/1503.02531.

\bibitem[{Ho et~al.(2023)Ho, Schmid, and Yun}]{DBLP:conf/acl/HoSY23}
Namgyu Ho, Laura Schmid, and Se{-}Young Yun. 2023.
\newblock \href {https://doi.org/10.18653/V1/2023.ACL-LONG.830} {Large language models are reasoning teachers}.
\newblock In \emph{Proceedings of the 61st Annual Meeting of the Association for Computational Linguistics (Volume 1: Long Papers), {ACL} 2023, Toronto, Canada, July 9-14, 2023}, pages 14852--14882. Association for Computational Linguistics.

\bibitem[{Honovich et~al.(2023)Honovich, Scialom, Levy, and Schick}]{DBLP:conf/acl/HonovichSLS23}
Or~Honovich, Thomas Scialom, Omer Levy, and Timo Schick. 2023.
\newblock \href {https://doi.org/10.18653/V1/2023.ACL-LONG.806} {Unnatural instructions: Tuning language models with (almost) no human labor}.
\newblock In \emph{Proceedings of the 61st Annual Meeting of the Association for Computational Linguistics (Volume 1: Long Papers), {ACL} 2023, Toronto, Canada, July 9-14, 2023}, pages 14409--14428. Association for Computational Linguistics.

\bibitem[{Hsieh et~al.(2023)Hsieh, Li, Yeh, Nakhost, Fujii, Ratner, Krishna, Lee, and Pfister}]{DBLP:conf/acl/HsiehLYNFRKLP23}
Cheng{-}Yu Hsieh, Chun{-}Liang Li, Chih{-}Kuan Yeh, Hootan Nakhost, Yasuhisa Fujii, Alex Ratner, Ranjay Krishna, Chen{-}Yu Lee, and Tomas Pfister. 2023.
\newblock \href {https://doi.org/10.18653/V1/2023.FINDINGS-ACL.507} {Distilling step-by-step! outperforming larger language models with less training data and smaller model sizes}.
\newblock In \emph{Findings of the Association for Computational Linguistics: {ACL} 2023, Toronto, Canada, July 9-14, 2023}, pages 8003--8017. Association for Computational Linguistics.

\bibitem[{Huang et~al.(2023)Huang, Bai, Zhu, Zhang, Zhang, Su, Liu, Lv, Zhang, Lei, Fu, Sun, and He}]{DBLP:journals/corr/HuangBZ23}
Yuzhen Huang, Yuzhuo Bai, Zhihao Zhu, Junlei Zhang, Jinghan Zhang, Tangjun Su, Junteng Liu, Chuancheng Lv, Yikai Zhang, Jiayi Lei, Yao Fu, Maosong Sun, and Junxian He. 2023.
\newblock \href {https://doi.org/10.48550/ARXIV.2305.08322} {C-eval: {A} multi-level multi-discipline chinese evaluation suite for foundation models}.
\newblock \emph{CoRR}, abs/2305.08322.

\bibitem[{Jacobs et~al.(2023)Jacobs, Tanaka, Zhang, Zhang, Song, Rajbhandari, and He}]{DBLP:journals/corr/JacobsTZ23}
Sam~Ade Jacobs, Masahiro Tanaka, Chengming Zhang, Minjia Zhang, Shuaiwen~Leon Song, Samyam Rajbhandari, and Yuxiong He. 2023.
\newblock \href {https://doi.org/10.48550/ARXIV.2309.14509} {Deepspeed ulysses: System optimizations for enabling training of extreme long sequence transformer models}.
\newblock \emph{CoRR}, abs/2309.14509.

\bibitem[{Jha et~al.(2023)Jha, Groeneveld, Strubell, and Beltagy}]{DBLP:journals/corr/JhaGS23}
Ananya~Harsh Jha, Dirk Groeneveld, Emma Strubell, and Iz~Beltagy. 2023.
\newblock \href {https://doi.org/10.48550/ARXIV.2305.14864} {Large language model distillation doesn't need a teacher}.
\newblock \emph{CoRR}, abs/2305.14864.

\bibitem[{Ji et~al.(2023)Ji, Gong, Deng, Peng, Niu, Ma, and Li}]{DBLP:journals/corr/JiGD23}
Yunjie Ji, Yan Gong, Yong Deng, Yiping Peng, Qiang Niu, Baochang Ma, and Xiangang Li. 2023.
\newblock \href {https://doi.org/10.48550/ARXIV.2304.07854} {Towards better instruction following language models for chinese: Investigating the impact of training data and evaluation}.
\newblock \emph{CoRR}, abs/2304.07854.

\bibitem[{Jiang et~al.(2023)Jiang, Sablayrolles, Mensch, Bamford, Chaplot, de~Las~Casas, Bressand, Lengyel, Lample, Saulnier, Lavaud, Lachaux, Stock, Scao, Lavril, Wang, Lacroix, and Sayed}]{DBLP:journals/corr/JiangSM23}
Albert~Q. Jiang, Alexandre Sablayrolles, Arthur Mensch, Chris Bamford, Devendra~Singh Chaplot, Diego de~Las~Casas, Florian Bressand, Gianna Lengyel, Guillaume Lample, Lucile Saulnier, L{\'{e}}lio~Renard Lavaud, Marie{-}Anne Lachaux, Pierre Stock, Teven~Le Scao, Thibaut Lavril, Thomas Wang, Timoth{\'{e}}e Lacroix, and William~El Sayed. 2023.
\newblock \href {https://doi.org/10.48550/ARXIV.2310.06825} {Mistral 7b}.
\newblock \emph{CoRR}, abs/2310.06825.

\bibitem[{Jiao et~al.(2020)Jiao, Yin, Shang, Jiang, Chen, Li, Wang, and Liu}]{DBLP:conf/emnlp/JiaoYSJCL0L20}
Xiaoqi Jiao, Yichun Yin, Lifeng Shang, Xin Jiang, Xiao Chen, Linlin Li, Fang Wang, and Qun Liu. 2020.
\newblock \href {https://doi.org/10.18653/V1/2020.FINDINGS-EMNLP.372} {Tinybert: Distilling {BERT} for natural language understanding}.
\newblock In \emph{Findings of the Association for Computational Linguistics: {EMNLP} 2020, Online Event, 16-20 November 2020}, volume {EMNLP} 2020 of \emph{Findings of {ACL}}, pages 4163--4174. Association for Computational Linguistics.

\bibitem[{Kaplan et~al.(2020)Kaplan, McCandlish, Henighan, Brown, Chess, Child, Gray, Radford, Wu, and Amodei}]{DBLP:journals/corr/KaplanMH20}
Jared Kaplan, Sam McCandlish, Tom Henighan, Tom~B. Brown, Benjamin Chess, Rewon Child, Scott Gray, Alec Radford, Jeffrey Wu, and Dario Amodei. 2020.
\newblock \href {https://arxiv.org/abs/2001.08361} {Scaling laws for neural language models}.
\newblock \emph{CoRR}, abs/2001.08361.

\bibitem[{Kim et~al.(2021)Kim, Gholami, Yao, Mahoney, and Keutzer}]{DBLP:conf/icml/KimGYMK21}
Sehoon Kim, Amir Gholami, Zhewei Yao, Michael~W. Mahoney, and Kurt Keutzer. 2021.
\newblock \href {http://proceedings.mlr.press/v139/kim21d.html} {{I-BERT:} integer-only {BERT} quantization}.
\newblock In \emph{Proceedings of the 38th International Conference on Machine Learning, {ICML} 2021, 18-24 July 2021, Virtual Event}, volume 139 of \emph{Proceedings of Machine Learning Research}, pages 5506--5518. {PMLR}.

\bibitem[{Kim and Rush(2016)}]{DBLP:conf/emnlp/KimR16}
Yoon Kim and Alexander~M. Rush. 2016.
\newblock \href {https://doi.org/10.18653/V1/D16-1139} {Sequence-level knowledge distillation}.
\newblock In \emph{Proceedings of the 2016 Conference on Empirical Methods in Natural Language Processing, {EMNLP} 2016, Austin, Texas, USA, November 1-4, 2016}, pages 1317--1327. The Association for Computational Linguistics.

\bibitem[{Komatsuzaki et~al.(2023)Komatsuzaki, Puigcerver, Lee{-}Thorp, Ruiz, Mustafa, Ainslie, Tay, Dehghani, and Houlsby}]{DBLP:conf/iclr/KomatsuzakiPLRM23}
Aran Komatsuzaki, Joan Puigcerver, James Lee{-}Thorp, Carlos~Riquelme Ruiz, Basil Mustafa, Joshua Ainslie, Yi~Tay, Mostafa Dehghani, and Neil Houlsby. 2023.
\newblock \href {https://openreview.net/pdf?id=T5nUQDrM4u} {Sparse upcycling: Training mixture-of-experts from dense checkpoints}.
\newblock In \emph{The Eleventh International Conference on Learning Representations, {ICLR} 2023, Kigali, Rwanda, May 1-5, 2023}. OpenReview.net.

\bibitem[{Kudo and Richardson(2018)}]{DBLP:conf/emnlp/KudoR18}
Taku Kudo and John Richardson. 2018.
\newblock \href {https://doi.org/10.18653/V1/D18-2012} {Sentencepiece: {A} simple and language independent subword tokenizer and detokenizer for neural text processing}.
\newblock In \emph{Proceedings of the 2018 Conference on Empirical Methods in Natural Language Processing, {EMNLP} 2018: System Demonstrations, Brussels, Belgium, October 31 - November 4, 2018}, pages 66--71. Association for Computational Linguistics.

\bibitem[{Li et~al.(2021)Li, Shao, Sun, Yan, Qiu, and Huang}]{DBLP:conf/acl/LiSSYQH20}
Xiaonan Li, Yunfan Shao, Tianxiang Sun, Hang Yan, Xipeng Qiu, and Xuanjing Huang. 2021.
\newblock \href {https://doi.org/10.18653/V1/2021.ACL-LONG.16} {Accelerating {BERT} inference for sequence labeling via early-exit}.
\newblock In \emph{Proceedings of the 59th Annual Meeting of the Association for Computational Linguistics and the 11th International Joint Conference on Natural Language Processing, {ACL/IJCNLP} 2021, (Volume 1: Long Papers), Virtual Event, August 1-6, 2021}, pages 189--199. Association for Computational Linguistics.

\bibitem[{Li et~al.(2023)Li, Bubeck, Eldan, Giorno, Gunasekar, and Lee}]{DBLP:journals/corr/LiBE23}
Yuanzhi Li, S{\'{e}}bastien Bubeck, Ronen Eldan, Allie~Del Giorno, Suriya Gunasekar, and Yin~Tat Lee. 2023.
\newblock \href {https://doi.org/10.48550/ARXIV.2309.05463} {Textbooks are all you need {II:} phi-1.5 technical report}.
\newblock \emph{CoRR}, abs/2309.05463.

\bibitem[{Li(2023)}]{BiLLA23}
Zhongli Li. 2023.
\newblock \href {https://github.com/Neutralzz/BiLLa} {Billa: A bilingual llama with enhanced reasoning ability}.

\bibitem[{Lin et~al.(2022)Lin, Hilton, and Evans}]{DBLP:conf/acl/LinHE22}
Stephanie Lin, Jacob Hilton, and Owain Evans. 2022.
\newblock \href {https://doi.org/10.18653/V1/2022.ACL-LONG.229} {Truthfulqa: Measuring how models mimic human falsehoods}.
\newblock In \emph{Proceedings of the 60th Annual Meeting of the Association for Computational Linguistics (Volume 1: Long Papers), {ACL} 2022, Dublin, Ireland, May 22-27, 2022}, pages 3214--3252. Association for Computational Linguistics.

\bibitem[{Liu et~al.(2019)Liu, Ott, Goyal, Du, Joshi, Chen, Levy, Lewis, Zettlemoyer, and Stoyanov}]{DBLP:journals/corr/LiuOG19}
Yinhan Liu, Myle Ott, Naman Goyal, Jingfei Du, Mandar Joshi, Danqi Chen, Omer Levy, Mike Lewis, Luke Zettlemoyer, and Veselin Stoyanov. 2019.
\newblock \href {https://arxiv.org/abs/1907.11692} {Roberta: {A} robustly optimized {BERT} pretraining approach}.
\newblock \emph{CoRR}, abs/1907.11692.

\bibitem[{Ma et~al.(2023)Ma, Fang, and Wang}]{DBLP:journals/corr/MaFW23}
Xinyin Ma, Gongfan Fang, and Xinchao Wang. 2023.
\newblock \href {https://doi.org/10.48550/ARXIV.2305.11627} {Llm-pruner: On the structural pruning of large language models}.
\newblock \emph{CoRR}, abs/2305.11627.

\bibitem[{Magister et~al.(2023)Magister, Mallinson, Ad{\'{a}}mek, Malmi, and Severyn}]{DBLP:conf/acl/MagisterMAMS23}
Lucie~Charlotte Magister, Jonathan Mallinson, Jakub Ad{\'{a}}mek, Eric Malmi, and Aliaksei Severyn. 2023.
\newblock \href {https://doi.org/10.18653/V1/2023.ACL-SHORT.151} {Teaching small language models to reason}.
\newblock In \emph{Proceedings of the 61st Annual Meeting of the Association for Computational Linguistics (Volume 2: Short Papers), {ACL} 2023, Toronto, Canada, July 9-14, 2023}, pages 1773--1781. Association for Computational Linguistics.

\bibitem[{Men et~al.(2024)Men, Xu, Zhang, Wang, Lin, Lu, Han, and Chen}]{DBLP:journals/corr/MenXZ24}
Xin Men, Mingyu Xu, Qingyu Zhang, Bingning Wang, Hongyu Lin, Yaojie Lu, Xianpei Han, and Weipeng Chen. 2024.
\newblock \href {https://doi.org/10.48550/ARXIV.2403.03853} {Shortgpt: Layers in large language models are more redundant than you expect}.
\newblock \emph{CoRR}, abs/2403.03853.

\bibitem[{Merity et~al.(2017)Merity, Xiong, Bradbury, and Socher}]{DBLP:conf/iclr/MerityX0S17}
Stephen Merity, Caiming Xiong, James Bradbury, and Richard Socher. 2017.
\newblock \href {https://openreview.net/forum?id=Byj72udxe} {Pointer sentinel mixture models}.
\newblock In \emph{5th International Conference on Learning Representations, {ICLR} 2017, Toulon, France, April 24-26, 2017, Conference Track Proceedings}. OpenReview.net.

\bibitem[{Michel et~al.(2019)Michel, Levy, and Neubig}]{DBLP:conf/nips/MichelLN19}
Paul Michel, Omer Levy, and Graham Neubig. 2019.
\newblock \href {https://proceedings.neurips.cc/paper/2019/hash/2c601ad9d2ff9bc8b282670cdd54f69f-Abstract.html} {Are sixteen heads really better than one?}
\newblock In \emph{Advances in Neural Information Processing Systems 32: Annual Conference on Neural Information Processing Systems 2019, NeurIPS 2019, December 8-14, 2019, Vancouver, BC, Canada}, pages 14014--14024.

\bibitem[{Mirzadeh et~al.(2020)Mirzadeh, Farajtabar, Li, Levine, Matsukawa, and Ghasemzadeh}]{DBLP:conf/aaai/MirzadehFLLMG20}
Seyed{-}Iman Mirzadeh, Mehrdad Farajtabar, Ang Li, Nir Levine, Akihiro Matsukawa, and Hassan Ghasemzadeh. 2020.
\newblock \href {https://doi.org/10.1609/AAAI.V34I04.5963} {Improved knowledge distillation via teacher assistant}.
\newblock In \emph{The Thirty-Fourth {AAAI} Conference on Artificial Intelligence, {AAAI} 2020, The Thirty-Second Innovative Applications of Artificial Intelligence Conference, {IAAI} 2020, The Tenth {AAAI} Symposium on Educational Advances in Artificial Intelligence, {EAAI} 2020, New York, NY, USA, February 7-12, 2020}, pages 5191--5198. {AAAI} Press.

\bibitem[{Molchanov et~al.(2017)Molchanov, Tyree, Karras, Aila, and Kautz}]{DBLP:conf/iclr/MolchanovTKAK17}
Pavlo Molchanov, Stephen Tyree, Tero Karras, Timo Aila, and Jan Kautz. 2017.
\newblock \href {https://openreview.net/forum?id=SJGCiw5gl} {Pruning convolutional neural networks for resource efficient inference}.
\newblock In \emph{5th International Conference on Learning Representations, {ICLR} 2017, Toulon, France, April 24-26, 2017, Conference Track Proceedings}. OpenReview.net.

\bibitem[{Muralidharan et~al.(2024)Muralidharan, Sreenivas, Joshi, Chochowski, Patwary, Shoeybi, Catanzaro, Kautz, and Molchanov}]{DBLP:journals/corr/MuralidharanSJ24}
Saurav Muralidharan, Sharath~Turuvekere Sreenivas, Raviraj Joshi, Marcin Chochowski, Mostofa Patwary, Mohammad Shoeybi, Bryan Catanzaro, Jan Kautz, and Pavlo Molchanov. 2024.
\newblock \href {https://doi.org/10.48550/ARXIV.2407.14679} {Compact language models via pruning and knowledge distillation}.
\newblock \emph{CoRR}, abs/2407.14679.

\bibitem[{OpenAI(2022)}]{OpenAI22}
OpenAI. 2022.
\newblock \href {https://openai.com/blog/chatgpt} {Introducing chatgpt}.

\bibitem[{OpenBuddyAI(2023)}]{OpenBuddy23}
OpenBuddyAI. 2023.
\newblock \href {https://openbuddy.ai} {Openbuddy - open multilingual chatbot for everyone}.

\bibitem[{Paperno et~al.(2016)Paperno, Kruszewski, Lazaridou, Pham, Bernardi, Pezzelle, Baroni, Boleda, and Fern{\'{a}}ndez}]{DBLP:conf/acl/PapernoKLPBPBBF16}
Denis Paperno, Germ{\'{a}}n Kruszewski, Angeliki Lazaridou, Quan~Ngoc Pham, Raffaella Bernardi, Sandro Pezzelle, Marco Baroni, Gemma Boleda, and Raquel Fern{\'{a}}ndez. 2016.
\newblock \href {https://doi.org/10.18653/V1/P16-1144} {The {LAMBADA} dataset: Word prediction requiring a broad discourse context}.
\newblock In \emph{Proceedings of the 54th Annual Meeting of the Association for Computational Linguistics, {ACL} 2016, August 7-12, 2016, Berlin, Germany, Volume 1: Long Papers}. The Association for Computer Linguistics.

\bibitem[{Penedo et~al.(2023)Penedo, Malartic, Hesslow, Cojocaru, Cappelli, Alobeidli, Pannier, Almazrouei, and Launay}]{DBLP:journals/corr/PenedoMH23}
Guilherme Penedo, Quentin Malartic, Daniel Hesslow, Ruxandra Cojocaru, Alessandro Cappelli, Hamza Alobeidli, Baptiste Pannier, Ebtesam Almazrouei, and Julien Launay. 2023.
\newblock \href {https://doi.org/10.48550/ARXIV.2306.01116} {The refinedweb dataset for falcon {LLM:} outperforming curated corpora with web data, and web data only}.
\newblock \emph{CoRR}, abs/2306.01116.

\bibitem[{Pereyra et~al.(2017)Pereyra, Tucker, Chorowski, Kaiser, and Hinton}]{DBLP:conf/iclr/PereyraTCKH17}
Gabriel Pereyra, George Tucker, Jan Chorowski, Lukasz Kaiser, and Geoffrey~E. Hinton. 2017.
\newblock \href {https://openreview.net/forum?id=HyhbYrGYe} {Regularizing neural networks by penalizing confident output distributions}.
\newblock In \emph{5th International Conference on Learning Representations, {ICLR} 2017, Toulon, France, April 24-26, 2017, Workshop Track Proceedings}. OpenReview.net.

\bibitem[{Radford et~al.(2018)Radford, Narasimhan, Salimans, and Sutskever}]{DBLP:journals/corr/RadfordNS18}
Alec Radford, Karthik Narasimhan, Tim Salimans, and Ilya Sutskever. 2018.
\newblock \href {https://s3-us-west-2.amazonaws.com/openai-assets/research-covers/language-unsupervised/language_understanding_paper.pdf} {Improving language understanding by generative pre-training}.

\bibitem[{Radford et~al.(2019)Radford, Wu, Child, Luan, Amodei, and Sutskever}]{DBLP:journals/corr/RadfordWC19}
Alec Radford, Jeff Wu, Rewon Child, David Luan, Dario Amodei, and Ilya Sutskever. 2019.
\newblock \href {https://d4mucfpksywv.cloudfront.net/better-language-models/language_models_are_unsupervised_multitask_learners.pdf} {Language models are unsupervised multitask learners}.

\bibitem[{Rafailov et~al.(2023)Rafailov, Sharma, Mitchell, Ermon, Manning, and Finn}]{DBLP:journals/corr/RafailovSM23}
Rafael Rafailov, Archit Sharma, Eric Mitchell, Stefano Ermon, Christopher~D. Manning, and Chelsea Finn. 2023.
\newblock \href {https://doi.org/10.48550/ARXIV.2305.18290} {Direct preference optimization: Your language model is secretly a reward model}.
\newblock \emph{CoRR}, abs/2305.18290.

\bibitem[{Raffel et~al.(2020)Raffel, Shazeer, Roberts, Lee, Narang, Matena, Zhou, Li, and Liu}]{DBLP:journals/jmlr/RaffelSRLNMZLL20}
Colin Raffel, Noam Shazeer, Adam Roberts, Katherine Lee, Sharan Narang, Michael Matena, Yanqi Zhou, Wei Li, and Peter~J. Liu. 2020.
\newblock \href {http://jmlr.org/papers/v21/20-074.html} {Exploring the limits of transfer learning with a unified text-to-text transformer}.
\newblock \emph{J. Mach. Learn. Res.}, 21:140:1--140:67.

\bibitem[{Rasley et~al.(2020)Rasley, Rajbhandari, Ruwase, and He}]{DBLP:conf/kdd/RasleyRRH20}
Jeff Rasley, Samyam Rajbhandari, Olatunji Ruwase, and Yuxiong He. 2020.
\newblock \href {https://doi.org/10.1145/3394486.3406703} {Deepspeed: System optimizations enable training deep learning models with over 100 billion parameters}.
\newblock In \emph{{KDD} '20: The 26th {ACM} {SIGKDD} Conference on Knowledge Discovery and Data Mining, Virtual Event, CA, USA, August 23-27, 2020}, pages 3505--3506. {ACM}.

\bibitem[{Rivi{\`{e}}re et~al.(2024)Rivi{\`{e}}re, Pathak, Sessa, Hardin, Bhupatiraju, Hussenot, Mesnard, Shahriari, Ram{\'{e}}, Ferret, Liu, Tafti, Friesen, Casbon, Ramos, Kumar, Lan, Jerome, Tsitsulin, Vieillard, Stanczyk, Girgin, Momchev, Hoffman, Thakoor, Grill, Neyshabur, Bachem, Walton, Severyn, Parrish, Ahmad, Hutchison, Abdagic, Carl, Shen, Brock, Coenen, Laforge, Paterson, Bastian, Piot, Wu, Royal, Chen, Kumar, Perry, Welty, Choquette{-}Choo, Sinopalnikov, Weinberger, Vijaykumar, Rogozinska, Herbison, Bandy, Wang, Noland, Moreira, Senter, Eltyshev, Visin, Rasskin, Wei, Cameron, Martins, Hashemi, Klimczak{-}Plucinska, Batra, Dhand, Nardini, Mein, Zhou, Svensson, Stanway, Chan, Zhou, Carrasqueira, Iljazi, Becker, Fernandez, van Amersfoort, Gordon, Lipschultz, Newlan, Ji, Mohamed, Badola, Black, Millican, McDonell, Nguyen, Sodhia, Greene, Sj{\"{o}}sund, Usui, Sifre, Heuermann, Lago, and McNealus}]{DBLP:journals/corr/RivierePS24}
Morgane Rivi{\`{e}}re, Shreya Pathak, Pier~Giuseppe Sessa, Cassidy Hardin, Surya Bhupatiraju, L{\'{e}}onard Hussenot, Thomas Mesnard, Bobak Shahriari, Alexandre Ram{\'{e}}, Johan Ferret, Peter Liu, Pouya Tafti, Abe Friesen, Michelle Casbon, Sabela Ramos, Ravin Kumar, Charline~Le Lan, Sammy Jerome, Anton Tsitsulin, Nino Vieillard, Piotr Stanczyk, Sertan Girgin, Nikola Momchev, Matt Hoffman, Shantanu Thakoor, Jean{-}Bastien Grill, Behnam Neyshabur, Olivier Bachem, Alanna Walton, Aliaksei Severyn, Alicia Parrish, Aliya Ahmad, Allen Hutchison, Alvin Abdagic, Amanda Carl, Amy Shen, Andy Brock, Andy Coenen, Anthony Laforge, Antonia Paterson, Ben Bastian, Bilal Piot, Bo~Wu, Brandon Royal, Charlie Chen, Chintu Kumar, Chris Perry, Chris Welty, Christopher~A. Choquette{-}Choo, Danila Sinopalnikov, David Weinberger, Dimple Vijaykumar, Dominika Rogozinska, Dustin Herbison, Elisa Bandy, Emma Wang, Eric Noland, Erica Moreira, Evan Senter, Evgenii Eltyshev, Francesco Visin, Gabriel Rasskin, Gary Wei, Glenn Cameron, Gus
  Martins, Hadi Hashemi, Hanna Klimczak{-}Plucinska, Harleen Batra, Harsh Dhand, Ivan Nardini, Jacinda Mein, Jack Zhou, James Svensson, Jeff Stanway, Jetha Chan, Jin~Peng Zhou, Joana Carrasqueira, Joana Iljazi, Jocelyn Becker, Joe Fernandez, Joost van Amersfoort, Josh Gordon, Josh Lipschultz, Josh Newlan, Ju{-}yeong Ji, Kareem Mohamed, Kartikeya Badola, Kat Black, Katie Millican, Keelin McDonell, Kelvin Nguyen, Kiranbir Sodhia, Kish Greene, Lars~Lowe Sj{\"{o}}sund, Lauren Usui, Laurent Sifre, Lena Heuermann, Leticia Lago, and Lilly McNealus. 2024.
\newblock \href {https://doi.org/10.48550/ARXIV.2408.00118} {Gemma 2: Improving open language models at a practical size}.
\newblock \emph{CoRR}, abs/2408.00118.

\bibitem[{Sakaguchi et~al.(2020)Sakaguchi, Bras, Bhagavatula, and Choi}]{DBLP:conf/aaai/SakaguchiBBC20}
Keisuke Sakaguchi, Ronan~Le Bras, Chandra Bhagavatula, and Yejin Choi. 2020.
\newblock \href {https://doi.org/10.1609/AAAI.V34I05.6399} {Winogrande: An adversarial winograd schema challenge at scale}.
\newblock In \emph{The Thirty-Fourth {AAAI} Conference on Artificial Intelligence, {AAAI} 2020, The Thirty-Second Innovative Applications of Artificial Intelligence Conference, {IAAI} 2020, The Tenth {AAAI} Symposium on Educational Advances in Artificial Intelligence, {EAAI} 2020, New York, NY, USA, February 7-12, 2020}, pages 8732--8740. {AAAI} Press.

\bibitem[{Sanh et~al.(2019)Sanh, Debut, Chaumond, and Wolf}]{DBLP:journals/corr/SanhDC19}
Victor Sanh, Lysandre Debut, Julien Chaumond, and Thomas Wolf. 2019.
\newblock \href {https://arxiv.org/abs/1910.01108} {Distilbert, a distilled version of {BERT:} smaller, faster, cheaper and lighter}.
\newblock \emph{CoRR}, abs/1910.01108.

\bibitem[{Scao et~al.(2022)Scao, Fan, Akiki, Pavlick, Ilic, Hesslow, Castagn{\'{e}}, Luccioni, Yvon, Gall{\'{e}}, Tow, Rush, Biderman, Webson, Ammanamanchi, Wang, Sagot, Muennighoff, del Moral, Ruwase, Bawden, Bekman, McMillan{-}Major, Beltagy, Nguyen, Saulnier, Tan, Suarez, Sanh, Lauren{\c{c}}on, Jernite, Launay, Mitchell, Raffel, Gokaslan, Simhi, Soroa, Aji, Alfassy, Rogers, Nitzav, Xu, Mou, Emezue, Klamm, Leong, van Strien, Adelani, and et~al.}]{DBLP:journals/corr/ScaoFA22}
Teven~Le Scao, Angela Fan, Christopher Akiki, Ellie Pavlick, Suzana Ilic, Daniel Hesslow, Roman Castagn{\'{e}}, Alexandra~Sasha Luccioni, Fran{\c{c}}ois Yvon, Matthias Gall{\'{e}}, Jonathan Tow, Alexander~M. Rush, Stella Biderman, Albert Webson, Pawan~Sasanka Ammanamanchi, Thomas Wang, Beno{\^{\i}}t Sagot, Niklas Muennighoff, Albert~Villanova del Moral, Olatunji Ruwase, Rachel Bawden, Stas Bekman, Angelina McMillan{-}Major, Iz~Beltagy, Huu Nguyen, Lucile Saulnier, Samson Tan, Pedro~Ortiz Suarez, Victor Sanh, Hugo Lauren{\c{c}}on, Yacine Jernite, Julien Launay, Margaret Mitchell, Colin Raffel, Aaron Gokaslan, Adi Simhi, Aitor Soroa, Alham~Fikri Aji, Amit Alfassy, Anna Rogers, Ariel~Kreisberg Nitzav, Canwen Xu, Chenghao Mou, Chris Emezue, Christopher Klamm, Colin Leong, Daniel van Strien, David~Ifeoluwa Adelani, and et~al. 2022.
\newblock \href {https://doi.org/10.48550/ARXIV.2211.05100} {{BLOOM:} {A} 176b-parameter open-access multilingual language model}.
\newblock \emph{CoRR}, abs/2211.05100.

\bibitem[{Srivastava et~al.(2022)Srivastava, Rastogi, Rao, Shoeb, Abid, Fisch, Brown, Santoro, Gupta, Garriga{-}Alonso, Kluska, Lewkowycz, Agarwal, Power, Ray, Warstadt, Kocurek, Safaya, Tazarv, Xiang, Parrish, Nie, Hussain, Askell, Dsouza, Rahane, Iyer, Andreassen, Santilli, Stuhlm{\"{u}}ller, Dai, La, Lampinen, Zou, Jiang, Chen, Vuong, Gupta, Gottardi, Norelli, Venkatesh, Gholamidavoodi, Tabassum, Menezes, Kirubarajan, Mullokandov, Sabharwal, Herrick, Efrat, Erdem, Karakas, and et~al.}]{DBLP:journals/corr/SrivastavaRR22}
Aarohi Srivastava, Abhinav Rastogi, Abhishek Rao, Abu Awal~Md Shoeb, Abubakar Abid, Adam Fisch, Adam~R. Brown, Adam Santoro, Aditya Gupta, Adri{\`{a}} Garriga{-}Alonso, Agnieszka Kluska, Aitor Lewkowycz, Akshat Agarwal, Alethea Power, Alex Ray, Alex Warstadt, Alexander~W. Kocurek, Ali Safaya, Ali Tazarv, Alice Xiang, Alicia Parrish, Allen Nie, Aman Hussain, Amanda Askell, Amanda Dsouza, Ameet Rahane, Anantharaman~S. Iyer, Anders Andreassen, Andrea Santilli, Andreas Stuhlm{\"{u}}ller, Andrew~M. Dai, Andrew La, Andrew~K. Lampinen, Andy Zou, Angela Jiang, Angelica Chen, Anh Vuong, Animesh Gupta, Anna Gottardi, Antonio Norelli, Anu Venkatesh, Arash Gholamidavoodi, Arfa Tabassum, Arul Menezes, Arun Kirubarajan, Asher Mullokandov, Ashish Sabharwal, Austin Herrick, Avia Efrat, Aykut Erdem, Ayla Karakas, and et~al. 2022.
\newblock \href {https://doi.org/10.48550/ARXIV.2206.04615} {Beyond the imitation game: Quantifying and extrapolating the capabilities of language models}.
\newblock \emph{CoRR}, abs/2206.04615.

\bibitem[{StabilityAI(2023)}]{StableLM23}
StabilityAI. 2023.
\newblock \href {https://stability.wandb.io/stability-llm/stable-lm/reports/StableLM-3B-4E1T--VmlldzoyMjU4?accessToken=u3zujipenkx5g7rtcj9qojjgxpconyjktjkli2po09nffrffdhhchq045vp0wyfo} {Stablelm-3b-4e1t technical report}.

\bibitem[{Sun et~al.(2023)Sun, Liu, Bair, and Kolter}]{DBLP:journals/corr/SunLB23}
Mingjie Sun, Zhuang Liu, Anna Bair, and J.~Zico Kolter. 2023.
\newblock \href {https://doi.org/10.48550/ARXIV.2306.11695} {A simple and effective pruning approach for large language models}.
\newblock \emph{CoRR}, abs/2306.11695.

\bibitem[{Sun et~al.(2019)Sun, Cheng, Gan, and Liu}]{DBLP:conf/emnlp/SunCGL19}
Siqi Sun, Yu~Cheng, Zhe Gan, and Jingjing Liu. 2019.
\newblock \href {https://doi.org/10.18653/V1/D19-1441} {Patient knowledge distillation for {BERT} model compression}.
\newblock In \emph{Proceedings of the 2019 Conference on Empirical Methods in Natural Language Processing and the 9th International Joint Conference on Natural Language Processing, {EMNLP-IJCNLP} 2019, Hong Kong, China, November 3-7, 2019}, pages 4322--4331. Association for Computational Linguistics.

\bibitem[{Suzgun et~al.(2023)Suzgun, Scales, Sch{\"{a}}rli, Gehrmann, Tay, Chung, Chowdhery, Le, Chi, Zhou, and Wei}]{DBLP:conf/acl/SuzgunSSGTCCLCZ23}
Mirac Suzgun, Nathan Scales, Nathanael Sch{\"{a}}rli, Sebastian Gehrmann, Yi~Tay, Hyung~Won Chung, Aakanksha Chowdhery, Quoc~V. Le, Ed~Chi, Denny Zhou, and Jason Wei. 2023.
\newblock \href {https://doi.org/10.18653/V1/2023.FINDINGS-ACL.824} {Challenging big-bench tasks and whether chain-of-thought can solve them}.
\newblock In \emph{Findings of the Association for Computational Linguistics: {ACL} 2023, Toronto, Canada, July 9-14, 2023}, pages 13003--13051. Association for Computational Linguistics.

\bibitem[{Taori et~al.(2023)Taori, Gulrajani, Zhang, Dubois, Li, Guestrin, Liang, and Hashimoto}]{Alpaca23}
Rohan Taori, Ishaan Gulrajani, Tianyi Zhang, Yann Dubois, Xuechen Li, Carlos Guestrin, Percy Liang, and Tatsunori~B. Hashimoto. 2023.
\newblock \href {https://github.com/tatsu-lab/stanford_alpaca} {Stanford alpaca: An instruction-following llama model}.

\bibitem[{Teknium(2023)}]{GPTeacher23}
Teknium. 2023.
\newblock \href {https://github.com/teknium1/GPTeacher} {A collection of modular datasets generated by gpt-4, general-instruct - roleplay-instruct - code-instruct - and toolformer}.

\bibitem[{Together(2023)}]{RedPajama23}
Together. 2023.
\newblock \href {https://together.ai/blog/redpajama} {Redpajama, a project to create leading open-source models, starts by reproducing llama training dataset of over 1.2 trillion tokens}.

\bibitem[{Touvron et~al.(2023{\natexlab{a}})Touvron, Lavril, Izacard, Martinet, Lachaux, Lacroix, Rozi{\`{e}}re, Goyal, Hambro, Azhar, Rodriguez, Joulin, Grave, and Lample}]{DBLP:journals/corr/TouvronLI23}
Hugo Touvron, Thibaut Lavril, Gautier Izacard, Xavier Martinet, Marie{-}Anne Lachaux, Timoth{\'{e}}e Lacroix, Baptiste Rozi{\`{e}}re, Naman Goyal, Eric Hambro, Faisal Azhar, Aur{\'{e}}lien Rodriguez, Armand Joulin, Edouard Grave, and Guillaume Lample. 2023{\natexlab{a}}.
\newblock \href {https://doi.org/10.48550/ARXIV.2302.13971} {Llama: Open and efficient foundation language models}.
\newblock \emph{CoRR}, abs/2302.13971.

\bibitem[{Touvron et~al.(2023{\natexlab{b}})Touvron, Martin, Stone, Albert, Almahairi, Babaei, Bashlykov, Batra, Bhargava, Bhosale, Bikel, Blecher, Canton{-}Ferrer, Chen, Cucurull, Esiobu, Fernandes, Fu, Fu, Fuller, Gao, Goswami, Goyal, Hartshorn, Hosseini, Hou, Inan, Kardas, Kerkez, Khabsa, Kloumann, Korenev, Koura, Lachaux, Lavril, Lee, Liskovich, Lu, Mao, Martinet, Mihaylov, Mishra, Molybog, Nie, Poulton, Reizenstein, Rungta, Saladi, Schelten, Silva, Smith, Subramanian, Tan, Tang, Taylor, Williams, Kuan, Xu, Yan, Zarov, Zhang, Fan, Kambadur, Narang, Rodriguez, Stojnic, Edunov, and Scialom}]{DBLP:journals/corr/TouvronMS23}
Hugo Touvron, Louis Martin, Kevin Stone, Peter Albert, Amjad Almahairi, Yasmine Babaei, Nikolay Bashlykov, Soumya Batra, Prajjwal Bhargava, Shruti Bhosale, Dan Bikel, Lukas Blecher, Cristian Canton{-}Ferrer, Moya Chen, Guillem Cucurull, David Esiobu, Jude Fernandes, Jeremy Fu, Wenyin Fu, Brian Fuller, Cynthia Gao, Vedanuj Goswami, Naman Goyal, Anthony Hartshorn, Saghar Hosseini, Rui Hou, Hakan Inan, Marcin Kardas, Viktor Kerkez, Madian Khabsa, Isabel Kloumann, Artem Korenev, Punit~Singh Koura, Marie{-}Anne Lachaux, Thibaut Lavril, Jenya Lee, Diana Liskovich, Yinghai Lu, Yuning Mao, Xavier Martinet, Todor Mihaylov, Pushkar Mishra, Igor Molybog, Yixin Nie, Andrew Poulton, Jeremy Reizenstein, Rashi Rungta, Kalyan Saladi, Alan Schelten, Ruan Silva, Eric~Michael Smith, Ranjan Subramanian, Xiaoqing~Ellen Tan, Binh Tang, Ross Taylor, Adina Williams, Jian~Xiang Kuan, Puxin Xu, Zheng Yan, Iliyan Zarov, Yuchen Zhang, Angela Fan, Melanie Kambadur, Sharan Narang, Aur{\'{e}}lien Rodriguez, Robert Stojnic, Sergey Edunov,
  and Thomas Scialom. 2023{\natexlab{b}}.
\newblock \href {https://doi.org/10.48550/ARXIV.2307.09288} {Llama 2: Open foundation and fine-tuned chat models}.
\newblock \emph{CoRR}, abs/2307.09288.

\bibitem[{Tunstall et~al.(2023)Tunstall, Beeching, Lambert, Rajani, Rasul, Belkada, Huang, von Werra, Fourrier, Habib, Sarrazin, Sanseviero, Rush, and Wolf}]{DBLP:journals/corr/TunstallBL23}
Lewis Tunstall, Edward Beeching, Nathan Lambert, Nazneen Rajani, Kashif Rasul, Younes Belkada, Shengyi Huang, Leandro von Werra, Clémentine Fourrier, Nathan Habib, Nathan Sarrazin, Omar Sanseviero, Alexander~M. Rush, and Thomas Wolf. 2023.
\newblock \href {https://arxiv.org/abs/2310.16944} {Zephyr: Direct distillation of lm alignment}.
\newblock \emph{CoRR}, abs/2310.16944.

\bibitem[{Vaswani et~al.(2017)Vaswani, Shazeer, Parmar, Uszkoreit, Jones, Gomez, Kaiser, and Polosukhin}]{DBLP:conf/nips/VaswaniSPUJGKP17}
Ashish Vaswani, Noam Shazeer, Niki Parmar, Jakob Uszkoreit, Llion Jones, Aidan~N. Gomez, Lukasz Kaiser, and Illia Polosukhin. 2017.
\newblock \href {https://proceedings.neurips.cc/paper/2017/hash/3f5ee243547dee91fbd053c1c4a845aa-Abstract.html} {Attention is all you need}.
\newblock In \emph{Advances in Neural Information Processing Systems 30: Annual Conference on Neural Information Processing Systems 2017, December 4-9, 2017, Long Beach, CA, {USA}}, pages 5998--6008.

\bibitem[{Wang et~al.(2021)Wang, Bao, Huang, Dong, and Wei}]{DBLP:conf/acl/WangBHDW21}
Wenhui Wang, Hangbo Bao, Shaohan Huang, Li~Dong, and Furu Wei. 2021.
\newblock \href {https://doi.org/10.18653/V1/2021.FINDINGS-ACL.188} {Minilmv2: Multi-head self-attention relation distillation for compressing pretrained transformers}.
\newblock In \emph{Findings of the Association for Computational Linguistics: {ACL/IJCNLP} 2021, Online Event, August 1-6, 2021}, volume {ACL/IJCNLP} 2021 of \emph{Findings of {ACL}}, pages 2140--2151. Association for Computational Linguistics.

\bibitem[{Wen et~al.(2023)Wen, Li, Du, and Mou}]{DBLP:conf/acl/Wen0DM23}
Yuqiao Wen, Zichao Li, Wenyu Du, and Lili Mou. 2023.
\newblock \href {https://doi.org/10.18653/V1/2023.ACL-LONG.605} {f-divergence minimization for sequence-level knowledge distillation}.
\newblock In \emph{Proceedings of the 61st Annual Meeting of the Association for Computational Linguistics (Volume 1: Long Papers), {ACL} 2023, Toronto, Canada, July 9-14, 2023}, pages 10817--10834. Association for Computational Linguistics.

\bibitem[{Wolf et~al.(2019)Wolf, Debut, Sanh, Chaumond, Delangue, Moi, Cistac, Rault, Louf, Funtowicz, and Brew}]{DBLP:journals/corr/WolfDS19}
Thomas Wolf, Lysandre Debut, Victor Sanh, Julien Chaumond, Clement Delangue, Anthony Moi, Pierric Cistac, Tim Rault, R{\'{e}}mi Louf, Morgan Funtowicz, and Jamie Brew. 2019.
\newblock \href {https://arxiv.org/abs/1910.03771} {Huggingface's transformers: State-of-the-art natural language processing}.
\newblock \emph{CoRR}, abs/1910.03771.

\bibitem[{Xia et~al.(2023)Xia, Gao, Zeng, and Chen}]{DBLP:journals/corr/XiaGZC23}
Mengzhou Xia, Tianyu Gao, Zhiyuan Zeng, and Danqi Chen. 2023.
\newblock \href {https://doi.org/10.48550/ARXIV.2310.06694} {Sheared llama: Accelerating language model pre-training via structured pruning}.
\newblock \emph{CoRR}, abs/2310.06694.

\bibitem[{Xia et~al.(2022)Xia, Zhong, and Chen}]{DBLP:conf/acl/XiaZC22}
Mengzhou Xia, Zexuan Zhong, and Danqi Chen. 2022.
\newblock \href {https://doi.org/10.18653/V1/2022.ACL-LONG.107} {Structured pruning learns compact and accurate models}.
\newblock In \emph{Proceedings of the 60th Annual Meeting of the Association for Computational Linguistics (Volume 1: Long Papers), {ACL} 2022, Dublin, Ireland, May 22-27, 2022}, pages 1513--1528. Association for Computational Linguistics.

\bibitem[{Xiao et~al.(2023)Xiao, Lin, Seznec, Wu, Demouth, and Han}]{DBLP:conf/icml/XiaoLSWDH23}
Guangxuan Xiao, Ji~Lin, Micka{\"{e}}l Seznec, Hao Wu, Julien Demouth, and Song Han. 2023.
\newblock \href {https://proceedings.mlr.press/v202/xiao23c.html} {Smoothquant: Accurate and efficient post-training quantization for large language models}.
\newblock In \emph{International Conference on Machine Learning, {ICML} 2023, 23-29 July 2023, Honolulu, Hawaii, {USA}}, volume 202 of \emph{Proceedings of Machine Learning Research}, pages 38087--38099. {PMLR}.

\bibitem[{Xie et~al.(2023)Xie, Pham, Dong, Du, Liu, Lu, Liang, Le, Ma, and Yu}]{DBLP:journals/corr/XiePD23}
Sang~Michael Xie, Hieu Pham, Xuanyi Dong, Nan Du, Hanxiao Liu, Yifeng Lu, Percy Liang, Quoc~V. Le, Tengyu Ma, and Adams~Wei Yu. 2023.
\newblock \href {https://doi.org/10.48550/ARXIV.2305.10429} {Doremi: Optimizing data mixtures speeds up language model pretraining}.
\newblock \emph{CoRR}, abs/2305.10429.

\bibitem[{Xin et~al.(2020)Xin, Tang, Lee, Yu, and Lin}]{DBLP:conf/acl/XinTLYL20}
Ji~Xin, Raphael Tang, Jaejun Lee, Yaoliang Yu, and Jimmy Lin. 2020.
\newblock \href {https://doi.org/10.18653/V1/2020.ACL-MAIN.204} {Deebert: Dynamic early exiting for accelerating {BERT} inference}.
\newblock In \emph{Proceedings of the 58th Annual Meeting of the Association for Computational Linguistics, {ACL} 2020, Online, July 5-10, 2020}, pages 2246--2251. Association for Computational Linguistics.

\bibitem[{Xu et~al.(2023{\natexlab{a}})Xu, Sun, Zheng, Geng, Zhao, Feng, Tao, and Jiang}]{DBLP:journals/corr/XuSZ23}
Can Xu, Qingfeng Sun, Kai Zheng, Xiubo Geng, Pu~Zhao, Jiazhan Feng, Chongyang Tao, and Daxin Jiang. 2023{\natexlab{a}}.
\newblock \href {https://doi.org/10.48550/ARXIV.2304.12244} {Wizardlm: Empowering large language models to follow complex instructions}.
\newblock \emph{CoRR}, abs/2304.12244.

\bibitem[{Xu et~al.(2023{\natexlab{b}})Xu, Guo, Duan, and McAuley}]{DBLP:journals/corr/XuGD23}
Canwen Xu, Daya Guo, Nan Duan, and Julian~J. McAuley. 2023{\natexlab{b}}.
\newblock \href {https://doi.org/10.48550/ARXIV.2304.01196} {Baize: An open-source chat model with parameter-efficient tuning on self-chat data}.
\newblock \emph{CoRR}, abs/2304.01196.

\bibitem[{Yang et~al.(2023)Yang, Xiao, Wang, Zhang, Bian, Yin, Lv, Pan, Wang, Yan, Yang, Deng, Wang, Liu, Ai, Dong, Zhao, Xu, Sun, Zhang, Liu, Ji, Xie, Dai, Fang, Su, Song, Liu, Ru, Ma, Wang, Liu, Lin, Nie, Guo, Sun, Zhang, Li, Li, Cheng, Chen, Zeng, Wang, Chen, Men, Yu, Pan, Shen, Wang, Li, Jiang, Gao, Zhang, Zhou, and Wu}]{DBLP:journals/corr/YangXW23}
Aiyuan Yang, Bin Xiao, Bingning Wang, Borong Zhang, Ce~Bian, Chao Yin, Chenxu Lv, Da~Pan, Dian Wang, Dong Yan, Fan Yang, Fei Deng, Feng Wang, Feng Liu, Guangwei Ai, Guosheng Dong, Haizhou Zhao, Hang Xu, Haoze Sun, Hongda Zhang, Hui Liu, Jiaming Ji, Jian Xie, Juntao Dai, Kun Fang, Lei Su, Liang Song, Lifeng Liu, Liyun Ru, Luyao Ma, Mang Wang, Mickel Liu, MingAn Lin, Nuolan Nie, Peidong Guo, Ruiyang Sun, Tao Zhang, Tianpeng Li, Tianyu Li, Wei Cheng, Weipeng Chen, Xiangrong Zeng, Xiaochuan Wang, Xiaoxi Chen, Xin Men, Xin Yu, Xuehai Pan, Yanjun Shen, Yiding Wang, Yiyu Li, Youxin Jiang, Yuchen Gao, Yupeng Zhang, Zenan Zhou, and Zhiying Wu. 2023.
\newblock \href {https://doi.org/10.48550/ARXIV.2309.10305} {Baichuan 2: Open large-scale language models}.
\newblock \emph{CoRR}, abs/2309.10305.

\bibitem[{Yang et~al.(2022)Yang, Zhang, and Song}]{DBLP:conf/emnlp/YangZS22}
Yi~Yang, Chen Zhang, and Dawei Song. 2022.
\newblock \href {https://doi.org/10.18653/V1/2022.EMNLP-MAIN.258} {Sparse teachers can be dense with knowledge}.
\newblock In \emph{Proceedings of the 2022 Conference on Empirical Methods in Natural Language Processing, {EMNLP} 2022, Abu Dhabi, United Arab Emirates, December 7-11, 2022}, pages 3904--3915. Association for Computational Linguistics.

\bibitem[{Yuan et~al.(2021)Yuan, Zhao, Du, Ding, Liu, Cen, Zou, Yang, and Tang}]{DBLP:journals/aiopen/YuanZDDLCZYT21}
Sha Yuan, Hanyu Zhao, Zhengxiao Du, Ming Ding, Xiao Liu, Yukuo Cen, Xu~Zou, Zhilin Yang, and Jie Tang. 2021.
\newblock \href {https://doi.org/10.1016/J.AIOPEN.2021.06.001} {Wudaocorpora: {A} super large-scale chinese corpora for pre-training language models}.
\newblock \emph{{AI} Open}, 2:65--68.

\bibitem[{Zhang et~al.(2023{\natexlab{a}})Zhang, Wang, and Song}]{DBLP:journals/corr/ZhangWS23}
Chen Zhang, Benyou Wang, and Dawei Song. 2023{\natexlab{a}}.
\newblock \href {https://doi.org/10.48550/ARXIV.2311.07204} {On elastic language models}.
\newblock \emph{CoRR}, abs/2311.07204.

\bibitem[{Zhang et~al.(2023{\natexlab{b}})Zhang, Yang, Liu, Wang, Xian, Wang, and Song}]{DBLP:conf/acl/ZhangYLWXWS23}
Chen Zhang, Yang Yang, Jiahao Liu, Jingang Wang, Yunsen Xian, Benyou Wang, and Dawei Song. 2023{\natexlab{b}}.
\newblock \href {https://doi.org/10.18653/V1/2023.ACL-LONG.249} {Lifting the curse of capacity gap in distilling language models}.
\newblock In \emph{Proceedings of the 61st Annual Meeting of the Association for Computational Linguistics (Volume 1: Long Papers), {ACL} 2023, Toronto, Canada, July 9-14, 2023}, pages 4535--4553. Association for Computational Linguistics.

\bibitem[{Zhang et~al.(2023{\natexlab{c}})Zhang, Yang, Wang, and Song}]{DBLP:journals/corr/ZhangYWS23}
Chen Zhang, Yang Yang, Jingang Wang, and Dawei Song. 2023{\natexlab{c}}.
\newblock \href {https://doi.org/10.48550/ARXIV.2305.12330} {Task-agnostic distillation of encoder-decoder language models}.
\newblock \emph{CoRR}, abs/2305.12330.

\bibitem[{Zhang et~al.(2022{\natexlab{a}})Zhang, Yang, Wang, Liu, Wang, Wu, and Song}]{DBLP:journals/corr/ZhangYW22}
Chen Zhang, Yang Yang, Qifan Wang, Jiahao Liu, Jingang Wang, Wei Wu, and Dawei Song. 2022{\natexlab{a}}.
\newblock \href {https://doi.org/10.48550/ARXIV.2205.14570} {Minidisc: Minimal distillation schedule for language model compression}.
\newblock \emph{CoRR}, abs/2205.14570.

\bibitem[{Zhang et~al.(2023{\natexlab{d}})Zhang, Shi, Liu, Yuan, Li, Dong, Shu, Li, Wang, Lin, Huang, and Fu}]{DBLP:journals/corr/ZhangSL23}
Ge~Zhang, Yemin Shi, Ruibo Liu, Ruibin Yuan, Yizhi Li, Siwei Dong, Yu~Shu, Zhaoqun Li, Zekun Wang, Chenghua Lin, Wenhao Huang, and Jie Fu. 2023{\natexlab{d}}.
\newblock \href {https://doi.org/10.48550/ARXIV.2304.07987} {Chinese open instruction generalist: {A} preliminary release}.
\newblock \emph{CoRR}, abs/2304.07987.

\bibitem[{Zhang et~al.(2023{\natexlab{e}})Zhang, Zeng, Wang, and Lu}]{TinyLLaMA23}
Peiyuan Zhang, Guangtao Zeng, Tianduo Wang, and Wei Lu. 2023{\natexlab{e}}.
\newblock \href {https://github.com/jzhang38/TinyLlama} {Tinyllama}.

\bibitem[{Zhang et~al.(2022{\natexlab{b}})Zhang, Roller, Goyal, Artetxe, Chen, Chen, Dewan, Diab, Li, Lin, Mihaylov, Ott, Shleifer, Shuster, Simig, Koura, Sridhar, Wang, and Zettlemoyer}]{DBLP:journals/corr/ZhangRG22}
Susan Zhang, Stephen Roller, Naman Goyal, Mikel Artetxe, Moya Chen, Shuohui Chen, Christopher Dewan, Mona~T. Diab, Xian Li, Xi~Victoria Lin, Todor Mihaylov, Myle Ott, Sam Shleifer, Kurt Shuster, Daniel Simig, Punit~Singh Koura, Anjali Sridhar, Tianlu Wang, and Luke Zettlemoyer. 2022{\natexlab{b}}.
\newblock \href {https://doi.org/10.48550/ARXIV.2205.01068} {{OPT:} open pre-trained transformer language models}.
\newblock \emph{CoRR}, abs/2205.01068.

\bibitem[{Zhang et~al.(2022{\natexlab{c}})Zhang, Lin, Liu, Li, Sun, and Zhou}]{DBLP:conf/acl/ZhangL00S022}
Zhengyan Zhang, Yankai Lin, Zhiyuan Liu, Peng Li, Maosong Sun, and Jie Zhou. 2022{\natexlab{c}}.
\newblock \href {https://doi.org/10.18653/V1/2022.FINDINGS-ACL.71} {Moefication: Transformer feed-forward layers are mixtures of experts}.
\newblock In \emph{Findings of the Association for Computational Linguistics: {ACL} 2022, Dublin, Ireland, May 22-27, 2022}, pages 877--890. Association for Computational Linguistics.

\bibitem[{Zheng et~al.(2023)Zheng, Chiang, Sheng, Zhuang, Wu, Zhuang, Lin, Li, Li, Xing, Zhang, Gonzalez, and Stoica}]{DBLP:journals/corr/ZhengCS23}
Lianmin Zheng, Wei{-}Lin Chiang, Ying Sheng, Siyuan Zhuang, Zhanghao Wu, Yonghao Zhuang, Zi~Lin, Zhuohan Li, Dacheng Li, Eric~P. Xing, Hao Zhang, Joseph~E. Gonzalez, and Ion Stoica. 2023.
\newblock \href {https://doi.org/10.48550/ARXIV.2306.05685} {Judging llm-as-a-judge with mt-bench and chatbot arena}.
\newblock \emph{CoRR}, abs/2306.05685.

\bibitem[{ZhipuAI(2023{\natexlab{a}})}]{ChatGLM23}
ZhipuAI. 2023{\natexlab{a}}.
\newblock \href {https://github.com/THUDM/ChatGLM-6B} {Chatglm-6b: An open bilingual dialogue language model}.

\bibitem[{ZhipuAI(2023{\natexlab{b}})}]{ChatGLM223}
ZhipuAI. 2023{\natexlab{b}}.
\newblock \href {https://github.com/THUDM/ChatGLM2-6B} {Chatglm2-6b: An open bilingual chat llm}.

\bibitem[{Zhou et~al.(2022)Zhou, Xu, and McAuley}]{DBLP:conf/acl/ZhouXM22}
Wangchunshu Zhou, Canwen Xu, and Julian~J. McAuley. 2022.
\newblock \href {https://doi.org/10.18653/V1/2022.ACL-LONG.485} {{BERT} learns to teach: Knowledge distillation with meta learning}.
\newblock In \emph{Proceedings of the 60th Annual Meeting of the Association for Computational Linguistics (Volume 1: Long Papers), {ACL} 2022, Dublin, Ireland, May 22-27, 2022}, pages 7037--7049. Association for Computational Linguistics.

\bibitem[{Zhu et~al.(2023)Zhu, Li, Liu, Ma, and Wang}]{DBLP:journals/corr/ZhuLL23}
Xunyu Zhu, Jian Li, Yong Liu, Can Ma, and Weiping Wang. 2023.
\newblock \href {https://doi.org/10.48550/ARXIV.2308.07633} {A survey on model compression for large language models}.
\newblock \emph{CoRR}, abs/2308.07633.

\bibitem[{Zuo et~al.(2022)Zuo, Zhang, Liang, He, Zhao, and Chen}]{DBLP:conf/naacl/ZuoZLHZC22}
Simiao Zuo, Qingru Zhang, Chen Liang, Pengcheng He, Tuo Zhao, and Weizhu Chen. 2022.
\newblock \href {https://doi.org/10.18653/V1/2022.NAACL-MAIN.116} {Moebert: from {BERT} to mixture-of-experts via importance-guided adaptation}.
\newblock In \emph{Proceedings of the 2022 Conference of the North American Chapter of the Association for Computational Linguistics: Human Language Technologies, {NAACL} 2022, Seattle, WA, United States, July 10-15, 2022}, pages 1610--1623. Association for Computational Linguistics.

\end{thebibliography}

\appendix

\section{Adaptation of LLaMA2}
\label{app:llama_adaptation}

It is well identified that LLaMA and LLaMA2 are weak in Chinese language understanding and generation~\citep{DBLP:journals/corr/CuiYY23}, impeding the distillation from LLaMA to \textsc{MiniMA} that could potentially be applied to Chinese scenarios. The core rationales underlying the sub-optimal performance of LLaMA2 on Chinese can be attributed to two parts, where the first part is that LLaMA2 is not trained on many Chinese tokens and the second part is that LLaMA2 is not trained with adequate Chinese vocabulary. While the former one could be counted by adaptation, the latter should be handled by expanding the original vocabulary with a Chinese vocabulary. 

At the time of formulating this work, it is the first option to do so since there are not any available bilingual language models with better performance than LLaMA2. And nowadays, many great open-source bilingual language models (\citealp[e.g., Baichuan2,][]{DBLP:journals/corr/YangXW23};~\citealp[Qwen,][]{DBLP:journals/corr/BaiBC23}) have emerged that can be utilized without the necessity of adaptation.

\begin{table}[ht]
    \centering
    \caption{The token per byte compression rates. The results are measured on some English and Chinese Wikipedia articles.}
    \begin{adjustbox}{width=0.47\textwidth,center}
    \begin{tabular}{lcc}
    \toprule
        \textbf{LM} & \textbf{Vocabulary} & \textbf{Token/Byte Compression Rate} \\
    \midrule
        LLaMA2 & 32,000 & 46.2\% \\
        \textsc{MiniMA} & 49,216 & 35.0\% \\
    \bottomrule
    \end{tabular}
    \end{adjustbox}
    \label{tab:minima_byte_comp_rate}
\end{table}

Thereby, before the adaptation, we append many frequently used Chinese characters and phrases in SentencePiece~\citep{DBLP:conf/emnlp/KudoR18} so that the original encoding and decoding behaviors of English are not changed while those of Chinese are improved. In doing so, we expand the vocabulary of LLaMA2 from 32,000 to 49,216 with much better Chinese encoding and decoding performance in terms of token per byte compression rate as in Table~\ref{tab:minima_byte_comp_rate}. With the purpose of making LLaMA2 familiar with more Chinese tokens including those appended yet not learned elements in the new vocabulary, we should necessarily adapt the expanded LLaMA2 on curated data as in Table~\ref{tab:minima_data_stat}. And after the adaptation, the English perplexity is not declined that much while the Chinese perplexity is also considerable as in Table~\ref{tab:minima_perplexity}. The discrepancy in Chinese perplexity is perhaps due to that LLaMA2 may fallback to bytes for unknown Chinese characters and adapted LLaMA2 is still under-trained.

\begin{table}[ht]
    \centering
    \caption{The perplexities. The results are measured on some English and Chinese Wikipedia articles.}
    \begin{adjustbox}{width=0.35\textwidth,center}
    \begin{tabular}{lcc}
    \toprule
        \textbf{LM} & \textbf{English} & \textbf{Chinese} \\
    \midrule
        LLaMA2-7B & 3.74 & 3.96 \\
        adapted LLaMA2-7B & 3.98 & 13.91 \\
    \bottomrule
    \end{tabular}
    \end{adjustbox}
    \label{tab:minima_perplexity}
\end{table}

The data for adaptation is packed into sequences, and each of which is of 4,096 tokens in alignment with that of LLaMA2. We take a batch size of $\sim$4M tokens (or say 1,024 batches), a learning rate of 3e-5, a weight decay of 1e-1. The training lasts for 1 epoch, and the learning rate is scheduled to warm up linearly for the first 1\% steps of all and decay down in a sinusoidal way for the rest steps. Gradients whose norms accumulate over 1 are necessarily clipped. The training is executed on 16 A100 80G GPUs. The training efficiency is sufficiently guaranteed by DeepSpeed Zero3 this time and Flash Attention. Gradient checkpointing is enabled to further reduce the memory footprint and bfloat16 precision is chosen to improve the training stability.

\begin{table}[ht]
    \centering
    \caption{The statistics of data mixture used for finetuning.}
    \begin{adjustbox}{width=0.5\textwidth,center}
    \begin{tabular}{lcrr}
    \toprule
        \textbf{Dataset} & \textbf{Language} & \textbf{Examples} & \textbf{Proportion} \\
    \midrule
        Alpaca & En & 52,002 & 4.7\% \\
        Alpaca-Chinese & Zh & 48,818 & 4.4\% \\
        CodeAlpaca & En \& Co & 20,022 & 1.8\% \\
        GPTeacher-Instruct & En & 18,194 & 1.7\% \\
        GPTeacher-RolePlay & En & 1,923 & 0.2\% \\
        GPTeacher-CodeGen & En \& Co & 4,535 & 0.4\% \\
        Baize-Quora & En & 54,456 & 4.9\% \\
        Baize-StackOverflow & En & 57,046 & 5.2\% \\
        UnnaturalInstruction & En & 9,000 & 0.8\% \\
        Flan-CoT & En & 74,771 & 6.8\% \\
        Flan-CoT-Chinese & Zh & 74,771 & 6.8\% \\
        COIG-LeetCode & Zh \& Co & 11,737 & 1.1\% \\
        BELLE-Math & Zh & 49,696 & 4.5\% \\
        ShareGPT/Discord & En & 197,893 & 18.0\% \\
        MiniChat-Zhihu & Zh & 129,981 & 11.8\% \\
        MiniChat-Wiki & Zh & 234,291 & 21.3\% \\
        MiniChat-Math & Zh & 39,336 & 3.6\% \\ 
        MiniChat-Life & Zh & 22,697 & 2.1\% \\
    \midrule
        Mixture & -- & 1,101,169 & 100.0\% \\
    \bottomrule
    \end{tabular}
    \end{adjustbox}
    \label{tab:minichat_data_stat}
\end{table}

\section{Statistics of Instruction-following Data}
\label{app:stat_inst_data}

The statistics are presented in Table~\ref{tab:minichat_data_stat}, accommodating data gathered from different languages and categories, such as Alpaca~\citep{Alpaca23}, CodeAlpaca~\citep{CodeAlpaca23}, GPTeacher~\citep{GPTeacher23}, Baize~\citep{DBLP:journals/corr/XuGD23}, Unnatural Instruction~\citep{DBLP:conf/acl/HonovichSLS23}, Flan-CoT~\citep{DBLP:journals/corr/ChungHL22}, COIG~\citep{DBLP:journals/corr/ZhangSL23}, and ShareGPT.

In regard to lacking Chinese data, we additionally incorporate data generated from ChatGPT following the \textit{self-chat} technique~\citep{DBLP:journals/corr/XuGD23}, which necessarily provides seed topics to ChatGPT and has it generate chat data fluently. The seed topics in Chinese we use range from Zhihu trending questions and Wikipedia entries to Math questions and Life news. 

\section{Pruning of \textsc{MiniMA}}
\label{app:minima_pruning}

The motivation of using a heuristic pruning priority for \textsc{MiniMA} is that asymmetric shapes across layers need quite a few monkey-patches to widely used libraries (\citealp[e.g., HuggingFace Transformers,][]{DBLP:journals/corr/WolfDS19}) in spite of their slightly superior performance. In contrast, symmetric shaping naturally fits in these libraries, thus flexible for widespread use.

And the reasons pertain behind the design choice for symmetric shaping are as follows. The local pruning priority is basically required by the symmetric shaping. However, the global pruning priority can lead to entire-layer pruning yet the local one cannot. So we introduce a heuristic rule here that distributes a portion of the sparsity to layer dropping and consequently reduces the sparsity of hidden states, MHA heads, and FFN neurons.

\begin{figure}[ht]
    \centering
    \includegraphics[width=0.5\textwidth]{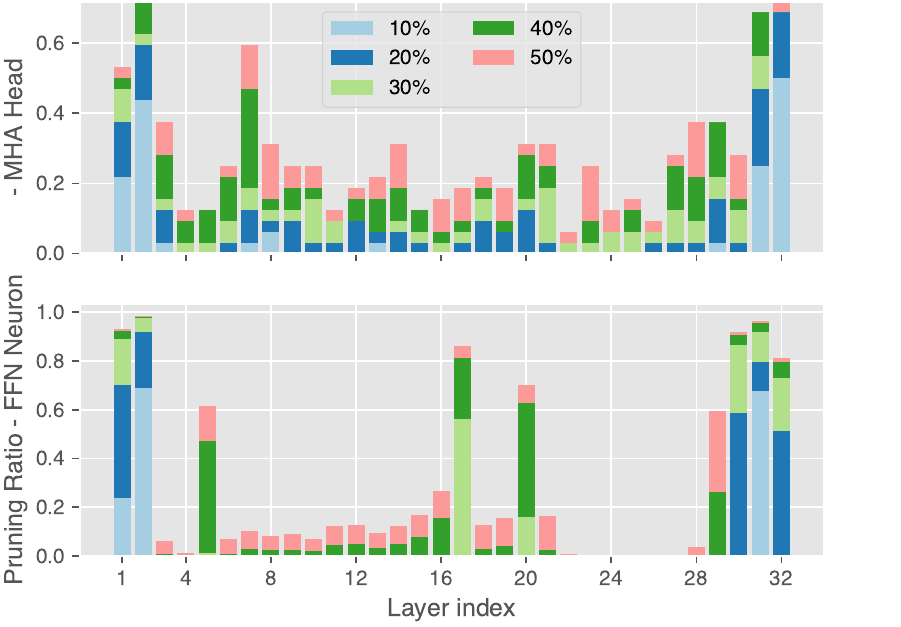}
    \caption{The pruning ratios across layers of LLaMA2-7B along the rise of sparsity. The variations of pruning ratios of MHA heads and FFN neurons are listed separately. Peaks and valleys are obvious to see, and peaks stand at the most bottom and top layers.}
    \label{fig:minima_pruning_stat}
\end{figure}

As for justification on why evenly dropping the most bottom and top layers rather than other layers, we uncover that the global pruning priority tends to prune the most bottom and top MHA heads and FFN neurons as in Figure~\ref{fig:minima_pruning_stat} even with the mentioned $\ell_{2}$ norm protection from skewed pruning ratios~\citep{DBLP:conf/iclr/MolchanovTKAK17}. This suggests that the most bottom and top layers of LLaMA are confidently of least significance among all layers, encouraging us to prune them in the first place. And the even distribution is just an approximation.

\begin{table*}[ht]
    \centering
    \caption{The results against existing LMs on standard benchmarks.}
    \begin{adjustbox}{width=0.87\textwidth,center}
    \begin{tabular}{lrcccccc}
    \toprule
        \textbf{LM} & \textbf{Tokens} & \makecell[c]{\textbf{MMLU}\\\textbf{Acc}} & \makecell[c]{\textbf{CEval}\\\textbf{Acc}} & \makecell[c]{\textbf{DROP}\\\textbf{EM Score}} & \makecell[c]{\textbf{BBH}\\\textbf{EM Score}} & \makecell[c]{\textbf{GSM8K}\\\textbf{Maj1@1}} & \makecell[c]{\textbf{HumanEval}\\\textbf{Pass@1}} \\
    \midrule
        LLaMA-7B~\citeyearpar{DBLP:journals/corr/TouvronLI23} & 1,000 B & 35.10 & 28.00 & 27.46 & 30.93 & 9.17 & 10.37 \\
        LLaMA2-7B~\citeyearpar{DBLP:journals/corr/TouvronMS23} & 2,000 B & 46.00 & 34.40 & 31.57 & 32.02 & 14.10 & 12.80 \\
        Baichuan-7B~\citeyearpar{Baichuan123} & 1,200 B & 42.60 & 43.50 & 19.82 & 31.94 & 8.57 & 7.93 \\
        Baichuan2-7B~\citeyearpar{DBLP:journals/corr/YangXW23} & 2,600 B & 54.31 & 55.27 & 25.97 & 35.21 & 13.19 & 16.46 \\
        Mistral-7B~\citeyearpar{DBLP:journals/corr/JiangSM23} & -- & 62.67 & 45.91 & 46.59 & 43.88 & 41.02 & 28.05 \\
    \midrule
        Mamba-2.8B~\citeyearpar{DBLP:journals/corr/GuD23} & 300 B & 25.58 & 24.74 & 15.72 & 29.37 & 3.49 & 7.32 \\
        ShearedLLaMA-2.7B~\citeyearpar{DBLP:journals/corr/XiaGZC23} & 50 B & 26.97 & 22.88 & 19.98 & 30.48 & 3.56 & 4.88 \\
        CerebrasGPT-2.7B~\citeyearpar{DBLP:journals/corr/DeyGC23} & 53 B & 24.66 & 23.18 & 11.46 & 29.32 & 2.43 & 3.66 \\
        OPT-2.7B~\citeyearpar{DBLP:journals/corr/ZhangRG22} & 180 B & 26.02 & 24.52 & 13.70 & 28.71 & 1.90 & 0.00 \\
        BLOOM-3B~\citeyearpar{DBLP:journals/corr/ScaoFA22} & 341 B & 26.60 & 23.77 & 14.32 & 29.84 & 2.04 & 0.61 \\
        Pythia-2.8B~\citeyearpar{DBLP:conf/icml/BidermanSABOHKP23} & 300 B & 26.28 & 23.11 & 16.04 & 29.30 & 2.73 & 5.49 \\
        OpenLLaMA-3B~\citeyearpar{OpenLLaMA23} & 1,000 B & 26.70 & 26.30 & 20.14 & 30.56 & 3.11 & 0.00 \\
        OpenLLaMAv2-3B~\citeyearpar{OpenLLaMA23} & 1,000 B & 26.36 & 25.41 & 18.19 & 30.45 & 4.62 & 7.93 \\
        BTLM-3B~\citeyearpar{DBLP:journals/corr/DeySA23} & 627 B & 27.20 & 26.00 & 17.84 & 30.87 & 4.55 & 10.98 \\
        StableLM-3B~\citeyearpar{StableLM23} & 4,000 B & 44.75 & 31.05 & 22.35 & 32.59 & 10.99 & 15.85 \\
    \midrule
        Mamba-1.4B~\citeyearpar{DBLP:journals/corr/GuD23} & 300 B & 25.97 & 25.85 & 12.36 & 29.46 & 1.67 & 6.71 \\
        ShearedLLaMA-1.3B~\citeyearpar{DBLP:journals/corr/XiaGZC23} & 50 B & 25.72 & 24.22 & 14.53 & 29.22 & 3.03 & 1.83 \\
        CerebrasGPT-1.3B~\citeyearpar{DBLP:journals/corr/DeyGC23} & 26 B & 26.50 & 23.03 & 8.54 & 29.11 & 2.58 & 2.44 \\
        OPT-1.3B~\citeyearpar{DBLP:journals/corr/ZhangRG22} & 180 B & 26.66 & 25.71 & 14.38 & 28.77 & 3.18 & 0.00 \\
        BLOOM-1.1B~\citeyearpar{DBLP:journals/corr/ScaoFA22} & 341 B & 27.33 & 26.75 & 8.62 & 28.42 & 2.43 & 0.00 \\
        Pythia-1.4B~\citeyearpar{DBLP:conf/icml/BidermanSABOHKP23} & 300 B & 25.20 & 26.08 & 12.22 & 28.87 & 1.90 & 5.49 \\
        TinyLLaMA-1.1B~\citeyearpar{TinyLLaMA23} & 2,500 B & 25.84 & 25.19 & 15.42 & 29.35 & 2.12 & 7.93 \\
        Falcon-1B~\citeyearpar{DBLP:journals/corr/PenedoMH23} & 350 B & 25.52 & 25.93 & 15.94 & 27.71 & 1.67 & 0.00 \\
        Phi1.5-1.3B~\citeyearpar{DBLP:journals/corr/LiBE23} & 150 B & 42.86 & 26.60 & 14.94 & 30.84 & 28.43 & 32.93 \\
        Qwen-1.8B~\citeyearpar{DBLP:journals/corr/BaiBC23} & 2,200 B & 44.05 & 54.75 & 12.97 & 30.80 & 22.97 & 14.02 \\
    \midrule
        \textsc{MiniMA} & 126 B & 28.51 & 28.23 & 22.50 & 31.61 & 8.11 & 10.98 \\
    \bottomrule
    \end{tabular}
    \end{adjustbox}
    \label{tab:bench_lm}
\end{table*}

\section{Evaluation Protocols of Standard Benchmarks}
\label{app:bench_eval_proto}

MMLU~\citep{DBLP:conf/iclr/HendrycksBBZMSS21} is an multiple-choice dataset covering multiple disciplines. CEval~\citep{DBLP:journals/corr/HuangBZ23} is also a multiple-choice dataset covering multiple disciplines however in Chinese, which could be viewed as a Chinese MMLU. DROP~\citep{DBLP:conf/naacl/DuaWDSS019} is a reading comprehension dataset requiring discrete reasoning. BBH~\citep{DBLP:conf/acl/SuzgunSSGTCCLCZ23} is the challenging counterpart of BIG-Bench~\citep{DBLP:journals/corr/SrivastavaRR22}, which is essentially a dataset measuring the reasoning abilities of LMs. GSM8K~\citep{DBLP:journals/corr/CobbeKB21} is a grade-school level math problem-solving dataset. HumanEval~\citep{DBLP:journals/corr/ChenTJ21} is a coding problem-solving dataset. By category, MMLU and CEval are taken to evaluate the knowledge of LMs, DROP, BBH, GSM8K are taken to evaluate the reasoning of LMs, and HumanEval is taken to evaluate the coding of LMs.

Following the evaluation guidelines of LLaMA~\citep{DBLP:journals/corr/TouvronLI23,DBLP:journals/corr/TouvronMS23}, we devise evaluation protocols of considered datasets as follows:
\begin{itemize}
    \item MMLU: 5-shot direct prompting performance evaluated by accuracy.
    \item CEval: 5-shot direct prompting performance evaluated by accuracy.
    \item DROP: 3-shot direct prompting performance evaluated by exact match score.
    \item BBH: 3-shot direct prompting performance evaluated by exact match score.
    \item GSM8K: 8-shot chain-of-thought prompting performance evaluated by maj1@1.
    \item HumanEval: 0-shot prompting performance evaluated by pass@1.
\end{itemize}
Here, accuracy and extach match score are two commonly seen metrics. maj1@k is a specialized metric that gives a positive judgement if there is at least 1 correct answer among k candidate answers output by LMs. pass@1 is also a specialized one that shares similar idea with maj1@k yet with a much more complex formulation.

We report the results on validation sets of CEval and DROP due to the unavailability of their test sets and the results on test sets on the others.

\begin{figure}[ht]
    \centering
    \includegraphics[width=0.5\textwidth]{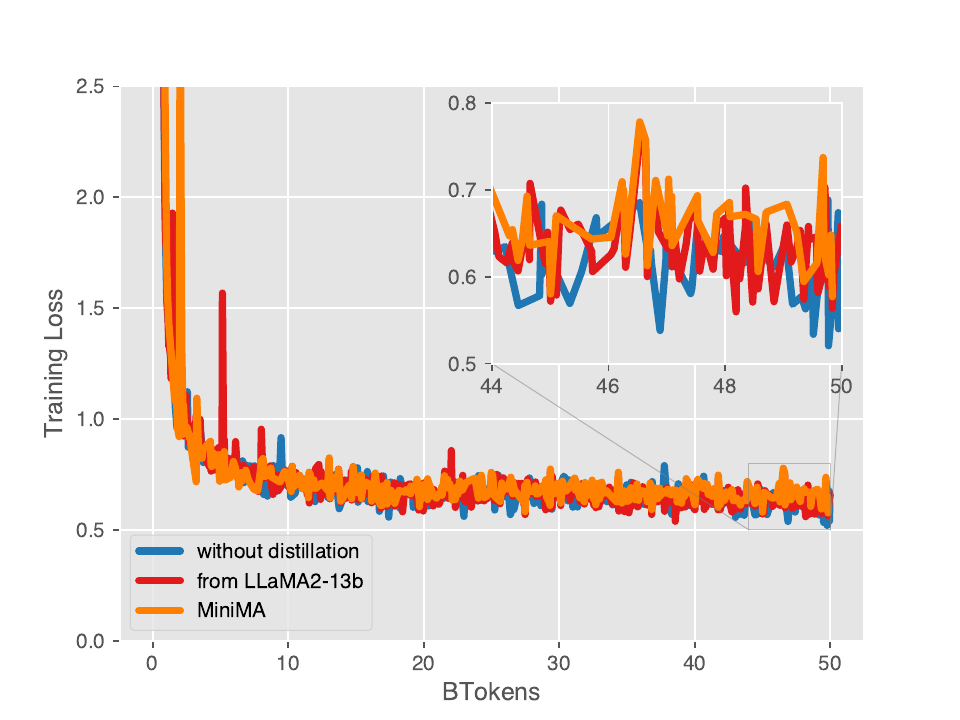}
    \caption{The training losses of \textsc{MiniMA} versus its variants.}
    \label{fig:minima_train_loss}
\end{figure}

\section{Training Loss of \textsc{MiniMA}}
\label{app:minima_train_loss}

Training loss is usually taken as a sort of criterion of convergence and can signal the performance of a LM in pretraining as in~\citet{DBLP:journals/corr/TouvronLI23}. So we conduct training loss-level comparisons among \textsc{MiniMA} and its variants.

However, the signal is not valid anymore in distillation since \textsc{MiniMA} and its variants have very similar training losses as in Figure~\ref{fig:minima_train_loss}. Direct training has a slightly lower training loss than distillation from LLaMA2-13B, and \textsc{MiniMA} actually owns a slightly higher training loss than distillation from LLaMA2-13B does. We conjecture that distillation loss (alignment to distributed probabilities) is inherently more difficult to optimize than direct training loss (alignment to one-hot labels). In literature~\citep{DBLP:conf/iclr/PereyraTCKH17}, larger models would confidently produce distributed probabilities that are more akin to one-hot labels. So it is no wonder even distillation from LLaMA2-13B can has a lower training loss than \textsc{MiniMA}.

\section{Benchmarking Results against Existing LMs}
\label{app:bench_lm}

We include the results of existing LMs here.

The results in Table~\ref{tab:bench_lm} present that \textsc{MiniMA} can with no doubt overshadow a large part of existing LMs. Reversely, some LMs beat other scale-matched LMs and \textsc{MiniMA} by substantial margins on some datasets thanks to neat data quality or huge data quantity, which initiates additional expenses. Some results on other related standard benchmarks are detailed in Appendix~\ref{app:more_standard}.

\section{More Standard Benchmarks}
\label{app:more_standard}

We additionally incorporate results on other tasks for a broader comparison including ARC~\citep{DBLP:journals/corr/ClarkCE18}, Winogrande~\citep{DBLP:conf/aaai/SakaguchiBBC20}, TruthfulQA~\citep{DBLP:conf/acl/LinHE22}. The performance on these tasks is evaluated by accuracy with 25-shot direct prompting, by accuracy with 5-shot direct prompting, and by mc2 with 0-shot direct prompting, respectively.

\begin{table}[ht]
    \centering
    \caption{The results on other standard benchmarks.}
    \begin{adjustbox}{width=0.50\textwidth,center}
    \begin{tabular}{lccc}
    \toprule
        \textbf{LM} & \makecell[c]{\textbf{ARC}\\\textbf{Acc}} & \makecell[c]{\textbf{Winogrande}\\\textbf{Acc}} & \makecell[c]{\textbf{TruthfulQA}\\\textbf{MC2}} \\
    \midrule
        OpenLLaMAv2-3B~\citeyearpar{OpenLLaMA23} & 40.27 & 67.01 & 34.78 \\
        ShearedLLaMA-2.7B~\citeyearpar{DBLP:journals/corr/XiaGZC23} & 41.72 & 67.01 & 37.32 \\
    \midrule
        \textsc{MiniMA} & 43.43 & 65.98 & 39.76 \\
    \bottomrule
    \end{tabular}
    \end{adjustbox}
    \label{tab:more_standard}
\end{table}

\begin{table}[ht]
    \centering
    \caption{The estimates of trainig compute of considered LMs.}
    \begin{adjustbox}{width=0.49\textwidth,center}
    \begin{threeparttable}
    \begin{tabular}{lrrr}
    \toprule
        \textbf{LM} & \textbf{Params} & \textbf{Tokens} & \textbf{Compute} \\
    \midrule
        LLaMA-7B & 7 B & 1,000 B & 42.0$\times$10\textsuperscript{9} TFLOPs \\
        LLaMA2-7B & 7 B & 2,000 B & 84.0$\times$10\textsuperscript{9} TFLOPs \\
        Baichuan-7B & 7 B & 1,200 B & 50.4$\times$10\textsuperscript{9} TFLOPs \\
        Baichuan2-7B & 7 B & 2,600 B & 109.2$\times$10\textsuperscript{9} TFLOPs \\
        Mistral-7B~\tnote{\ding{192}} & 7 B & 4,000 B & 168.0$\times$10\textsuperscript{9} TFLOPs \\
        Mixtral-8x7B~\tnote{\ding{192}} & 47 B & 4,000 B & 336.0$\times$10\textsuperscript{9} TFLOPs \\
    \midrule
        Mamba-2.8B~\tnote{\ding{193}} & 2.8 B & 300 B & 4.6$\times$10\textsuperscript{9} TFLOPs \\
        ShearedLLaMA-2.7B & 2.7 B & 50 B & 0.8$\times$10\textsuperscript{9} TFLOPs \\
        CerebrasGPT-2.7B & 2.7 B & 53 B & 0.9$\times$10\textsuperscript{9} TFLOPs \\
        OPT-2.7B & 2.7 B & 180 B & 2.9$\times$10\textsuperscript{9} TFLOPs \\
        BLOOM-3B & 3 B & 341 B & 6.1$\times$10\textsuperscript{9} TFLOPs \\
        Pythia-2.8B & 2.8 B & 300 B & 5.0$\times$10\textsuperscript{9} TFLOPs \\
        OpenLLaMA-3B & 3 B & 1,000 B & 18.0$\times$10\textsuperscript{9} TFLOPs \\
        OpenLLaMAv2-3B & 3 B & 1,000 B & 18.0$\times$10\textsuperscript{9} TFLOPs \\
        BTLM-3B & 3 B & 627 B & 11.2$\times$10\textsuperscript{9} TFLOPs \\
        StableLM-3B & 3 B & 4,000 B & 72.0$\times$10\textsuperscript{9} TFLOPs \\
        Phi2-2.8B~\tnote{\ding{195}} & 2.8 B & 1,400 B & 159.9$\times$10\textsuperscript{9} TFLOPs \\
    \midrule
        Mamba-1.4B~\tnote{\ding{193}} & 1.4 B & 300 B & 2.3$\times$10\textsuperscript{9} TFLOPs \\
        ShearedLLaMA-1.3B & 1.3 B & 50 B & 0.4$\times$10\textsuperscript{9} TFLOPs \\
        CerebrasGPT-1.3B & 1.3 B & 26 B & 0.2$\times$10\textsuperscript{9} TFLOPs \\
        OPT-1.3B & 1.3 B & 180 B & 1.0$\times$10\textsuperscript{9} TFLOPs \\
        BLOOM-1.1B & 1.1 B & 341 B & 2.3$\times$10\textsuperscript{9} TFLOPs \\
        Pythia-1.4B & 1.4 B & 300 B & 2.5$\times$10\textsuperscript{9} TFLOPs \\
        TinyLLaMA-1.1B~\tnote{\ding{194}} & 1.3 B & 2,500 B & 16.5$\times$10\textsuperscript{9} TFLOPs \\
        Falcon-1B & 1 B & 350 B & 2.1$\times$10\textsuperscript{9} TFLOPs \\
        Phi1.5-1.3B~\tnote{\ding{195}} & 1.3 B & 150 B & 17.5$\times$10\textsuperscript{9} TFLOPs \\
        Qwen-1.8B & 1.8 B & 2,200 B & 23.8$\times$10\textsuperscript{9} TFLOPs \\
    \midrule
        \textsc{MiniMA}~\tnote{\ding{196}} & 3 B & 126 B & 4.0$\times$10\textsuperscript{9} TFLOPs \\
    \bottomrule
    \end{tabular}
    \begin{tablenotes}
        \item [\ding{192}] Mistral is rumored to be trained on 8,000 B tokens, and we here choose 4,000 for estimation. Mixtral is an MoE model, in which 2 out of 8 experts are effective.
        \item [\ding{193}] Mamba is a state space model benefiting for long-context inference, and its compute is estimated from the original paper. 
        \item [\ding{194}] TinyLLaMA is planned to be on 3,000 B tokens, but as of December 11st it has been trained on 2,500 B tokens.
        \item [\ding{195}] Phi is trained on tokens generated or filtered by GPT4, so the number of teacher model parameters is estimated according to the rumor that GPT4 is a 2-in-16 MoE model with 1.8 trillion parameters. 
        \item [\ding{196}] \textsc{MiniMA} is distilled from a 7 B teacher.
    \end{tablenotes}
    \end{threeparttable}
    \end{adjustbox}
    \label{tab:est_compute}
\end{table}

The results in Table~\ref{tab:more_standard} shows an akin trend as that of Table~\ref{tab:minima_standard_bench}, likewise indicating the superiority of \textsc{MiniMA}.

\section{Estimates of Training Compute}
\label{app:est_compute}

The training compute estimation is quoted from~\citet{FLOPs22}, which explains the compute constitutions in an organized manner. Plus to that, this estimate is also mentioned and used in~\citet{DBLP:journals/corr/KaplanMH20} as:
\begin{equation}
    C=\underbrace{6\cdot N\cdot D}_{\text{total compute}}=\underbrace{2\cdot N\cdot D}_{\text{forward compute}}+\underbrace{4\cdot N\cdot D}_{\text{backward compute}},
\end{equation}
where $C$ is the training compute, $N$ is the number of model parameters, and $D$ the number of consumed tokens. And the total compute can be decomposed to forward and backward parts.

Particularly for distillation that additionally engages the inference compute of the teacher, the training computed should be expanded as:
\begin{equation}
    C^{\sf distillation}=\underbrace{6\cdot N^{\sf student}\cdot D}_{\text{student compute}}+\underbrace{2\cdot N^{\sf teacher}\cdot D}_{\text{teacher compute}},
\end{equation}
where $N^{\sf student}$ and $N^{\sf teacher}$ indicate the numbers of teacher and student model parameters, respectively.

So here we attach the training compute of considered LMs in Table~\ref{tab:est_compute}, with necessary reference sources.

\begin{table*}[ht]
    \centering
    \caption{The results of \textsc{MiniChat} on standard benchmarks.}
    \begin{adjustbox}{width=0.75\textwidth,center}
    \begin{tabular}{lcccccc}
    \toprule
        \textbf{LM} & \makecell[c]{\textbf{MMLU}\\\textbf{Acc}} & \makecell[c]{\textbf{CEval}\\\textbf{Acc}} & \makecell[c]{\textbf{DROP}\\\textbf{EM Score}} & \makecell[c]{\textbf{BBH}\\\textbf{EM Score}} & \makecell[c]{\textbf{GSM8K}\\\textbf{Maj1@1}} & \makecell[c]{\textbf{HumanEval}\\\textbf{Pass@1}} \\
    \midrule
        OpenLLaMA-3B~\citeyearpar{OpenLLaMA23} & 26.70 & 26.30 & 20.14 & 30.56 & 3.11 & 0.00 \\
        OpenBuddy-3B~\citeyearpar{OpenBuddy23} & 23.88 & 24.67 & 15.89 & 29.08 & 11.07 & 3.66 \\
        $\Delta$ & \textcolor{red}{-2.82} & \textcolor{red}{-1.63} & \textcolor{red}{-4.25} & \textcolor{red}{-1.48} & \textcolor{green}{+7.96} & \textcolor{green}{+3.66} \\
    \midrule
        \textsc{MiniMA} & 28.51 & 28.23 & 22.50 & 31.61 & 8.11 & 10.98 \\
        \textsc{MiniChat} & 38.40 & 36.48 & 22.58 & 31.36 & 29.72 & 18.29 \\
        $\Delta$ & \textcolor{green}{+9.89} & \textcolor{green}{+8.25} & \textcolor{green}{+0.08} & \textcolor{red}{-0.25} & \textcolor{green}{+21.61} & \textcolor{green}{+7.31} \\
    \bottomrule
    \end{tabular}
    \end{adjustbox}
    \label{tab:minichat_bench}
\end{table*}

\section{Evaluation Protocols of GPT4 Assessments}
\label{app:gpt4_eval_proto}

Vicuna-Bench~\citep{Vicuna23} is an instruction-following dataset encompassing 80 instructions from 9 fields. And BELLE-Bench is a Chinese counterpart of Vicuna-Bench comprised of 1,000 instructions from 10 fields. Note that both of them only examine single-turn capability, and we could incorporate multi-turn judgements in the future~\citep[e.g., MT-Bench,][]{DBLP:journals/corr/ZhengCS23}

In commonsense, it is more convincing to grade responses via human labor. However, regarding the labor intensity, we instead derive an evaluation protocol using GPT4 following their recommendations. Concretely, given a pair of responses from two LMs, GPT4 is instructed to score each of them in a head-to-head manner. And different instructions are offered to GPT4 for distinguished considerations of general, coding, and math questions. Please refer to the instructions in~\citet{Vicuna23}. Macro average scores are further computed as summaries. Besides the pairwise scores, we also deduce win-lose-tie proportions correspondingly from them.




\section{Benchmarking Results of \textsc{MiniChat}}
\label{app:minichat_bench}

We include the results of \textsc{MiniChat} on standard benchmarks in case of potential curiosities, since it is long suspected that instruction data could boost benchmarking performance.

The results in Table~\ref{tab:minichat_bench} indicate that proper finetuning could indeed promote the benchmarking performance to certain degrees. In spite of this, it is not very clear whether the improvements are resulted by data contamination or not.

\section{Field-specific Scores in GPT4 Assessments}
\label{app:gpt4_eval_field}

We also provide field-specific scores of regarded baselines and \textsc{MiniChat} for possible interests of rigorous studies.

\begin{figure*}[]
    \centering
    \includegraphics[width=0.92\textwidth]{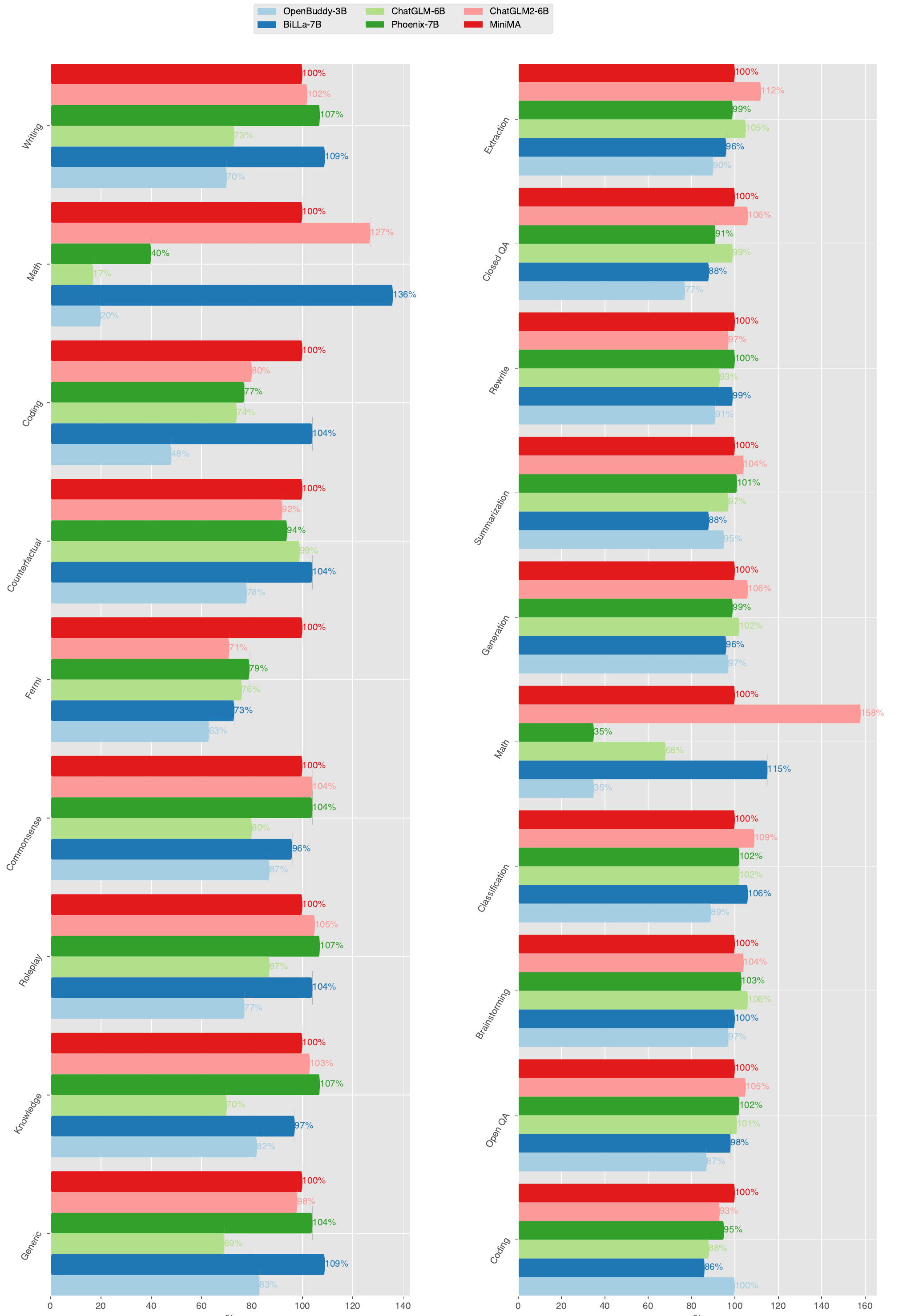}
    \caption{The field-specific results of \textsc{MiniChat} on GPT4 assessments. The scores are rebased to relative percents with respect to those of \textsc{MiniChat}.}
    \label{fig:minichat_field_score}
\end{figure*}

The results in Figure~\ref{fig:minichat_field_score} again show that \textsc{MiniChat} is cost-effective across fields. 

\begin{table}[ht]
    \centering
    \caption{The results of \textsc{MiniChat} on human assessments. Elo ratings are reported.}
    \begin{adjustbox}{width=0.49\textwidth,center}
    \begin{tabular}{lccc}
    \toprule
        \textbf{LM} & \makecell[c]{\textbf{Vicuna-Bench}\\\textbf{Elo Rating}} & \textbf{Human Rank} & \textbf{GPT4 Rank} \\
    \midrule
        \textsc{MiniChat} & 1013.9 & 1 & 2\\
        Phoenix-7B & 1010.8 & 2 & 4\\
        ChatGLM2-6B & 1004.2 & 3 & 3\\
        BiLLa-7B & 996.0 & 4 & 1 \\
        OpenBuddy-3B & 992.0 & 5 & 6 \\
        ChatGLM-6B & 983.2 & 6 & 5 \\
    \bottomrule
    \end{tabular}
    \end{adjustbox}
    \label{tab:minichat_human_assess}
\end{table}

\section{Human Assessments}
\label{app:human_elo}

It is for a long time doubted that whether GPT4 is a convincing referee in response assessment. On that account, we conduct a fast human assessment to verify the consistency between GPT4 and human in judging the quality of model responses. As it is too expensive to ask humans to evaluate these responses in a head-to-head style, we learn from~\citet{DBLP:journals/corr/ZhengCS23} and instead adopt an elo rating-based arena. The arena quickly reduces the evaluation complexity from $\mathcal{O}$(\#models$\times$\#questions) to $\mathcal{O}$(\#questions). To further simplify the arena, we attain the models responses to questions in the dataset in advance. During the whole process, model identities are anonymized from humans. Multiple experienced researchers are recruited to mitigate human biases. 

The results in Table~\ref{tab:minichat_human_assess} inform that the rank disparity between GPT4 and human is computed as a spearman correlation between two rank permutations, i.e., 0.54, so the consistency is roughly 54\%, which is not very high.

To deliver more intuitions on how \textsc{MiniChat} performs, we select a few representative cases here to display. 

The cases in Appendix~\ref{app:comp_case} picture that \textsc{MiniChat} is a good problem-solver.

\section{Comparative Cases}
\label{app:comp_case}

In Table~\ref{tab:minichat_comparative_case}, we supply two cases that are separately concerned about coding and writing and attach them with responses from OpenBuddy-3B, ChatGLM2-6B, and \textsc{MiniChat} for better comparisons.

In the coding case, we find that: OpenBuddy-3B follows the problem requirements but fails to solve the problem; ChatGLM2-6B fails to follow the problem requirements that extra data structures should not be used, and also fails to implement a correct hash table class; \textsc{MiniChat} gives a neat solution.

In the writing case, we find that: OpenBuddy-3B gives an overly general blog mentioning only two items ; ChatGLM2-6B mentions a few more respectively for cultural experiences and must-see attractions, however, it formats the blog very much like a scientific paper rather than a blog; on the contrary, \textsc{MiniChat} seems to have more elegant language but uses very interleaved logistics.

\section{MiniMA Family}
\label{app:minima_family}

The details should be left to their model cards on the HuggingFace, since they are marginally related to this paper.

\begin{itemize}
    \item \textsc{MiniMA}-2-3B. This is a model continued from \textsc{MiniMA} with a more diverse mixture of data.
    \item MiniChat-2-3B. This is an instruction-following model trained from \textsc{MiniMA}-2-3B. 
    \item MiniLoong-3B. This is a long-context model extended from \textsc{MiniMA}-2-3B with the aid of distributed attention~\citep{DBLP:journals/corr/JacobsTZ23}.
    \item MiniMix-2/4x3B. This is a mixture-of-experts model upcycled from \textsc{MiniMA}-2-3B~\citep{DBLP:conf/iclr/KomatsuzakiPLRM23}.
\end{itemize}

\begin{table*}[]
    \centering
    \caption{The comparative cases.}
    \begin{adjustbox}{width=0.99\textwidth,center}
    \begin{tabular}{l}
    \toprule
        \textbf{Question:} Implement a program to find the common elements in two arrays without using any extra data structures. \\
    \midrule
        \textbf{OpenBuddy-3B:} To find the common elements in two arrays without using any extra data structures, you can use the following alg-\\orithm:\\\\1. Sort the two arrays in ascending order (using an appropriate sorting function).\\2. Compare the first element of each array. If it's the same, then the rest of the array is common.\\3. If the first element is different, then the rest of the array is common.\\4. Repeat steps 2 and 3 until either all the elements in the first array are different or all the elements in the second array are different.\\\\Here is the Python code for the algorithm:\\\begin{lstlisting}[language=python]
        def find_common_elements(arr1, arr2):
            sorted_arr1 = sorted(arr1)
            sorted_arr2 = sorted(arr2)
            for i in range(len(sorted_arr1)):
                if sorted_arr1[i] == sorted_arr2[i]:
                    return sorted_arr1
            return []
        \end{lstlisting}\\This algorithm can be used to find the common elements in any two arrays. \\
    \midrule
        \textbf{ChatGLM2-6B:} One way to implement a program to find the common elements in two arrays without using any extra data structu-\\res is to use a hash table. A hash table is a data structure that stores key-value pairs and allows for fast lookup, insertion, and delet-\\ion of elements.\\\\To use a hash table to find the common elements in two arrays, you would first need to create a hash table with a key as the index and\\a value as the key-value pair. Then, for each element in the first array, you would use the hash table to look up the element and com-\\pare it to the elements in the second array. If a match is found, the element and its index would be added to the hash table.\\\\Here is a simple example of how this could be implemented in Python:\\\begin{lstlisting}[language=python]
        class Element:
            def __init__(self, value):
                self.value = value
                
        class HashTable:
            def __init__(self):
                self.table = {}
                
            def put(self, key, value):
                self.table[key] = value
                
            def get(self, key):
                return self.table.get(key)
                
            def contains(self, key):
                return self.table.contains(key)
                
        # Example usage
        element1 = Element('a')
        element2 = Element('b')
        table = HashTable()
        table.put('a', element1)
        table.put('b', element2)
        print(table.get('a'))  # prints the Element object for 'a'
        print(table.contains('a'))  # prints True
        print(table.get('z'))  # raises a KeyError\end{lstlisting}\\Note that this approach has a time complexity of O(n * m), where n and m are the lengths of the arrays, which may be slow for large\\arrays. There are other data structures that could be used to implement this, such as a linked list or a tree-based data structure, but a\\hash table is a simple and efficient solution for this problem. \\
    \midrule
        \textbf{\textsc{MiniChat}:} Here is a Python implementation:\\\begin{lstlisting}[language=python]
        def common_elements(arr1, arr2):
            if len(arr1) == 0:
                return []
            if len(arr2) == 0:
                return arr1
            common_elements = []
            for element in arr1:
                if element in arr2:
                    common_elements.append(element)
            return common_elements
        \end{lstlisting} \\
    \midrule
        \textbf{Question:} Compose an engaging travel blog post about a recent trip to Hawaii, highlighting cultural experiences and must-see attractions. \\
    \midrule
        \textbf{OpenBuddy-3B:} Hi there! I am excited to share with you my recent trip to Hawaii and all the amazing cultural experiences I had during\\my stay. From the stunning beaches to the vibrant culture, I fell in love with this beautiful island state. \\\\First of all, I must say that I was blown away by the incredible natural beauty of Hawaii. The lush green landscapes, crystal-clear waters,\\
    \end{tabular}
    \end{adjustbox}
    \label{tab:minichat_comparative_case}
\end{table*}

\begin{table*}[]
    \centering
    \begin{adjustbox}{width=0.99\textwidth,center}
    \begin{tabular}{l}
        and stunning sandy beaches all left a lasting impression on me. My favorite part of my trip was kayaking through the beautiful Wailea\\coastline, where I saw dolphins, seahorses, and even a couple of whale sharks swimming in the ocean. It was such a surreal and unforget-\\table experience that I will always remember.\\\\Another highlight of my trip was visiting the Polynesian cultural center, where I experienced traditional Polynesian dance and music. I \\was fascinated by the intricate and complex dances that the performers were performing, and I even managed to join in on a few of the\\demonstrations. It was an incredible moment that I will never forget.\\\\Furthermore, I had the opportunity to experience the vibrant cultural lifestyle of Hawaii. From attending traditional Hawaiian festivals to\\visiting local art galleries, I felt like I was truly immersed in the local culture. I even had the chance to try some delicious Hawaiian cuis-\\ine, such as poke bowls and plate lunches, which were some of the most delicious dishes I have ever tasted.\\\\Finally, I have to say that my trip to Hawaii was nothing short of incredible. The stunning beaches, warm weather, and vibrant culture\\made for the perfect vacation. I can't wait to return to this beautiful island state and explore even more of its rich cultural heritage.\\\\If you're planning a trip to Hawaii, I highly recommend taking the time to experience the local culture and attractions. It will be an unfo-\\rgettable experience that you won't want to miss. \\
    \midrule
        \textbf{ChatGLM2-6B:} Title: A Journey Through the Aloha State: Exploring Hawaii's Cultural Experiences and Must-See Attractions\\\\Introduction\\Hawaii, a state known for its lush greenery, stunning landscapes, and tropical climate, is a paradise for many. But there's more to Hawaii\\than what meets the eye. This recent trip to the Aloha State offered me a unique opportunity to immerse myself in the local culture and\\traditions. As I discovered, Hawaii is a place where one can truly feel at home and experience the perfect blend of relaxation, adventure,\\and culture.\\\\Cultural Experiences\\Hawaii is a place where culture and tradition are deeply ingrained in the very fabric of the island. I had the chance to experience this first-\\hand by attending a traditional Hawaiian luau, a dinner party that is a staple of the Aloha State. The luau was a fascinating event that inc-\\luded a host of local dishes, music, and dance. I was amazed by the energy and excitement of the party, and I felt truly at home in the mi-\\dst of the Aloha State.\\\\Another way to experience the local culture in Hawaii is by visiting the Pu'uhonua o Honaunau National Historical Park. This park is ho-\\me to a collection of ancient Hawaiian temples, and it offers a glimpse into the island's spiritual traditions. I found it fascinating to learn\\about the history and significance of these temples, and I appreciated the opportunity to reflect on the culture and traditions of Hawaii.\\\\Must-See Attractions\\Of course, any trip to Hawaii is not complete without visiting some of the state's many natural attractions. I had the chance to snorkel in\\the crystal-clear waters of the Polynesian Islands, and I was amazed by the beauty of the coral reefs and the diverse array of marine life. I\\also took the opportunity to visit the Waikato Waterfall, a natural wonder that is a must-see attraction in Hawaii.\\\\Another must-see attraction in Hawaii is the Diamond Head, a volcanic crater that offers stunning views of the surrounding landscape. I\\took the hike up to the top of the crater and was rewarded with breathtaking views of Honolulu and the ocean below.\\\\Conclusion\\Hawaii is a place where culture, tradition, and natural beauty are all around. I was truly touched by the Aloha State and its unique way\\of life. I hope you'll have the chance to experience it for yourself and discover the beauty and magic of Hawaii. \\
    \midrule
        \textbf{\textsc{MiniChat}:} Visiting Hawaii: A Journey Through the Ocean's Heart\\\\As I step off the plane, I am immediately greeted by the warm, inviting sights and sounds of Hawaii. The colors of the sunset are breath-\\taking, and the sound of the waves is soothing. I feel as though I am in a different world, surrounded by the beauty of nature. \\\\My first stop is Waikiki Beach, where I take a leisurely stroll along the sandy shoreline, feeling the softness of the golden sand between\\my toes. The ocean is crystal clear, with gentle waves gently lapping at the shore. I stop for a moment to watch the surfers as they\\ride the waves and catch the sun. \\\\After soaking up the sun, I head to Pearl Harbor to pay my respects at the USS Arizona Memorial at the National Memorial Cemetery of\\the Pacific. As I walk through the exhibits, I am reminded of the sacrifices that were made during World War II. The USS Arizona Muse-\\um is a moving tribute to the bravery and sacrifices of the sailors and soldiers who gave their lives for their country. \\\\Next, I make my way to Hanauma Bay, where I go scuba diving in the crystal-clear waters. The visibility is excellent, and the coral reefs\\are a sight to behold. The fish are vibrant, and the colors are so vivid that it's like a dream. \\\\After an exhilarating day of exploring, I head to the historic town of Hilo to experience the local culture. The Hilo Museum is a great pla-\\ce to learn about the history of the area, and the Hilo Bay Fishpond is a fascinating example of traditional Hawaiian aquaculture. \\\\My trip to Hawaii has been an unforgettable experience. Whether it be the stunning natural beauty or the rich cultural heritage, there is so-\\mething for everyone in this incredible destination. I am already planning my next visit, and I know that I will be back to explore even mo-\\re of what Hawaii has to offer. \\
    \bottomrule
    \end{tabular}
    \end{adjustbox}
\end{table*}

\end{document}